\documentclass[a4paper,oneside,10pt]{article}%
\usepackage{float}
\usepackage{adjustbox}
\usepackage{amsmath}
\usepackage{amsfonts}
\usepackage{amssymb}
\usepackage{multirow}
\usepackage{graphicx}
\usepackage{subcaption}
\usepackage{url}

\graphicspath{{figures/ex1/}}

\usepackage{epsfig}
\usepackage{booktabs}
\usepackage[square,numbers,sort&compress]{natbib}
\usepackage{diagbox}%
\usepackage[pagewise]{lineno}
\usepackage[ruled,vlined]{algorithm2e} 
\usepackage{caption}
\usepackage{threeparttable}
\usepackage{makecell}
\usepackage[title]{appendix}

\providecommand{\U}[1]{\protect\rule{.1in}{.1in}}
\allowdisplaybreaks[4]

\newtheorem{theorem}{Theorem}[section]

\newtheorem{lemma}[theorem]{Lemma}

\newtheorem{remark}[theorem]{Remark}

\newenvironment{proof}[1][Proof]{\noindent \textbf{#1.} }{\  \rule{0.5em}{0.5em}}
\numberwithin{equation}{section}
\renewcommand \arraystretch{1.5}
\begin{document}

\title{Deep backward multistep high-order regression methods for high dimensional nonlinear parabolic partial differential equations
}
\author{Qiang Han \thanks{School of Mathematics, Yangzhou University, Yangzhou 225002, China. Email: hanqiang@yzu.edu.cn.}
\and Shaolin Ji\thanks{Zhongtai Securities Institute for Financial Studies,
Shandong University, Jinan, Shandong 250100, China. jsl@sdu.edu.cn.}
\and Zihao Xv\thanks{
School of Mathematics, Yangzhou University, Yangzhou 225002, China. Email: MX120240321@stu.yzu.edu.cn.}
}

\maketitle

\textbf{Abstract}.
We propose a deep backward multistep high-order regression (MHDBR) method
for semilinear parabolic partial differential equations through their
backward stochastic differential equation representations. The method
combines multistep time discretization with regression targets formed
by conditional expectations. Under explicit regularity, starting-value,
forward-approximation and network-moment assumptions, we prove an error
bound separating time discretization, conditional Monte Carlo sampling
and neural-network approximation. The analysis assumes exact population
minimization; finite-training-sample generalization and optimization
errors are not included. For generators independent of the gradient
variable, suitably constrained GroupSort networks and sufficiently large
conditional sample sizes give order $J$ in the stated nodal $L^2$ error
norm. The required network sizes depend on dimension, so this result
does not establish dimension-independent complexity. We present a
reproducible time-discretization test satisfying the analytical
assumptions and three broader neural-network benchmarks, whose
relation to the assumptions is discussed explicitly.



\addcontentsline{toc}{section}{\hspace*{1.8em}Abstract}

\section{Introduction}
This paper develops a neural network architecture grounded in probabilistic multistep high-order methods for high-dimensional semilinear parabolic partial differential equations (PDEs)
\begin{equation}
\begin{aligned}
\begin{cases}
\partial_t u+\mu \cdot D_x u+\frac{1}{2} \operatorname{Tr}\left(\sigma\sigma^{\top} D_x^2 u\right)+f\left(\cdot, \cdot, u, D_x u \sigma\right) =0 & \text { on }[0, T) \times \mathbb{R}^d, \\
u(T, \cdot)=g(\cdot)& \text { on } \mathbb{R}^d,
\end{cases}
\end{aligned}
\label{PDE}
\end{equation}
where $\mu$ maps $[0,T]\times\mathbb{R}^{d}\rightarrow \mathbb{R}^{d}$ and $\sigma$ maps $[0,T]\times\mathbb{R}^{d}\rightarrow \mathbb{R}^{d\times d}$;
$f:[0,T]\times\mathbb{R}^{d}\times\mathbb{R}^{n}\times\mathbb{R}^{n\times d}\rightarrow\mathbb{R}^{n}$
and $g:\mathbb{R}^{d}\rightarrow \mathbb{R}^{n}$  represent the nonlinear generator and the terminal function, respectively;
the operators $D_x$ and $D_x^2$ refer to the first and second order spatial derivatives;
the symbol $\cdot$ represents the scalar product and $^\top$ denotes the transpose of a vector or matrix.
For $n>1$, the differential operators in \eqref{PDE} act componentwise.
Unless stated otherwise, $|\cdot|$ denotes the Euclidean norm for vectors
and the Frobenius norm for matrices. For a square-integrable random
vector or matrix $\xi$, write
$\mathbb D_i[\xi]=\mathbb E_i|\xi-\mathbb E_i\xi|^2$ and
$\|\xi\|_2=(\mathbb E|\xi|^2)^{1/2}$.
Appendices B--C explicitly specify the different output norm used
only for the spatial Lipschitz constraints.

PDEs characterize the continuous evolution of physical quantities in space and time, serving as a fundamental mathematical framework for diverse disciplines such as fluid mechanics, quantum mechanics, financial mathematics, and mathematical biology and so on. From the Fourier equation governing heat conduction and the Navier-Stokes equations describing fluid motion, to the Black-Scholes equation for option pricing and the Schr\"{o}dinger equation for quantum state evolution, PDEs are ubiquitous in scientific modeling. However, the vast majority of practical PDEs lack closed-form solutions, necessitating numerical approximations. Consequently, the development of efficient, stable, and highly accurate numerical methods for PDEs remains a central research theme in computational mathematics and scientific computing.

Tensor-product finite difference, finite element and spectral grids
can become prohibitively expensive as the dimension increases.
Sparse grids \cite{BHJGM04}, low-rank tensor methods \cite{KBN14}
and probabilistic methods \cite{FATNWX11,GWZJZJ15,GELSJGTPVC16}
exploit additional structure, with efficiency depending on the problem.
High-dimensional PDEs arise, for example, in multi-asset pricing,
stochastic control and quantum many-body models. The growth of the
state space with dimension motivates alternatives to full spatial grids.

Neural networks provide flexible function approximators, while automatic
differentiation and stochastic sampling are useful for differential
operators and high-dimensional expectations. Universal approximation
\cite{HKSMWH89} guarantees density on compact sets under suitable
activation assumptions, but by itself gives neither a polynomial
parameter bound in dimension nor a training guarantee. Deep learning
approaches to PDEs can be grouped into deterministic and probabilistic
methods.

The first branch comprises deterministic approach, including physics-informed neural networks proposed by \cite{RMPPKGE19}, 
the deep Galerkin method \cite{SJSK18},
the deep Ritz method \cite{EWYB18} and
the weak adversarial method \cite{ZYBGYXZH20}.
When specific boundary conditions are enforced, the standard loss function must be suitably adapted. For example, the deep Nitsche method  \cite{LYWP21} extends the deep Ritz method by incorporating Nitsche’s technique to address essential boundary conditions, whereas the penalty-free neural network method \cite{SHYC21} employs an auxiliary network to strictly enforce boundary constraints.
It is worth noting that there exist many other efficient physics-informed neural networks based deep learning methods for solving high-dimensional PDEs, which we do not enumerate exhaustively here.

The second is the stochastic approach, which utilizes the Feynman-Kac formula \cite{PEPS92,PS91} to transform the PDE into a backward stochastic differential equation (BSDE) and employs neural networks to approximate the solutions and their gradient  terms within the BSDE. The deep BSDE method introduced by  \cite{HJJAEW18} is the pioneering  deep learning method developed for high dimensional PDEs.  \cite{HJJAEW18} successfully solved the 100-dimensional nonlinear Black–Scholes equation, the 100-dimensional Hamilton–Jacobi–Bellman equation, and the 100-dimensional Allen–Cahn equation, reporting relative errors on the order of $10^{-2}$ to $10^{-3}$ in these experiments.
Subsequently, various variants of the deep BSDE method have been developed (see \cite{EWHMJAKT19,HJHRLJ24,JSPSPYZX20,JSPSPYZX22,JSPSPYZX25,HJHWLJZY26,CWFSZT25,CWFSZT25a} and many others). To derive better stability and accuracy properties, the deep backward dynamic programming method \cite{HCPHWX20},
the deep splitting method \cite{BCBSCPJANA21} and 
the deep backward regression  method \cite{HQJSLY26}
are designed to optimize the deep BSDE method.

To provide better accuracy, the 
deep backward multistep dynamic programming method (MDBDP)
\cite{GMPHWX22} is proposed. 
It estimates the solutions and their gradient  by the loss functions, for $i=N-1,N-2,\cdots,0$
\begin{footnotesize}
\begin{equation}\label{MDBDP-loss}
J_i^{HPW}(\theta_{i}) = \mathbb{E}\left[\left|
\mathcal{Y}_i - g(X_N^\pi) 
- \sum\limits_{j=i+1}^{N-1}\left(hf(t_j,X_j^\pi,\mathcal{Y}_j^*,\mathcal{Z}_j^*)
- \mathcal{Z}_j^* \Delta W_{j,j+1}\right)
-hf(t_i,X_i^\pi,\mathcal{Y}_i,\mathcal{Z}_i) +\mathcal{Z}_i\Delta W_{i,i+1}
\right|^2\right],
\end{equation}
\end{footnotesize}
at each time step $t_i$, where $\mathcal{Y}_i: \mathbb{R}^d \rightarrow \mathbb{R}^n$ and
$\mathcal{Z}_i: \mathbb{R}^d \rightarrow \mathbb{R}^{n\times d}$ denote the sequences of network functions. The optimized pair $(\mathcal{Y}_i^*,\mathcal{Z}_i^*)$ is obtained by solving the corresponding local
minimization problem.
The loss function (\ref{MDBDP-loss}) is derived from the multistep forward dynamic programming equation \cite{GETP16}  for the time-discretization of the BSDE. 
The loss function \eqref{MDBDP-loss} uses an Euler-based time
discretization. Higher temporal order can be useful when sufficient
solution regularity is available and the remaining error sources are
controlled. Pathwise stochastic targets also have sampling variance,
which can affect regression and optimization. These observations
motivate the present construction, but do not imply a universal
accuracy or computational-cost ranking of neural BSDE methods.

To address these problems, we propose a new deep backward 
multistep high-order
regression-based (MHDBR) method to approximate the solutions and their gradient of high dimensional PDEs by the loss functions
\begin{equation}\label{DBR-loss-y}
F_{y,i}(\theta_{y,i}) = \mathbb{E}\left[ \left|\mathcal{Y}_i - \mathbb{E}_i\left[\mathcal{Y}_{i+1}^*
    +\alpha_{0}hf(t_i,X_i^\pi,\mathcal{Y}_i,\mathcal{Z}_i^* ) 
    +\sum\limits_{j=1}^{J} \alpha_{j}hf(t_{i+j},X_{i+j}^{\pi},\mathcal{Y}_{i+j}^*,\mathcal{Z}_{i+j}^*)
    \right]
    \right|^2\right],
\end{equation}
and
\begin{equation}\label{DBR-loss-z}
F_{z,i}(\theta_{z,i}) = \mathbb{E}\left[ \left|\mathcal{Z}_i 
-\sum\limits_{j=0}^{J}\beta_{J,j}\mathbb{E}_{i}\left[\mathcal{Y}_{i+j}^*\Delta W_{i,i+j}^\top\right]
    \right|^2\right],
\end{equation}
where $\mathbb{E}_i\left[\cdot\right] = \mathbb{E}\left[\cdot\big|\mathcal{F}_{t_i}\right]$
is the conditional expectation; $\{\alpha_j\}_{0\le j\le J}$ and $\{\beta_{J,j}\}_{0\le j\le J}$ denote parameters. The above multistep high-order regression-based loss functions  use conditional expectations to construct the regression targets.
For any square-integrable target $H$, conditional expectation satisfies
$\mathbb E|H|^2=\mathbb E|\mathbb E_iH|^2+
\mathbb E|H-\mathbb E_iH|^2$. In this precise sense it removes
conditional target variance. Its Monte Carlo approximation still has
sampling error, and this identity alone does not establish lower
stochastic-gradient variance or better generalization. The MHDBR
construction is a neural-network realization of the multistep scheme.

The remainder of this paper is organized as follows.
In Section 2, some essential materials  that are used in the whole paper are provided.
The MHDBR method is designed in Subsection 3.1;
The upper bound error of the proposed MHDBR method is rigorously derived theoretically in Subsection 3.2.
In addition, we provide the convergence analysis for the  proposed MHDBR method.
We report the numerical results in Section 4.
In the end, we compile the conclusion of this paper in Section 5.
For brevity,  several  auxiliary results are deferred to the Appendix.

\section{Preliminaries}

In this section, we introduce the notation and preliminary results used throughout the paper. Specifically, we review the probabilistic representation of PDEs, define the relevant neural network function classes, and propose the multistep high order scheme for forward backward stochastic differential equations (FBSDEs).

\subsection{Deep neural networks as function approximators}

In the numerical schemes developed below, neural networks are employed to
parametrize the deterministic functions arising from the conditional
expectations in the time-discretized backward equations. This choice is
supported by the classical universal approximation results \cite{HKSMWH89}, which establish that feedforward neural
networks with suitable activation functions are dense in broad classes of
functions, including the space of continuous functions on compact sets.
We next introduce the notation used to describe the corresponding network
classes.
Let \(d_{1},d_{2}\in\mathbb{N}\), and let
\(\overline{\varrho}:\mathbb{R}\to\mathbb{R}\) be a scalar activation
function. Its componentwise extension to \(\mathbb{R}^{d_{2}}\)
is denoted by
$\varrho(z)=
\bigl(
\overline{\varrho}(z_1),\ldots,
\overline{\varrho}(z_{d_{2}})
\bigr)^\top$, $z\in\mathbb{R}^{d_{2}}.$
We define the collection of activated affine layers by
\[
\mathcal{LL}_{d_{1},d_{2}}^{\varrho}
=
\left\{
\psi:\mathbb{R}^{d_{1}}
\to\mathbb{R}^{d_{2}}
\;:\;
\psi(x)=\varrho(\mathcal{W}x+\beta),
\quad
(\mathcal{W},\beta)
\in
\mathbb{R}^{d_{2}\times d_{1}}
\times\mathbb{R}^{d_{2}}
\right\}.
\]
When no activation is applied, that is, when \(\varrho\) is the identity
mapping, we write
\[
\mathcal{LL}_{d_{1},d_{2}}
=
\left\{
x\mapsto \mathcal{W}x+\beta
\right\}.
\]

Let \(L\in\mathbb{N}\) be the number of hidden layers and let
$
\mathbf{n}=(n_0,n_1,\ldots,n_{L-1})
$
specify their widths. For an input dimension \(d_1\), an output dimension
\(d_{2}\), and hidden-layer activation functions
$\boldsymbol{\varrho}=(\varrho_0,\varrho_1,\ldots,\varrho_{L-1})$,
we introduce the feedforward neural-network class
\[
\begin{aligned}
\mathcal{NN}_{d_1,d_{2}}^{\boldsymbol{\varrho},L,\mathbf{n}}
=
\Bigl\{&
\varphi:\mathbb{R}^{d_1}\to\mathbb{R}^{d_{2}}
\;:\;
\varphi
=
\psi_L\circ\psi_{L-1}\circ\cdots\circ\psi_0,
\quad
\psi_0
\in
\mathcal{LL}_{d_1,n_0}^{\varrho_0},
\\
&
\psi_\ell
\in
\mathcal{LL}_{n_{\ell-1},n_\ell}^{\varrho_\ell},
\quad
1\leq\ell\leq L-1,
\quad
\psi_L
\in
\mathcal{LL}_{n_{L-1},d_{2}}
\Bigr\}.
\end{aligned}
\]
Thus, nonlinear activation functions are applied in the hidden layers,
whereas the output layer is affine. The tuple
$
\bigl(
d_1,d_{2},L,\mathbf{n},
\boldsymbol{\varrho}
\bigr)
$
determines the architecture of the network, while the weight matrices and
bias vectors constitute its trainable parameters.
The aforementioned approximation results concern representational
capacity: under appropriate assumptions on the activation function and
with sufficiently many neurons, functions in the relevant approximation
space can be approximated with arbitrary accuracy. 

\subsection{Probabilistic representation for the solutions of PDEs}
The numerical methods proposed in this work are based on the connection
between semilinear parabolic PDEs and FBSDEs. Let $(\Omega,\mathcal{F},(\mathcal{F}_t)_{0\le t\le T},\mathbb{P})$ be a complete filtered probability space satisfying the usual conditions
and supporting a \(d\)-dimensional standard Brownian motion
\(W=(W_t)_{0\leq t\leq T}\). We assume that \((\mathcal{F}_t)_{0\le t\le T}\) is the usual
augmentation of the filtration generated by \(W\).
Consider the decoupled FBSDEs
\begin{equation}
\left\{
\begin{aligned}
dX_{t}= &\mu(t,X_{t})dt+\sigma(t,X_{t})dW_{t},\quad X_{0}= x_0,\\
-dY_{t}= &f(t,X_{t},Y_{t},Z_{t})dt-Z_{t}dW_{t},\quad Y_{T}= g(X_{T}),\\
\end{aligned}
\right.  \label{dFBSDE}%
\end{equation}
where $X_t\in\mathbb{R}^d$, $Y_t\in\mathbb{R}^n$, $Z_t\in\mathbb{R}^{n\times d}$
and \(x_0\in\mathbb{R}^d\) is an initial state. Under the standard well-posedness assumptions on
\(\mu,\sigma,f\), and \(g\), the nonlinear Feynman--Kac formula
\cite{PEPS90,PEPS92,PS91} establishes a probabilistic representation of
the solution \(u\) to the PDE \eqref{PDE}. If \(u\) is a sufficiently regular classical solution of
\eqref{PDE}, then the processes defined by
\begin{equation}
Y_t=u(t,X_t),
\qquad
Z_t=D_xu(t,X_t)\sigma(t,X_t),
\label{FK}
\end{equation}
solve the associated FBSDE.


\begin{remark}
The high-order analysis requires the solution regularity stated explicitly
in Assumption 2 below. Smoothness of $f$ and $g$ alone is insufficient:
regularity results also impose conditions on the drift and diffusion
coefficients and, depending on the result, nondegeneracy or other
structural assumptions. We do not deduce Assumption 2 from Assumption 1;
any use of a sufficient regularity theorem, such as those discussed in
\cite{DPGZJ02,GELC07}, requires verification of all its hypotheses.
\end{remark}

For the readers' convenience, before giving the time-discretization scheme of FBSDEs (\ref{dFBSDE}),
we introduce the following notation.
We denote the grid of the time interval $[0,T]$ by $\pi$, namely
$\pi=\{0=t_0<t_1< \cdots< t_N=T\}$, where $t_i=ih, h=\frac{T}{N}$
for $i=0,1,\cdots,N,N\in \mathbb{N}^+$
and $\Delta W_{i,i+j}=W_{t_{i+j}}-W_{t_i}$ the  Brownian motion increment.
Then the stable multistep time-discretization of the BSDE  in (\ref{dFBSDE}), at the mesh points $t_i$, is
\begin{equation}
\left\{
\begin{aligned}
Y_{i}^{\pi}&=  \mathbb{E}_{i}\Big[Y_{i+1}^{\pi}
+ \sum\limits_{j=0}^{J} \alpha_{j}hf(t_{i+j},X_{i+j}^{\pi},Y_{i+j}^{\pi},Z_{i+j}^{\pi})\Big],\\
Z_{i}^{\pi} &=  \sum\limits_{j=0}^{J}\beta_{J,j}\mathbb{E}_{i}\big[Y_{i+j}^{\pi}\Delta W_{i,i+j}^\top\big],
\end{aligned}
\right.
\label{NS}%
\end{equation}
where
$X_{i}^{\pi}$, $Y_{i}^{\pi}$ and $Z_{i}^{\pi}$ denote the time-discrete approximations of $X$, $Y$ and $Z$ at $t_{i}$ respectively;
and $\{\alpha_{j}\}_{0\le j\le J }$ are real numbers  and satisfy the condition:
\begin{equation}
1= \sum\limits_{j=0}^{J} \alpha_{j},
\quad 1= 2\sum\limits_{j=1}^{J} j\alpha_{j},
\quad\cdots,
\quad 1=(J+1)\sum\limits_{j=1}^{J} j^{J}\alpha_{j};\label{parameter-alpha}
\end{equation}
$\{\beta_{J, j}\}_{0\le j\le J }$ are real numbers and satisfy the condition:
\begin{equation}
\left[\begin{array}{ccccc}
1 & 1 & 1 & \cdots & 1 \\
0 & 1 & 2 & \cdots & J \\
0 & 1^2 & 2^2 & \cdots & J^2 \\
\vdots & \vdots & \vdots & \vdots & \vdots \\
0 & 1^J & 2^J & \cdots & J^J
\end{array}\right] \times\left[\begin{array}{c}
\beta_{J, 0} h \\
\beta_{J, 1} h \\
\beta_{J, 2} h \\
\vdots \\
\beta_{J, J}h
\end{array}\right]=\left[\begin{array}{c}
0 \\
1 \\
0 \\
\vdots \\
0
\end{array}\right].
\label{parameter-beta}
\end{equation}

Since the first equation in (\ref{NS}) is implicit
with respect to \(Y_i^\pi\), we impose a standard small time-step condition to ensure
its well-posedness. Let \(L_y\) denote a Lipschitz constant of \(f\) with respect to
the \(y\)-variable. Under \textbf{Assumption 1 (iii)} below, one can take \(L_y\leq L\). We
assume throughout that
\begin{equation}
|\alpha_0|hL_y\leq|\alpha_0|hL\leq\frac{1}{2}.
\label{time-step-condition}
\end{equation}
Indeed, for fixed \(i\), \(X_i^\pi\), and \(Z_i^\pi\), the map
\[
y\longmapsto
\mathbb{E}_i\left[Y_{i+1}^\pi 
+ \sum\limits_{j=1}^{J} \alpha_{j}hf(t_{i+j},X_{i+j}^{\pi},Y_{i+j}^{\pi},Z_{i+j}^{\pi})\right]
+
\alpha_{0}h f(t_i,X_i^\pi,y,Z_i^\pi)
\]
is a contraction on \(\mathbb{R}^n\) whenever \eqref{time-step-condition} holds.
Hence the implicit equation for \(Y_i^\pi\) admits a unique solution at each time
step.

\mbox{}
For the time discretization of the forward SDE in
\eqref{dFBSDE}, we use the one-step scheme
\begin{equation}\label{SDE-general-scheme}
X_{i+1}^{\pi}
=
X_i^{\pi}
+\Psi\bigl(
t_i,X_i^{\pi},h,
\Delta W_{i,i+1},\xi_{i,i+1}
\bigr),
\qquad i=0,\ldots,N-1,
\end{equation}
where $X_0^\pi=x_0$ and
$\Delta W_{i,i+1}=W_{t_{i+1}}-W_{t_i}$.
Here $\Psi$ denotes the numerical increment over
$[t_i,t_{i+1}]$, and $\xi_{i,i+1}$ collects the additional
random variables required by the chosen scheme,
including the iterated stochastic integrals used in
higher-order discretizations (\cite{KPEPE92}).
Particular realizations of $\Psi$ are given in Appendix~A. 

\section{MHDBR methods}

This section details the formulation of the MHDBR method, which leverages learned data to approximate the conditional expectations in (\ref{NS}) and yields a fully implementable algorithm. Furthermore, we provide a comprehensive error analysis and establish the corresponding convergence rate.

\subsection{Deep learning method to approximate the conditional expectation}

The implementation of $(Y_{i}^{\pi},Z_{i}^{\pi})$ via the scheme (\ref{NS}) necessitates approximating the conditional expectations, which we achieve using deep neural networks. The forward process $X_{\cdot}^{\pi,m} $ 
is then simulated via Monte Carlo method as follows,
 for $i=0,1,\cdots,N-1$
\begin{equation}\label{SDE-M}
X_{i+1}^{\pi,m} =X_{i}^{\pi,m} +
\Psi(t_{i},X_{i}^{\pi,m},h,\Delta W^m_{i, i+1},\xi^m_{i,i+1}),  \quad X_{0}^{\pi,m}= x_0,
\end{equation}
where $\Delta W_{i,i+1}^m \sim \mathcal{N}(0, hI_d)$,  $m = 1,2,\cdots,M$, $M\in\mathbb{N}$;
$\xi^m_{i,i+1}$ depends on the time discretization scheme of  SDE in (\ref{dFBSDE}) and  relates to $\Delta W_{i,i+1}^m$.
The backward process $Y_{\cdot}^{\pi,m} $ is simulated by Monte Carlo method in the following ways, for $i=N-J,N-J-1,\cdots,0$
\begin{equation}
\left\{
\begin{aligned}
Y_{i}^{\pi,m}&=  \mathbb{E}_{i}\Big[Y_{i+1}^{\pi,m}
+ \sum\limits_{j=0}^{J} \alpha_{j}hf(t_{i+j},X_{i+j}^{\pi,m},Y_{i+j}^{\pi,m},Z_{i+j}^{\pi,m})\Big],\\
Z_{i}^{\pi,m} &=  \sum\limits_{j=0}^{J}\beta_{J,j}\mathbb{E}_{i}\big[Y_{i+j}^{\pi,m}\cdot(\Delta W_{i,i+j}^m)^\top\big].
\end{aligned}
\right.
\label{NS-M}%
\end{equation}
At time $t_i$, for each outer state $X_i^{\pi,m}$, generate $K_i$
complete conditional branches. Set $X_i^{\pi,m,k}=X_i^{\pi,m}$ and,
for $q=0,\ldots,J-1$, use
\[
X_{i+q+1}^{\pi,m,k}=X_{i+q}^{\pi,m,k}
+\Psi(t_{i+q},X_{i+q}^{\pi,m,k},h,
      \Delta W_{i+q,i+q+1}^{m,k},\xi_{i+q,i+q+1}^{m,k}).
\]
Each one-step pair has the same joint law as the driving variables in
\eqref{SDE-M}, with
$\Delta W_{r,r+1}^{m,k}\sim\mathcal N(0,hI_d)$, and
$\Delta W_{i,i+j}^{m,k}=\sum_{r=i}^{i+j-1}\Delta W_{r,r+1}^{m,k}$.
Given the outer state, the complete branches are independent and
identically distributed across $k$. Within one branch, all future
states and increments share the same increment history. Fresh branch
drivers are independent of the past and of the fixed future network
functions used in the population analysis. The outer chain and all
conditional branches use the same one-step transition kernel.

We now detail the implementation of the approximation algorithm. Specifically, the terms on the right-hand side of (\ref{NS-M}) are expressed as
\begin{equation}
\left\{
\begin{aligned}
Y_{i}^{\pi,m}&=  \frac{1}{K_i}\sum\limits_{k=1}^{K_i} \left(Y_{i+1}^{\pi,m,k}
+\sum\limits_{j=0}^{J} \alpha_{j}hf(t_{i+j},X_{i+j}^{\pi,m,k},Y_{i+j}^{\pi,m,k},Z_{i+j}^{\pi,m,k})\right)+R_{i,1}^{y},\\
Z_{i}^{\pi,m} &=  \frac{1}{K_i}\sum\limits_{k=1}^{K_i} \sum\limits_{j=0}^{J}\beta_{J,j}Y_{i+j}^{\pi,m,k}\cdot(\Delta W_{i,i+j}^{m,k})^\top+R_{i,1}^{z},
\end{aligned}
\right.
\label{NS-MC}%
\end{equation}
by Monte Carlo method, where
given $X_i^{\pi,m}$, the branch values $Y_{i+1}^{\pi,m,k}$ are
independent copies with the conditional law of $Y_{i+1}^{\pi,m}$;
\begin{footnotesize}
\begin{equation}
\begin{aligned}
R_{i,1}^{y} =&   \mathbb{E}_{i}\Big[Y_{i+1}^{\pi,m}+\sum\limits_{j=0}^{J} \alpha_{j}hf(t_{i+j},X_{i+j}^{\pi,m},Y_{i+j}^{\pi,m},Z_{i+j}^{\pi,m})\Big]
- \frac{1}{K_i}\sum\limits_{k=1}^{K_i} \left(Y_{i+1}^{\pi,m,k}
+\sum\limits_{j=0}^{J} \alpha_{j}hf(t_{i+j},X_{i+j}^{\pi,m,k},Y_{i+j}^{\pi,m,k},Z_{i+j}^{\pi,m,k})\right),\\
R_{i,1}^{z} =& \sum\limits_{j=0}^{J}\beta_{J,j}\mathbb{E}_{i}\big[Y_{i+j}^{\pi,m}\cdot(\Delta W_{i,i+j}^m)^\top\big]
-  \frac{1}{K_i}\sum\limits_{k=1}^{K_i} \sum\limits_{j=0}^{J}\beta_{J,j}Y_{i+j}^{\pi,m,k}\cdot(\Delta W_{i,i+j}^{m,k})^\top.
\end{aligned}
\nonumber
\end{equation} 
\end{footnotesize}
Thus, the computational formula for $(Y_{i}^{\pi,m},Z_{i}^{\pi,m})$ is 
\begin{equation}
\left\{
\begin{aligned}
Y_{i,K_i}^{\pi,m}&= \frac{1}{K_i}\sum\limits_{k=1}^{K_i} \left(Y_{i+1}^{\pi,m,k}
+\sum\limits_{j=0}^{J} \alpha_{j}hf(t_{i+j},X_{i+j}^{\pi,m,k},Y_{i+j}^{\pi,m,k},Z_{i+j}^{\pi,m,k})\right),\\
Z_{i,K_i}^{\pi,m}&= \frac{1}{K_i}\sum\limits_{k=1}^{K_i} \sum\limits_{j=0}^{J}\beta_{J,j}Y_{i+j}^{\pi,m,k}\cdot(\Delta W_{i,i+j}^{m,k})^\top.
\end{aligned}
\right.
\label{NS-MC-1=hq}%
\end{equation}
We approximate the local Monte Carlo estimates in (\ref{NS-MC-1=hq}) using two deep feedforward neural networks: 
\begin{equation}
\left\{
\begin{aligned}
Y_{i,K_i}^{\pi,m}
= & \mathcal{NN}_{y,i,1+d, n}^{\varrho, L, \mathbf{n}} (p_i^m,\theta^0_{y,i})+ R_{i,2}^{y,0},\\
Z_{i,K_i}^{\pi,m}
= & \mathcal{NN}_{z,i,1+d, n\times d}^{\varrho, L, \mathbf{n}}(p^m_i,\theta^0_{z,i})+ R_{i,2}^{z,0},\\
\end{aligned}
\right.
\label{NS-MC-1}%
\end{equation}
where $p_i^m = (t_i,X_{i}^{\pi,m})$; $\theta^0_{y,i} $ and 
$\theta^0_{z,i} $ denote initial parameters of the two neural networks at each time step;
\begin{equation}
\begin{aligned}
R_{i,2}^{y,0}&= Y_{i,K_i}^{\pi,m}
-\mathcal{NN}_{y,i,1+d, n}^{\varrho, L, \mathbf{n}}(p_i^m,\theta^0_{y,i}),\\
 R_{i,2}^{z,0} &=Z_{i,K_i}^{\pi,m}
-\mathcal{NN}_{z,i,1+d, n\times d}^{\varrho, L, \mathbf{n}}(p_i^m,\theta^0_{z,i}).
\end{aligned}
\nonumber
\end{equation}
To train the deep feedforward neural networks, we introduce the expectation loss functions as below
\begin{equation}
\left\{
\begin{aligned}
F_{y,i}^{r_j}(\theta_{y,i}^{r_j})= & \frac{1}{M}\sum_{m=1}^{M} \left| S_{y,i}
 - \mathcal{NN}_{y,i,1+d, n}^{\varrho, L, \mathbf{n}}(p_i^m,\theta_{y,i}^{r_j})\right|^2,\\
F_{z,i}^{r_j}(\theta_{z,i}^{r_j})= & \frac{1}{M}\sum_{m=1}^{M} \left| S_{z,i}
 - \mathcal{NN}_{z,i,1+d, n\times d}^{\varrho, L, \mathbf{n}}(p_i^m,\theta_{z,i}^{r_j})
\right|^2,
\end{aligned}
\right.
\label{NS-MC-3}%
\end{equation}
where 
$p_{i}^{ m, k}= (t_i,X_{i}^{\pi,m,k})$;
$r_j\in \mathbb{N}^+$ denotes stochastic gradient descent (SGD) iteration index; 
$\theta^{r_j}_{y,i}  = \theta_{y,i}^{r_j-1}-\rho_{y,i}^{r_j-1}\nabla_{\theta_{y,i}^{r_j-1}}F_{y,i}^{r_j-1}(\theta_{y,i}^{r_j-1}),$
$\theta^{r_j}_{z,i}  = \theta_{z,i}^{r_j-1}-\rho_{z,i}^{r_j-1}\nabla_{\theta_{z,i}^{r_j-1}}F_{z,i}^{r_j-1}(\theta_{z,i}^{r_j-1})$;
$\rho_{y,i}^{r_j},\rho_{z,i}^{r_j}\in(0,+\infty)$ denote the learning rates;
$\theta^{*}_{y,i} = \arg \min\limits_{\theta_{y,i}^{r_j}} F_{y,i}^{r_j}(\theta_{y,i}^{r_j})$,
$\theta^{*}_{z,i} = \arg \min\limits_{\theta_{z,i}^{r_j}} F_{z,i}^{r_j}(\theta_{z,i}^{r_j})$.
Note that
for the calculation of the loss functions \( F_{y,i}^{r_j} \) and \( F_{z,i}^{r_j} \), terms containing future network outputs are handled as constant target values. To alleviate the risk of vanishing or exploding gradients caused by long-term dependencies, the gradients of these terms are truncated during backpropagation, which aligns with the DBR implementation strategy \cite{HQJSLY26};
\begin{align}
S_{z,i}
&=\frac{1}{K_i}\sum\limits_{k=1}^{K_i} \sum\limits_{j=0}^{J}\beta_{J,j}\mathcal{NN}_{y,i+j,1+d, n}^{\varrho, L, \mathbf{n}}(p_{i+j}^{m,k},\theta_{y,i+j}^*) \left(\Delta W_{i,i+j}^{m,k}\right)^\top,\nonumber\\
S_{y,i}
&= \frac{1}{K_i}\sum\limits_{k=1}^{K_i} \mathcal{NN}_{y,i+1,1+d, n}^{\varrho, L, \mathbf{n}}(p_{i+1}^{m,k},\theta_{y,i+1}^*)
+\alpha_0 hf(t_{i},X_{i}^{\pi,m},\mathcal{NN}_{y,i,1+d, n}^{\varrho, L, \mathbf{n}}(p_{i}^{m},\theta_{y,i}^{r_j}),\mathcal{NN}_{z,i,1+d, n\times d}^{\varrho, L, \mathbf{n}}(p_{i}^{m},\theta_{z,i}^*))\nonumber\\
&\quad+\frac{1}{K_i}\sum\limits_{k=1}^{K_i}\sum\limits_{j=1}^{J} \alpha_{j}hf(t_{i+j},X_{i+j}^{\pi,m,k},\mathcal{NN}_{y,i+j,1+d, n}^{\varrho, L, \mathbf{n}}(p_{i+j}^{m,k},\theta_{y,i+j}^{*}),\mathcal{NN}_{z,i+j,1+d, n\times d}^{\varrho, L, \mathbf{n}}(p_{i+j}^{m,k},\theta_{z,i+j}^*)).\nonumber
\end{align}
Thus, by the optimization of the SGD method, (\ref{NS-MC-1}) is rewritten as
\begin{equation}
\left\{
\begin{aligned}
Y_{i,K_i}^{\pi,m}
= &  \mathcal{NN}_{y,i,1+d, n}^{\varrho, L, \mathbf{n}}(p^{m}_{i},\theta^{*}_{y,i})+ R_{i,2}^{y} ,\\
Z_{i,K_i}^{\pi,m}
= &\mathcal{NN}_{z,i,1+d, n\times d}^{\varrho, L, \mathbf{n}}(p^{m}_{i},\theta^{*}_{z,i})+ R_{i,2}^{z},
\end{aligned}
\right.
\label{NS-MC-2-1}%
\end{equation}
with 
\begin{equation}
\begin{aligned}
R_{i,2}^{y}&= 
 \frac{1}{K_i}\sum\limits_{k=1}^{K_i} \left(Y_{i+1}^{\pi,m,k}
+\sum\limits_{j=0}^{J} \alpha_{j}hf(t_{i+j},X_{i+j}^{\pi,m,k},Y_{i+j}^{\pi,m,k},Z_{i+j}^{\pi,m,k})\right)- \mathcal{NN}_{y,i,1+d, n}^{\varrho, L, \mathbf{n}}(p_{i}^m,\theta_{y,i}^*),\\
R_{i,2}^{z}&= 
\frac{1}{K_i}\sum\limits_{k=1}^{K_i} \sum\limits_{j=0}^{J}\beta_{J,j}Y_{i+j}^{\pi,m,k}\cdot(\Delta W_{i,i+j}^{m,k})^\top
 - \mathcal{NN}_{z,i,1+d, n\times d}^{\varrho, L, \mathbf{n}}(p_{i}^m,\theta_{z,i}^*).
\end{aligned}
\nonumber
\end{equation} 

Now, we propose MHDBR method for the FBSDEs (\ref{dFBSDE}) as below:
\begin{enumerate}
\item 
Given $x_0$ generate the forward samples $X_{i}^{\pi,m}$ using (\ref{SDE-general-scheme}) for $i = 1,2,\cdots, N$, $m = 1,2,\cdots,M$.

\item 
Prescribe the starting values $(\mathcal{Y}_{N-j},{Z}_{N-j})$, $j =0,1,\cdots,J-1$. For $i = N-J,N-J-1,\cdots,0$, we solve $(\mathcal{Y}_{i},{Z}_{i})$ in the following manner:
\begin{equation}
\left\{
\begin{aligned}
\mathcal{Y}_{i}
= &  \mathcal{NN}_{y,i,1+d, n}^{\varrho, L, \mathbf{n}}(p^{m}_{i},\theta^{*}_{y,i}),\\
\mathcal{Z}_{i}
= &\mathcal{NN}_{z,i,1+d, n\times d}^{\varrho, L, \mathbf{n}}(p^{m}_{i},\theta^{*}_{z,i}).
\end{aligned}
\right.
\label{NS-DMC}%
\end{equation}
\end{enumerate}

\begin{remark}\label{remark-st}
The scheme (\ref{NS-DMC}) provides a method to calculate $(\mathcal{Y}_{i},\mathcal{Z}_{i})$, for
$i=N-J,N-J-1,\cdots,0$.
The starting values $(\mathcal{Y}_{N-j},\mathcal{Z}_{N-j}),$
for $j=0,1,\cdots,J-1$ have to be given or computed. To that end, some other numerical schemes, such as the deep BSDE method \cite{HJJAEW18}, the deep backward dynamic programming method \cite{HCPHWX20},
the deep splitting method \cite{BCBSCPJANA21} and 
the deep backward regression  method \cite{HQJSLY26} or the scheme (\ref{NS}) equipped with $J=1$ may be adopted with a smaller time step. 
The MHDBR algorithm is provided to compute the numerical solutions $\mathcal{Y}_{i}$ and $\mathcal{Z}_{i}$
(see \textbf{Algorithm 1}).
\end{remark}

\begingroup
\captionsetup{labelformat=empty,skip=3pt}

\begin{algorithm}[!htbp]
\footnotesize
\linespread{0.94}\selectfont
\SetAlgoLined
\DontPrintSemicolon
\SetAlCapSkip{3pt}

\caption{MHDBR method for FBSDEs}
\label{alg:MHDBR}

\KwIn{
Grid \(0=t_0<\cdots<t_N=T\), \(h=T/N\);
order \(J\);
coefficients \(\{\alpha_j\}_{j=0}^{J}\) and
\(\{\beta_{J,j}\}_{j=0}^{J}\);
sample sizes \(M\) and \(K_i\);
starting values
\(\{(\mathcal{Y}_{N-j},\mathcal{Z}_{N-j})\}_{j=0}^{J-1}\).
}

\KwOut{
Approximations
\(\{(\mathcal{Y}_i,\mathcal{Z}_i)\}_{i=0}^{N-J}\).
}

\BlankLine
Generate the forward samples
\(\{X_i^{\pi,m}\}_{i=0}^{N}\), \(m=1,\ldots,M\), by
\[
X_{i+1}^{\pi,m}
=
X_i^{\pi,m}
+
\Psi(t_i,X_i^{\pi,m},h,
\Delta W_{i,i+1}^{m},\xi_{i,i+1}^{m}),
\qquad
X_0^{\pi,m}=x_0.
\]

Set the prescribed starting values
\[
(\mathcal{Y}_{N-j},\mathcal{Z}_{N-j}),
\qquad j=0,\ldots,J-1.
\]

\For{\(i=N-J,N-J-1,\ldots,0\)}{

Initialize \(\theta_{z,i}\) and \(\theta_{y,i}\)\;

\tcp{Training of the \(Z_i\)-network}

\For{each optimization iteration}{

For every \(X_i^{\pi,m}\), generate \(K_i\) conditional samples
\[
\bigl\{
X_{i+j}^{\pi,m,k},
\Delta W_{i,i+j}^{m,k}
\bigr\}_{j=1}^{J},
\qquad k=1,\ldots,K_i.
\]

Evaluate the future quantities
\(\mathcal{Y}_{i+j}^{m,k}\) using the prescribed starting values or
the previously trained networks, and compute
\[
S_{z,i}^{m}
=
\frac{1}{K_i}
\sum_{k=1}^{K_i}\sum_{j=1}^{J}
\beta_{J,j}
\mathcal{Y}_{i+j}^{m,k}
\left(\Delta W_{i,i+j}^{m,k}\right)^\top.
\]

Compute
\[
F_{z,i}(\theta_{z,i})
=
\frac{1}{M}\sum_{m=1}^{M}
\left\|
S_{z,i}^{m}
-
\mathcal{NN}_{z,i,1+d,n\times d}^{\varrho,L,\mathbf n}
\bigl(p_i^{m},\theta_{z,i}\bigr)
\right\|^2.
\]

Update \(\theta_{z,i}\) using a gradient-based optimizer\;
}

Denote the optimized parameter by \(\theta_{z,i}^{*}\) and set
\[
\mathcal{Z}_{i}^{m}
=
\mathcal{NN}_{z,i,1+d,n\times d}^{\varrho,L,\mathbf n}
\bigl(p_i^{m},\theta_{z,i}^{*}\bigr).
\]

\tcp{Training of the \(Y_i\)-network}

\For{each optimization iteration}{

Generate a new family of conditional samples and evaluate
\(\mathcal{Y}_{i+j}^{m,k}\) and
\(\mathcal{Z}_{i+j}^{m,k}\), \(j=1,\ldots,J\), using the prescribed
starting values or the previously trained networks\;

For \(m=1,\ldots,M\), compute
\[
\begin{aligned}
S_{y,i}^{m}
={}&
\frac{1}{K_i}\sum_{k=1}^{K_i}
\left[
\mathcal{Y}_{i+1}^{m,k}
+
h\sum_{j=1}^{J}\alpha_j
f\left(
t_{i+j},X_{i+j}^{\pi,m,k},
\mathcal{Y}_{i+j}^{m,k},
\mathcal{Z}_{i+j}^{m,k}
\right)
\right]
\\
&+
\alpha_0h
f\left(
t_i,X_i^{\pi,m},
\mathcal{NN}_{y,i,1+d,n}^{\varrho,L,\mathbf n}
\bigl(p_i^{m},\theta_{y,i}\bigr),
\mathcal{Z}_i^{m}
\right).
\end{aligned}
\]

Compute
\[
F_{y,i}(\theta_{y,i})
=
\frac{1}{M}\sum_{m=1}^{M}
\left|
S_{y,i}^{m}
-
\mathcal{NN}_{y,i,1+d,n}^{\varrho,L,\mathbf n}
\bigl(p_i^{m},\theta_{y,i}\bigr)
\right|^2.
\]

Update \(\theta_{y,i}\) using a gradient-based optimizer while keeping
\(\theta_{z,i}^{*}\) and all future network parameters fixed\;
}

Denote the optimized parameter by \(\theta_{y,i}^{*}\) and set
\[
\mathcal{Y}_{i}^{m}
=
\mathcal{NN}_{y,i,1+d,n}^{\varrho,L,\mathbf n}
\bigl(p_i^{m},\theta_{y,i}^{*}\bigr).
\]
}

\BlankLine
\textbf{Final estimation:}
\[
\mathcal{Y}_0
=
\mathcal{NN}_{y,0,1+d,n}^{\varrho,L,\mathbf n}
\bigl(p_0^{m},\theta_{y,0}^{*}\bigr).
\]

\end{algorithm}
\endgroup

\subsection{Error analysis}

We now proceed to bound the error between the numerical  solutions $(\mathcal{Y}_i,\mathcal{Z}_i)_{0\le i\le N-1}$
and the exact solutions  $(Y_{t_i},Z_{t_i})_{0\le i\le N-1}$.
This analysis yields an error estimate for the MHDBR method (\ref{NS-DMC}) in $L^2$-norm.
We state the following conditions with respect to the FBSDEs (\ref{dFBSDE}) associated with the semilinear PDEs (\ref{PDE}).
\textbf{Assumption 1.}
\begin{description}
\item[(i)]
The initial state $x_0\in\mathbb R^d$ is fixed and deterministic throughout this paper.
\item[(ii)]
The coefficients $\mu$ and $\sigma$ are Lipschitz continuous with respect to the spatial variable $x$;
assume
\[
\sup\limits_{0\le t\le T}\{|\mu(t,0)|+|\sigma(t,0)|\}\leq L,
\]
where  $L>0$  denotes the Lipschitz constant.
\item[(iii)]
The generator $f$ is $\frac{1}{2}-$H\"{o}lder continuous with respect to the time variable $t$ and
uniformly Lipschitz continuous in all other variables
$$
|f(t_1,x_1,y_1,z_{1})-f(t_2,x_2,y_2,z_{2})|\le
L(|t_{1}-t_{2}|^{\frac{1}{2}}+|x_{1}-x_{2}|+|y_{1}-y_{2}|+|z_{1}-z_{2}|),
$$
for all $t\in[0,T],$
$x_{1},x_{2}\in \mathbb{R}^d$, $y_{1},y_{2}\in \mathbb{R}^n$ and $z_{1},z_{2}\in\mathbb{R}^{n\times d}$;
and suppose
\[
\sup\limits_{0\le t\le T}\{|f(t,0,0,0)|\}\leq L.
\]
\item[(iv)]
The terminal function $g$ has at most linear growth.
\end{description}

\textbf{Assumption 2.}
Let $\mathcal L^0,\ldots,\mathcal L^d$ be the differential
operators defined in Appendix~A.
For a finite multi-index
$\gamma=(\gamma_1,\ldots,\gamma_p)$, with
$\gamma_j\in\{0,1,\ldots,d\}$, define
$
\ell(\gamma)=p,\quad
\mathcal L^\gamma
=
\mathcal L^{\gamma_1}\cdots\mathcal L^{\gamma_p}.
$
For the empty multi-index $\emptyset$, set
$\ell(\emptyset)=0$ and
$\mathcal L^\emptyset\nu=\nu$.
For an integer $k\geq0$, define
\[
\mathcal A_b^k
:=
\left\{
\begin{array}{l|l}
\nu:[0,T]\times\mathbb R^d\to\mathbb R^n
&
\begin{array}{l}
\mathcal L^\gamma\nu
\text{ exists, is continuous, and is bounded}\\
\text{for every multi-index }\gamma
\text{ with }\ell(\gamma)\leq k
\end{array}
\end{array}
\right\}.
\]

Define
\[
F_u(t,x)
:=
f\bigl(t,x,u(t,x),D_xu(t,x)\sigma(t,x)\bigr).
\]
For the prescribed integer $J\geq1$, assume that
\[
u\in\mathcal A_b^{J+2},
\qquad
F_u\in\mathcal A_b^{J+1}.
\]

\textbf{Assumption 3.}
Let $X_i^{\pi,m}$ be generated by the one-step transition
\eqref{SDE-M}, using the same Brownian motion as the exact SDE.
This transition may be an It\^{o}--Taylor approximation or an exact
transition when available. Assume that it satisfies
\begin{align}
&\max\limits_{0\le i\le N}\mathbb{E}\left[ 
\left|X_i^{\pi,m}\right|^p\right]
\le C_p(1+|x_0|^p),\quad p\ge 1,\label{X-err}\\
& \max_{0\leq i\leq N-1}
 \mathbb E\left|X_{t_{i+1}}- X_{i+1}^{\pi,m}\right|^2 \le C h^{2 \gamma_1},\label{phi-X-err}
\end{align}
where $p$, $\gamma_1$,  $C_p$, and $C$ 
are positive constants.
For $0\leq i\leq N-J$, define the conditional forward
process starting from $x$ at $t_i$ by
\begin{equation}\label{SDE-general-scheme-ix}
\begin{aligned}
X_i^{i,x}&=x,\\
X_{i+q+1}^{i,x}
&=X_{i+q}^{i,x}
+\Psi\bigl(
t_{i+q},X_{i+q}^{i,x},h,
\Delta W_{i+q,i+q+1},
\xi_{i+q,i+q+1}
\bigr),
\qquad q=0,\ldots,J-1.
\end{aligned}
\end{equation}
The transition rule and the distribution of the driving
variables are the same as those used to generate $X^\pi$.
Branches starting from $x,x'$ and using the same driving
variables satisfy
\begin{equation}\label{GS-X-stability}
 \mathbb E|X_{i+j}^{i,x}-X_{i+j}^{i,x'}|^2
 \leq(1+Ch)|x-x'|^2,
 \qquad 1\leq j\leq J,
\end{equation}
and have finite moments of the orders used below.
The constant in \eqref{GS-X-stability} is uniform in $i,N$ for
fixed $J$.

\begin{remark}
Under the corresponding coefficient smoothness, growth and stability
conditions, standard strong-error results give
$\gamma_1=1/2$ for Euler--Maruyama, $\gamma_1=1$ for Milstein,
and $\gamma_1=2$ for a strong order-two It\^{o}--Taylor scheme;
see \cite{KPEPE92}. These rates are not automatic consequences of
Assumption 1. The moment and branch-stability bounds in Assumption 3
must also hold, and any numerical approximation of iterated stochastic
integrals must preserve the asserted strong order.
\end{remark}


For the analysis, all starting approximations are deterministic
measurable functions of the current state. In particular, for
$q=N-J+1,\ldots,N$, write
$\mathcal Y_q=y_q^{\rm st}(X_q^{\pi,m})$ and
$\mathcal Z_q=z_q^{\rm st}(X_q^{\pi,m})$, and impose the same
Markov form on $(Y_q^{\pi,m},Z_q^{\pi,m})$.
Their required accuracy is specified in \eqref{starting-errors}.
Future network functions are kept fixed at each backward step.

Now, we investigate the errors of the scheme (\ref{NS-DMC}) and define, for $i=N-J, N-J-1,\cdots,0$
\begin{equation}
\left\{\begin{aligned}
\mathcal{U}_{i}^m & :=\mathbb{E}_i\left[\mathcal{Y}_{i+1}
+\alpha_0 hf(t_{i},X_{i}^{\pi,m},\mathcal{U}_{i}^m ,\mathcal{V}_{i}^m ) 
+\sum\limits_{j=1}^{J} \alpha_{j}hf\left(t_{i+j}, X_{i+j}^{\pi,m}, \mathcal{Y}_{i+j}, \mathcal{Z}_{i+j} \right)\right],\\
\mathcal{V}_{i}^m  & :=\mathbb{E}_i\left[\sum\limits_{j=0}^{J}\beta_{J,j}\mathcal{Y}_{i+j}\cdot(\Delta W_{i,i+j}^m)^\top\right].
\end{aligned}\right.
\label{Deep-NN-E}
\end{equation}
and by the Markov property of the discretized forward process
$\{X_i^{\pi,m}\}_{0\le i\le N}$, we have
\begin{equation}
\begin{aligned}
\mathcal{U}_{i}^m = \widetilde{u}(t_i,X^{\pi,m}_i),
\quad
\mathcal{V}_{i}^m = \widetilde{v}(t_i,X^{\pi,m}_i),
\end{aligned}
\label{Deep-NN-E--YZK-1}
\end{equation}
where $\widetilde{u}$ and $\widetilde{v}$ are some deterministic (but unknown) functions;
\begin{equation}
\left\{\begin{aligned}
\widetilde{\mathcal{U}}_{i}^m & :=\frac{1}{K_i}\sum\limits_{k=1}^{K_i}\mathcal{NN}_{y,i+1,1+d, n}^{\varrho, L, \mathbf{n}}(p_{i+1}^{m,k},\theta^{*}_{y,i+1})
+\alpha_0hf\left(t_i, X_{i}^{\pi,m}, \widetilde{\mathcal{U}}_{i}^m, \widetilde{\mathcal{V}}_{i}^m\right) \\
& +\frac{h}{K_i}\sum\limits_{k=1}^{K_i}\sum\limits_{j=1}^{J} \alpha_{j}f\left(t_{i+j}, X_{i+j}^{\pi,m,k}, \mathcal{NN}_{y,i+j,1+d, n}^{\varrho, L, \mathbf{n}}(p_{i+j}^{m,k},\theta^{*}_{y,i+j}), \mathcal{NN}_{z,i+j,1+d, n\times d}^{\varrho, L, \mathbf{n}}(p_{i+j}^{m,k},\theta^{*}_{z,i+j})\right),\\
\widetilde{\mathcal{V}}_{i}^m  & :=\frac{1}{K_i}\sum\limits_{k=1}^{K_i}
\sum\limits_{j=0}^{J}\beta_{J,j} \mathcal{NN}_{y,i+j,1+d, n}^{\varrho, L, \mathbf{n}}(p_{i+j}^{m,k},\theta^{*}_{y,i+j})\cdot(\Delta W_{i,i+j}^{m,k})^\top.
\end{aligned}\right.
\label{Deep-NN-E--YZK}
\end{equation}
Let us introduce
\begin{equation}
\begin{aligned}
&\varepsilon_{i}^{y} = \inf\limits_{\mathcal{NN}_{y,i,1+d, n}^{\varrho, L, \mathbf{n}}}\mathbb{E}\left[\left|\widetilde{u}(t_i,X^{\pi,m}_i)-\mathcal{NN}_{y,i,1+d, n}^{\varrho, L, \mathbf{n}}(p_{i}^{m},\theta^{r_j}_{y,i})\right|^2\right],\\
&\varepsilon_{i}^{z} = \inf\limits_{\mathcal{NN}_{z,i,1+d, n\times d}^{\varrho, L, \mathbf{n}}}\mathbb{E}\left[\left|\widetilde{v}(t_i,X^{\pi,m}_i)-\mathcal{NN}_{z,i,1+d, n\times d}^{\varrho, L, \mathbf{n}}(p_{i}^{m},\theta^{r_j}_{z,i})\right|^2\right],
\end{aligned}
\label{Deep-NN-E--YZK-21}
\end{equation}
and
\begin{equation}
\begin{aligned}
&\widetilde{\varepsilon}_{i}^{y} = \inf\limits_{\mathcal{NN}_{y,i,1+d, n}^{\varrho, L, \mathbf{n}}}\mathbb{E}\left[\left|\widetilde{\mathcal{U}}_i^{m}-\mathcal{NN}_{y,i,1+d, n}^{\varrho, L, \mathbf{n}}(p_{i}^{m},\theta^{r_j}_{y,i})\right|^2\right],\\
&\widetilde{\varepsilon}_{i}^{z} = \inf\limits_{\mathcal{NN}_{z,i,1+d, n\times d}^{\varrho, L, \mathbf{n}}}\mathbb{E}\left[\left|\widetilde{\mathcal{V}}_i^{m}-\mathcal{NN}_{z,i,1+d, n\times d}^{\varrho, L, \mathbf{n}}(p_{i}^{m},\theta^{r_j}_{z,i})\right|^2\right].
\end{aligned}
\label{Deep-NN-E--YZK-2}
\end{equation}

By a slight abuse of notation, we henceforth write $\widetilde{\mathcal{V}}_{i}^m$
to denote $S_{z,i}$,
to shift the focus from the specific functional form of the network to its role as an approximated function.

\noindent\textbf{Population loss convention.}
The loss functions in \eqref{NS-MC-3} constitute empirical objectives employed for the practical implementation of \textbf{Algorithm 1}. For the subsequent error analysis, however, we operate at the population level. Assume that each population infimum over the stated network class
is attained, and that the class contains the zero function. At each
backward step choose a deterministic exact minimizer, keeping all
future network functions fixed. Expectations are over an independent
evaluation forward path and its fresh conditional Monte Carlo branches.
Existence of these minimizers is a hypothesis, not a consequence of
universal approximation or of the practical optimization algorithm.
For fixed \(i\) and \(r_j\), we define the population loss associated with the \(Z\)-component as
\begin{equation}
\begin{aligned}
\mathcal{F}_{z,i}^{r_j}(\theta_{z,i}^{r_j})
:=&
\mathbb{E}\left[
\left|
\widetilde{\mathcal{V}}_{i}^m
-
\mathcal{NN}_{z,i,1+d,n\times d}^{\varrho,L,\mathbf{n}}
(p_i^m,\theta_{z,i}^{r_j})
\right|^2
\right].
\end{aligned}
\label{POP-LOSS-Z}
\end{equation}
Subsequently, upon denoting by \(\theta_{z,i}^{*}\) the exact minimizer of \eqref{POP-LOSS-Z}, we define the population loss associated with the \(Y\)-component as
\begin{equation}
\begin{aligned}
\mathcal{F}_{y,i}^{r_j}(\theta_{y,i}^{r_j})
:=&
\mathbb{E}\Bigg[
\bigg|
\widetilde{\mathcal{U}}_{i}^m
+\alpha_0h\Big(
f\big(t_i,X_i^{\pi,m},
\mathcal{NN}_{y,i,1+d,n}^{\varrho,L,\mathbf{n}}
(p_i^m,\theta_{y,i}^{r_j}),
\mathcal{NN}_{z,i,1+d,n\times d}^{\varrho,L,\mathbf{n}}
(p_i^m,\theta_{z,i}^{*})\big)
\\
&\qquad\qquad
-
f\big(t_i,X_i^{\pi,m},
\widetilde{\mathcal{U}}_{i}^m,
\widetilde{\mathcal{V}}_{i}^m\big)
\Big)
-
\mathcal{NN}_{y,i,1+d,n}^{\varrho,L,\mathbf{n}}
(p_i^m,\theta_{y,i}^{r_j})
\bigg|^2
\Bigg].
\end{aligned}
\label{POP-LOSS-Y}
\end{equation}

Throughout the proof of Theorem \ref{MHDBR-err-esti} , the symbols \(F_{y,i}^{r_j}\) and \(F_{z,i}^{r_j}\) are to be understood exclusively as the population losses \(\mathcal{F}_{y,i}^{r_j}\) and \(\mathcal{F}_{z,i}^{r_j}\), respectively. Given that the empirical losses in \eqref{NS-MC-3} serve as Monte Carlo estimators of these population objectives, Theorem \ref{MHDBR-err-esti}  constitutes an approximation-error estimate predicated on exact population minimization. Discrepancies arising from (i) the substitution of population losses with their empirical counterparts (generalization error), and (ii) non-exact stochastic optimization (optimization error), are decoupled from the present theorem; bounds for these additional errors require a separate analysis.

We now analyze the error of the MHDBR method.
To facilitate the proof of Theorem~\ref{MHDBR-err-esti},
we first establish the following lemma.

\begin{lemma}\label{noD0-stability}
Let $e_0^y,\ldots,e_N^y$ be adapted,
square-integrable $\mathbb R^n$-valued variables and let
$e_0^z,\ldots,e_N^z$ be adapted, square-integrable
$\mathbb R^{n\times d}$-valued variables. Define
$b_i=e_i^y-\mathbb E_i e_{i+1}^y$, 
$\mathfrak m_{i+1}=e_{i+1}^y-\mathbb E_i e_{i+1}^y$, $0\leq i<N$ .
Suppose that $\bar\beta_j=h\beta_{J,j}$, $1\leq j\leq J$,
are fixed real numbers independent of $N,h$, and that
\begin{equation}\label{noD0-z-relation}
 e_i^z=\sum_{j=1}^J\beta_{J,j}
        \mathbb E_i[e_{i+j}^y(\Delta W_{i,i+j})^\top]+\rho_i^z,
\end{equation}
where  $n_*=N-J$; $\rho_i^z$ is an $\mathcal F_{t_i}$-measurable,
square-integrable $\mathbb R^{n\times d}$-valued residual term.
Assume that nonnegative deterministic numbers $\mathfrak r_i$ and a
constant $C_b\geq1$, independent of $N,h$, satisfy, for 
$0\leq k\leq n_*$,
\begin{equation}\label{noD0-drift-bound}
 \sum_{i=k}^{n_*}\mathbb E|b_i|^2
 \leq C_bh^2\sum_{i=k}^{N}\mathbb E(|e_i^y|^2+|e_i^z|^2)
                      +C_b\sum_{i=k}^{n_*}\mathfrak r_i .
\end{equation}
Then, for sufficiently small $h$, depending only on the fixed
coefficients, $C_b,d,J,T$,
\begin{align}
 \max_{0\leq i\leq n_*}\mathbb E|e_i^y|^2
                    +h\sum_{i=0}^{n_*}\mathbb E|e_i^z|^2
\leq C\left\{
       \sum_{i=n_*+1}^{N}\mathbb E(|e_i^y|^2+h|e_i^z|^2)
       +\frac1h\sum_{i=0}^{n_*}\mathfrak r_i
                    +h\sum_{i=0}^{n_*}\mathbb E|\rho_i^z|^2
                      \right\}.
 \label{noD0-stability-bound}
\end{align}
\end{lemma}

\begin{proof}
Let
\begin{align}
 a_i=\mathbb E|e_i^y|^2,\quad c_i=\mathbb E|e_i^z|^2,\quad
 v_i=\mathbb E|\mathfrak m_{i+1}|^2,\quad
 \Xi=\sum_{i=n_*+1}^N(a_i+hc_i).
 \label{def-acv}
\end{align}
For $0\leq k\leq n_*$, also let
\[
 V_k=\sum_{i=k}^{n_*}v_i,\quad
 Q_k=h\sum_{i=k}^{n_*}c_i,\quad
 B_k=\sum_{i=k}^{n_*}\mathbb E|b_i|^2,\quad
 \mathcal R_k^y=\sum_{i=k}^{n_*}\mathfrak r_i,\quad
 \mathcal R_k^z=h\sum_{i=k}^{n_*}\mathbb E|\rho_i^z|^2.
\]
The definitions imply the exact identity
\[
 e_{i+j}^y=e_i^y+\sum_{r=i}^{i+j-1}(\mathfrak m_{r+1}-b_r).
\]
The variables $\mathfrak m_{r+1}$ are martingale differences.
For $i\leq r<s$, conditional orthogonality gives
\[
 \mathbb E_i\langle \mathfrak m_{r+1}, \mathfrak m_{s+1}\rangle=0.
\]
By martingale orthogonality and the Cauchy--Schwarz inequality,
we obtain
\begin{align}
 \mathbb E\{\mathbb D_i[e_{i+j}^y]\}
 &\leq\mathbb E|e_{i+j}^y-e_i^y|^2\notag\\
 &\leq2\sum_{r=i}^{i+j-1}v_r
             +2j\sum_{r=i}^{i+j-1}\mathbb E|b_r|^2 .
 \label{noD0-multistep-variance}
\end{align}
where $\mathbb D_i[\cdot]=\mathbb D[\cdot|\mathcal F_{t_i}]$ is the conditional variance.
Applying
$\mathbb E_i[\Delta W_{i,i+j}]=0$,
$\mathbb E_i[|\Delta W_{i,i+j}|^2]=djh$ and
the Cauchy--Schwarz inequality to
\eqref{noD0-z-relation}, one obtains
\[
 h c_i\leq
 2dJ\sum_{j=1}^J j|\bar\beta_j|^2
                    \mathbb E\{\mathbb D_i[e_{i+j}^y]\}
                   +2h\mathbb E|\rho_i^z|^2.
\]
For fixed $j$, each index $r$ occurs in the intervals
$i,\ldots,i+j-1$ at most $j$ times. Therefore
\begin{align}
 Q_k
 &\leq C_\beta\sum_{r=k}^{N-1}
                   \big(v_r+\mathbb E|b_r|^2\big)+2\mathcal R_k^z,
 \label{noD0-z-overlap}
\end{align}
where 
$C_\beta=1+4dJ\sum_{j=1}^Jj^3|\bar\beta_j|^2$.
For $n_*<r<N$, \eqref{def-acv} gives,
\[
 v_r\leq a_{r+1},\qquad
 \mathbb E|b_r|^2\leq2a_r+2a_{r+1}.
\]
Inserting the above two inequalities into \eqref{noD0-z-overlap} leads to
\begin{equation}\label{noD0-z-before-absorption}
 Q_k\leq C_0(V_k+B_k+\Xi)+2\mathcal R_k^z,
\end{equation}
with a constant $C_0$ independent of $h$, $k$.
By \eqref{noD0-drift-bound} and $h\leq1$, we have
\begin{equation}\label{noD0-B-tail}
 B_k\leq C_bh^2\sum_{i=k}^{n_*}a_i
                      +C_bh Q_k+C_bh\Xi+C_b\mathcal R_k^y.
\end{equation}
Choose $h$ such that $C_0C_bh\leq1/2$.
Then inserting \eqref{noD0-B-tail} into \eqref{noD0-z-before-absorption} and absorbing $Q_k$, we obtain
\begin{equation}\label{noD0-Q-tail}
 Q_k\leq C_*\left(
 V_k+h^2\sum_{i=k}^{n_*}a_i+\Xi+\mathcal R_k^y+\mathcal R_k^z
                      \right),
\end{equation}
where $C_*$ is independent of $h$.
Since $e_i^y-b_i$ is $\mathcal F_{t_i}$-measurable and
$\mathbb E_i \mathfrak m_{i+1}=0$, expanding
$e_{i+1}^y=e_i^y-b_i+\mathfrak m_{i+1}$ gives
\[
 a_{i+1}
 =
 \mathbb E|e_i^y-b_i|^2+v_i
 =
 a_i-2\mathbb E\langle e_i^y,b_i\rangle
 +\mathbb E|b_i|^2+v_i,
\]
which implies
\[
 a_i+v_i
 =a_{i+1}+2\mathbb E\langle e_i^y,b_i\rangle
                                      -\mathbb E|b_i|^2.
\]
Applying Young's inequality with $\gamma\geq1$, we obtain 
\[
 a_i+v_i\leq a_{i+1}+\gamma h a_i
                              +\frac1{\gamma h}\mathbb E|b_i|^2.
\]
By summing the above inequality over $i=k,\ldots,n_*$
and using \eqref{noD0-B-tail}, we obtain
\begin{align}
 a_k+V_k
 &\leq \Xi+\gamma h\sum_{i=k}^{n_*}a_i
                                      +\frac1{\gamma h}B_k\notag\\
 &\leq C\Xi+
       \left(\gamma+\frac{C_b}{\gamma}\right)
                          h\sum_{i=k}^{n_*}a_i
       +\frac{C_b}{\gamma}Q_k
                          +\frac{C_b}{\gamma h}\mathcal R_k^y .
 \label{noD0-Y-tail}
\end{align}
Choose $\gamma\geq\max\{1,4C_bC_*\}$. Inserting \eqref{noD0-Q-tail} into
\eqref{noD0-Y-tail} and absorbing the $V_k$, it is easy to get
\[
 a_k+\tfrac34V_k
 \leq C_\gamma h\sum_{i=k}^{n_*}a_i
       +C\left(\Xi+\frac{\mathcal R_k^y}{h}+\mathcal R_k^z\right).
\]
Taking $h$ smaller so that $C_\gamma h\leq1/2$
and absorbing $a_k$, we obtain
\begin{equation}\label{noD0-final-tail}
 a_k+\tfrac12 V_k
 \leq Ch\sum_{i=k+1}^{n_*}a_i
       +C\left(\Xi+\frac{\mathcal R_k^y}{h}+\mathcal R_k^z\right).
\end{equation}

\medskip
Since $\mathcal R_k^y\leq\mathcal R_0^y$ and
$\mathcal R_k^z\leq\mathcal R_0^z$,
applying the discrete Gronwall inequality to
\eqref{noD0-final-tail}, we obtain
\[
 \max_{0\leq i\leq n_*}a_i
       \leq C\left(\Xi+\frac{\mathcal R_0^y}{h}+\mathcal R_0^z\right).
\]
Taking $k=0$ in \eqref{noD0-final-tail} gives the same
bound for $V_0$. Finally, applying \eqref{noD0-Q-tail}, together with $h^2\sum_i a_i\leq hT\max_i a_i$, $\mathcal R_0^y\leq\mathcal R_0^y/h$, gives the same bound for $Q_0$.
This proves \eqref{noD0-stability-bound}.
\end{proof}

\begin{theorem} \label{MHDBR-err-esti}
Suppose that \textbf{Assumptions 1--3} hold. Assume the Markov form
of the starting values and the exact population minimization specified above.
Let $(Y_{t_i},Z_{t_i})$
and $(\mathcal{Y}_{i},\mathcal{Z}_{i})$ be solutions of the FBSDEs  (\ref{dFBSDE}) and solutions of the MHDBR method
(\ref{NS-DMC}) respectively.
Suppose also that the starting values
satisfy
\begin{align}
\sum_{i=N-J+1}^N
       \mathbb E\big[
        |Y_{t_i}-Y_i^{\pi,m}|^2+h|Z_{t_i}-Z_i^{\pi,m}|^2+ |Y_i^{\pi,m}-\mathcal Y_i|^2
                              +h|Z_i^{\pi,m}-\mathcal Z_i|^2\big]
\le Ch^{2J}.
\label{starting-errors}
\end{align}
Assume further that, for some $\delta>0$,
\begin{equation}\label{MC-network-moment}
 \max_{0\leq q\leq N}\mathbb E|\mathcal Y_q|^{2+\delta}\leq C,
\end{equation}
where $C$ is uniform in $N$ and $K_i$ for the selected minimizers.
This is an additional stability hypothesis on the selected network
outputs; it is not asserted for arbitrary network parameters.
Then, for sufficiently small $h$
\begin{equation}
\begin{aligned}
&\max\limits_{0\le i\le N-J}\mathbb{E}\left[\left|Y_{t_i}-\mathcal{Y}_{i}\right|^2\right]
+h\sum\limits_{i=0}^{N-J}\mathbb{E}\left[|Z_{t_i}-\mathcal{Z}_{i}|^2 \right]
\le   Ch^{2\min\{J, \gamma_1\}}
+C\sum\limits_{i=0}^{N-J}\mathbb{E}\left[\frac{\varepsilon_{i}^{y}}{h}
+h\varepsilon_{i}^{z}
+\frac{1}{hK}\right],
\nonumber
\end{aligned}
\end{equation}
where $K=\min\limits_{0\le i\le N-J}\{K_i\}$; 
$C$ represents a positive constant independent of $\pi$, which may depend on $d$, $J$, $T$ and change from line to line.
\end{theorem}
\begin{proof}
The proof consists of three steps. First, we estimate the time-discretization error. Second, we estimate the neural-network approximation and conditional Monte Carlo sampling errors. Finally, we combine these estimates to bound the total error. We begin with the decomposition
\begin{equation}
\begin{aligned}
&\max\limits_{0\le i\le N-J}\mathbb{E}\left[\left|Y_{t_i}-\mathcal{Y}_{i}\right|^2\right]
+\sum\limits_{i=0}^{N-J}h\mathbb{E}\left[|Z_{t_i}-\mathcal{Z}_{i}|^2 \right]\\
&\le  2\max\limits_{0\le i\le N-J}\mathbb{E}\left[\left|Y_{t_i}-Y_i^{\pi,m}\right|^2\right]
+2 \sum_{i=0}^{N-J}h\mathbb{E}\left[\left| Z_{t_i}-Z_{i}^{\pi,m}\right|^2\right]\\
&\quad+2\max\limits_{0\le i\le N-J}\mathbb{E}\left[\left|Y_i^{\pi,m}-\mathcal{Y}_{i} \right|^2\right]
+2 \sum_{i=0}^{N-J}h\mathbb{E}\left[\left|Z_i^{\pi,m}-\mathcal{Z}_i\right|^2\right].
\label{MHDBR-err-esti-1}
\end{aligned}
\end{equation}
We now estimate the two terms on  the right hand side  of  (\ref{MHDBR-err-esti-1}).

Step 1. \underline{Estimates for  $\mathbb{E}\left[\left|Y_{t_i}-Y_i^{\pi,m}\right|^2\right]+ \sum\limits_{i=0}^{N-J}\mathbb{E}\left[h\left| Z_{t_i}-Z_{i}^{\pi,m}\right|^2\right]$}. 
We first discretize the BSDE in (\ref{dFBSDE}) using the multistep method:
\begin{equation}
\left\{
\begin{aligned}
Y_{t_i}= & \mathbb{E}_{i}\Big[Y_{t_{i+1}}
+ \sum\limits_{j=0}^{J} \alpha_{j}hf_{t_{i+j}} \Big] + R^y_{i,3},\\
Z_{t_i} = & \sum_{j=0}^{J}\beta_{J,j}\mathbb{E}_{i}\big[Y_{t_{i+j}}\Delta W_{i,i+j}^\top\big] + R^z_{i,3},
\end{aligned}
\right.
\label{multistep-dis-time-NS}%
\end{equation}
where $f_{t_{i+j}}=f(t_{i+j},X_{t_{i+j}},Y_{t_{i+j}},Z_{t_{i+j}})$ and
\begin{align*}
 R^y_{i,3}&=\int_{t_i}^{t_{i+1}}\mathbb E_i[f_s]\,ds
             -h\sum_{j=0}^J\alpha_j\mathbb E_i[f_{t_{i+j}}],\\
 R^z_{i,3}&=\left.\frac{d}{dt}
       \mathbb E_i[Y_t(W_t-W_{t_i})^\top]\right|_{t=t_i}
       -\sum_{j=0}^J\beta_{J,j}
                  \mathbb E_i[Y_{t_{i+j}}\Delta W_{i,i+j}^\top].
\end{align*}
For completeness, the local consistency estimates follow from the
coefficient identities
\[
 \sum_{j=0}^J\alpha_jj^q=\frac1{q+1}\quad(0\leq q\leq J),
 \qquad
 \sum_{j=0}^J(h\beta_{J,j})j^q=\mathbf 1_{\{q=1\}}
 \quad(0\leq q\leq J).
\]
By Assumption 2 and repeated It\^{o}'s formula,
$s\mapsto\mathbb E_iF_u(t_i+s,X_{t_i+s})$ has bounded
conditional time derivatives through order $J+1$. Taylor expansion
and the first identities therefore give $|R_{i,3}^y|\leq Ch^{J+2}$.
For a Brownian coordinate $r$, apply the same argument on the
augmented state $(X_t,W_t-W_{t_i})$ to $u(t,x)w_r$.
For the space-time It\^{o} generator $\widehat{\mathcal L}$,
$\widehat{\mathcal L}(a w_r)=w_r\mathcal L^0a+\mathcal L^r a$.
Induction gives
\[
 \widehat{\mathcal L}^{\,m}(u w_r)
 =w_r(\mathcal L^0)^m u
 +\sum_{q=0}^{m-1}(\mathcal L^0)^q\mathcal L^r
                         (\mathcal L^0)^{m-1-q}u.
\]
For $m\leq J+1$, their conditional expectations are bounded on
$0\leq s\leq Jh$ by Assumption 2 and the Brownian increment moments.
Also the first derivative at $s=0$ is
$\mathcal L^r u(t_i,X_{t_i})=Z_{t_i}^{(r)}$.
Taylor expansion and the second coefficient identities yield
$|R_{i,3}^z|\leq Ch^J$. Consequently,
\begin{equation}\label{MHDBR-err-esti-2-6}
 |R_{i,3}^y|^2\leq Ch^{2J+4},\qquad
 |R_{i,3}^z|^2\leq Ch^{2J}.
\end{equation}
From (\ref{multistep-dis-time-NS}) and (\ref{NS-M}), we have
\begin{equation}\label{MHDBR-err-esti-2-1}
\begin{aligned}
 \left| Y_{t_i}-Y_{i}^{\pi,m}\right| = & \left| \mathbb{E}_{i}\left[Y_{t_{i+1}}-Y_{i+1}^{\pi,m} \right] 
+\mathbb{E}_{i}\left[ \sum\limits_{j=0}^{J} \alpha_{j}h\left(f_{t_{i+j}}
-f_{i+j}^{\pi,m})\right)\right]
+ R^y_{i,3} \right|,
    \end{aligned}
\end{equation}
and 
\begin{equation}\label{MHDBR-err-esti-2-2}
\begin{aligned}
 \left| Z_{t_i}-Z_{i}^{\pi,m}\right| = & \left| \sum_{j=0}^{J}\beta_{J,j}\mathbb{E}_{i}\left[\left(Y_{t_{i+j}}
 -Y_{i+j}^{\pi,m}\right)\left(\Delta W_{i,i+j}^m\right)^\top\right] + R^z_{i,3}
 \right|,
    \end{aligned}
\end{equation}
where $ f_{i+j}^{\pi,m}=f(t_{i+j},X_{i+j}^{\pi,m},Y_{i+j}^{\pi,m},Z_{i+j}^{\pi,m})$.
Let
\[
 e_i^y=Y_{t_i}-Y_i^{\pi,m},\qquad
 e_i^z=Z_{t_i}-Z_i^{\pi,m},\qquad
 b_i=e_i^y-\mathbb E_i e_{i+1}^y.
\]
From \eqref{MHDBR-err-esti-2-1}, Jensen's inequality,
and the Lipschitz condition on $f$, we obtain
\begin{align*}
 \mathbb E|b_i|^2
 &\leq
 Ch^2\sum_{j=0}^J\mathbb E\Big[
 |e_{i+j}^y|^2
 +|e_{i+j}^z|^2\Big]+Ch^2\sum_{j=0}^J
 \mathbb E|X_{t_{i+j}}-X_{i+j}^{\pi,m}|^2
 +C\mathbb E|R_{i,3}^y|^2.
\end{align*}
By adding up the above inequality over $i=k,\ldots,N-J$, we deduce
\begin{align}
\sum_{i=k}^{N-J} \mathbb E|b_i|^2
 &\leq
 Ch^2\sum_{i=k}^N\mathbb E\Big[
 |e_{i}^y|^2
+|e_{i}^z|^2\Big]+C\sum_{i=k}^{N-J} \mathfrak r_i,
\label{lemmay1}
\end{align}
where
$
 \mathfrak r_i
 =
 h^2\sum_{j=0}^J
 \mathbb E|X_{t_{i+j}}-X_{i+j}^{\pi,m}|^2
 +\mathbb E|R_{i,3}^y|^2.
$
From \eqref{MHDBR-err-esti-2-2}, we have
\begin{align}
     e_i^z=\sum_{j=1}^J\beta_{J,j}
        \mathbb E_i[e_{i+j}^y(\Delta W_{i,i+j})^\top]+\rho_i^z,
        \label{lemmaz1}
\end{align}
where $\rho_i^z=R_{i,3}^z$. By \eqref{lemmay1}, \eqref{lemmaz1} and Lemma~\ref{noD0-stability}, we obtain
\begin{align}
 &\max_{0\leq i\leq N-J}\mathbb E|Y_{t_i}-Y_i^{\pi,m}|^2
        +h\sum_{i=0}^{N-J}\mathbb E|Z_{t_i}-Z_i^{\pi,m}|^2
        \notag\\
 &\leq C\sum_{i=N-J+1}^N
       \mathbb E\big[
        |Y_{t_i}-Y_i^{\pi,m}|^2+h|Z_{t_i}-Z_i^{\pi,m}|^2\big]
        \notag\\
 &\quad+C\sum_{i=0}^{N-J}\left[
       h\sum_{j=0}^J\mathbb E|X_{t_{i+j}}-X_{i+j}^{\pi,m}|^2
       +\frac1h\mathbb E|R_{i,3}^y|^2
                              +h\mathbb E|R_{i,3}^z|^2\right].
 \label{noD0-time-forcing}
\end{align}
Inserting \eqref{phi-X-err} and \eqref{MHDBR-err-esti-2-6} into \eqref{noD0-time-forcing}, we obtain
\begin{align}
 \max_{0\leq i\leq N-J}\mathbb E|Y_{t_i}-Y_i^{\pi,m}|^2
        +h\sum_{i=0}^{N-J}\mathbb E|Z_{t_i}-Z_i^{\pi,m}|^2
 \leq C(h^{2J}+h^{2\gamma_1}+h^{2J+2})
                   \leq Ch^{2\min\{J,\gamma_1\}} .
 \label{noD0-time-error}
\end{align}

Step 2. \underline{Estimates for $\mathbb{E}\left[\left|Y_i^{\pi,m}-\mathcal{Y}_{i} \right|^2\right]+ \sum\limits_{i=0}^{N-J}\mathbb{E}\left[h\left|Z_{i}^{\pi,m}-\mathcal{Z}_{i}\right|^2\right]$}. 
We first let
\[
 e_i^y=Y_i^{\pi,m}-\mathcal Y_i,\qquad
 e_i^z=Z_i^{\pi,m}-\mathcal Z_i,\qquad
 b_i=e_i^y-\mathbb E_i e_{i+1}^y.
\]
Define 
\begin{align}
 \rho_i^y
 &=(\mathcal U_i^m-\mathcal Y_i)
       +\alpha_0h\big[
          f(t_i,X_i^{\pi,m},\mathcal Y_i,\mathcal Z_i)
          -f(t_i,X_i^{\pi,m},\mathcal U_i^m,\mathcal V_i^m)
                   \big],\notag\\
 \rho_i^z&=\mathcal V_i^m-\mathcal Z_i .
 \label{noD0-fitting-residuals}
\end{align}
From the Lipschitz condition of $f$, we have the estimate
\begin{equation}\label{noD0-rho-bound}
 \mathbb E|\rho_i^y|^2
 \leq C\mathbb E|\mathcal U_i^m-\mathcal Y_i|^2
                  +Ch^2\mathbb E|\mathcal V_i^m-\mathcal Z_i|^2.
\end{equation}
From \eqref{NS-M}, \eqref{Deep-NN-E} and \eqref{noD0-fitting-residuals}, we have

\begin{align} 
 e_i^y
 &=\mathbb E_i[e_{i+1}^y]
       +\rho_i^y+h\sum_{j=0}^J\alpha_j\mathbb E_i\Big[
       f(t_{i+j},X_{i+j}^{\pi,m},Y_{i+j}^{\pi,m},Z_{i+j}^{\pi,m})
       -f(t_{i+j},X_{i+j}^{\pi,m},\mathcal Y_{i+j},\mathcal Z_{i+j})
                                           \Big],\label{lemmay2}\\
 e_i^z
 &=\sum_{j=1}^J\beta_{J,j}
       \mathbb E_i\Big[
          e_{i+j}^y
                                 (\Delta W_{i,i+j}^m)^\top\Big]
       +\rho_i^z.\label{lemmaz2}
\end{align}
From \eqref{lemmay2} and
the Lipschitz condition on $f$, we have
\begin{align}
 \sum_{i=k}^{N-J}\mathbb E|b_i|^2
 &\leq Ch^2\sum_{i=k}^N\mathbb E\Big[
             |e_i^y|^2
             +|e_i^z|^2\Big]
                                 +C\sum_{i=k}^{N-J}\mathbb E|\rho_i^y|^2 .\label{lemmay22}
\end{align}
By \eqref{lemmaz2}, \eqref{lemmay22} and Lemma~\ref{noD0-stability}, we obtain
\begin{align}
 &\max_{0\leq i\leq N-J}\mathbb E|Y_i^{\pi,m}-\mathcal Y_i|^2
       +h\sum_{i=0}^{N-J}\mathbb E|Z_i^{\pi,m}-\mathcal Z_i|^2
       \notag\\
 &\leq C\sum_{i=N-J+1}^{N}\mathbb E\Big[
        |Y_i^{\pi,m}-\mathcal Y_i|^2
                              +h|Z_i^{\pi,m}-\mathcal Z_i|^2\Big]
       \notag\\
 &\quad+C\sum_{i=0}^{N-J}\left[
        \frac1h\mathbb E|\mathcal U_i^m-\mathcal Y_i|^2
                    +h\mathbb E|\mathcal V_i^m-\mathcal Z_i|^2\right].
 \label{MHDBR-err-esti-3-7}
\end{align}
We now estimate the last two terms.
Using the auxiliary variables defined in
\eqref{Deep-NN-E--YZK}, we have
\begin{align}
\frac1h \mathbb E|\mathcal U_i^m-\mathcal Y_i|^2
       +h\mathbb E|\mathcal V_i^m-\mathcal Z_i|^2
 &\leq\frac2h\mathbb E|\widetilde{\mathcal U}_i^m-\mathcal U_i^m|^2
       +2h\mathbb E|\widetilde{\mathcal V}_i^m-\mathcal V_i^m|^2
 +\frac2h\mathbb E|\widetilde{\mathcal U}_i^m-\mathcal Y_i|^2
       +2h\mathbb E|\widetilde{\mathcal V}_i^m-\mathcal Z_i|^2.
 \label{local-mc-split}
\end{align}
In what follows, we estimate the four terms on the right-hand side of \eqref{local-mc-split}.

\smallskip
\noindent\emph{(a) Monte Carlo errors.}
Conditional independence of the complete branches gives
\[
 \mathbb E|\widetilde{\mathcal V}_i^m-\mathcal V_i^m|^2
 =\frac1{K_i}\mathbb E\mathbb D_i[H_i],\qquad
 H_i=\sum_{j=1}^J\beta_{J,j}
                 \mathcal Y_{i+j}(\Delta W_{i,i+j}^m)^\top.
\]
No independence between different $j$ on one branch is assumed.
By \eqref{MC-network-moment} and H\"older's inequality,
\begin{align*}
 \mathbb E[|\mathcal Y_{i+j}|^2|\Delta W_{i,i+j}^m|^2]
 &\leq(\mathbb E|\mathcal Y_{i+j}|^{2+\delta})^{2/(2+\delta)}
       (\mathbb E|\Delta W_{i,i+j}^m|^{2(2+\delta)/\delta})^{\delta/(2+\delta)}
 \leq Ch.
\end{align*}
Since $J$ is fixed and $h\beta_{J,j}$ is independent of $h$,
$\mathbb E|H_i|^2\leq C/h$. Hence
\begin{equation}\label{MC-z-rate}
 \mathbb E|\widetilde{\mathcal V}_i^m-\mathcal V_i^m|^2
 \leq\frac C{K_i h}.
\end{equation}
From \eqref{Deep-NN-E}, \eqref{Deep-NN-E--YZK} and the Lipschitz condition,
we obtain
\begin{align}
 (1-|\alpha_0|hL)|\widetilde{\mathcal U}_i^m-\mathcal U_i^m|
       \notag
 &\leq\Bigg|
   \frac1{K_i}\sum_{k=1}^{K_i}\Big[
       \mathcal Y_{i+1}^{m,k}
       +h\sum_{j=1}^J\alpha_j
        f(t_{i+j},X_{i+j}^{\pi,m,k},
                          \mathcal Y_{i+j}^{m,k},\mathcal Z_{i+j}^{m,k})
                             \Big]\notag\\
 &\hspace{6mm}
   -\mathbb E_i\Big[
       \mathcal Y_{i+1}
       +h\sum_{j=1}^J\alpha_j
        f(t_{i+j},X_{i+j}^{\pi,m},
                              \mathcal Y_{i+j},\mathcal Z_{i+j})
                 \Big]\Bigg|\notag
 +|\alpha_0|hL
           |\widetilde{\mathcal V}_i^m-\mathcal V_i^m|.
 \label{MC-y-implicit}
\end{align}
By Young's inequality and 
\eqref{time-step-condition}, we have
\begin{equation}
 \mathbb E|\widetilde{\mathcal U}_i^m-\mathcal U_i^m|^2
 \leq\frac C{K_i}\mathbb E\mathbb D_i\left[
       \mathcal Y_{i+1}
       +h\sum_{j=1}^J\alpha_j
          f(t_{i+j},X_{i+j}^{\pi,m},\mathcal Y_{i+j},\mathcal Z_{i+j})
                                             \right]
 +Ch^2\mathbb E
              |\widetilde{\mathcal V}_i^m-\mathcal V_i^m|^2.
 \label{MC-y-complete}
\end{equation}
We next justify the moment bound for the complete $Y$-target,
including its learned $Z$-inputs. Since zero is an admissible network,
exact population minimization gives
\[
 \|\mathcal Z_i\|_2
 \leq\|\widetilde{\mathcal V}_i^m\|_2
      +\|\widetilde{\mathcal V}_i^m-\mathcal Z_i\|_2
 \leq2\|\widetilde{\mathcal V}_i^m\|_2\leq Ch^{-1/2}.
\]
The last bound follows from Jensen's inequality for the branch average
and $\mathbb E|H_i|^2\leq C/h$.
For the starting layers, Assumption 2 bounds
$Z_{t_q}^{(r)}=\mathcal L^r u(t_q,X_{t_q})$, and
\eqref{starting-errors} implies
$\mathbb E|\mathcal Z_q|^2\leq C(1+h^{2J-1})\leq C$.
Thus $\mathbb E|\mathcal Z_q|^2\leq C/h$ at every layer for $h\leq1$.
The linear-growth bound on $f$ from Assumption 1, the forward moments
in Assumption 3 and \eqref{MC-network-moment} now imply
\[
 \mathbb E\left|\mathcal Y_{i+1}
       +h\sum_{j=1}^J\alpha_j
          f(t_{i+j},X_{i+j}^{\pi,m},\mathcal Y_{i+j},\mathcal Z_{i+j})
       \right|^2
 \leq C\left[1+h^2\sum_{j=1}^J(1+\mathbb E|\mathcal Z_{i+j}|^2)\right]
 \leq C.
\]
Combining this estimate with \eqref{MC-y-complete} and
\eqref{MC-z-rate} gives
\begin{equation}\label{MC-y-rate}
 \mathbb E|\widetilde{\mathcal U}_i^m-\mathcal U_i^m|^2
       \leq\frac C{K_i}.
\end{equation}

\smallskip
\noindent\emph{(b) Neural network approximation errors.}
From \eqref{POP-LOSS-Z} and the minimizing property of
$\theta_{z,i}^*$, we obtain
\begin{equation}\label{local-fit-z-sampled}
 \mathbb E|\widetilde{\mathcal V}_i^m-\mathcal Z_i|^2
       =\widetilde{\varepsilon}_i^z.
\end{equation}
By the reverse triangle inequality and the Lipschitz
condition on $f$, we have
\begin{align}
 &(1-|\alpha_0|hL)
 \left|\widetilde{\mathcal U}_i^m-
       \mathcal{NN}_{y,i,1+d,n}^{\varrho,L,\mathbf n}
                              (p_i^m,\theta_{y,i}^{r_j})\right|
       \notag\\
 &\leq\Bigg|\widetilde{\mathcal U}_i^m
       +\alpha_0h\Bigg[
           f\left(t_i,X_i^{\pi,m},
              \mathcal{NN}_{y,i,1+d,n}^{\varrho,L,\mathbf n}
                                  (p_i^m,\theta_{y,i}^{r_j}),
              \mathcal{NN}_{z,i,1+d,n\times d}^{\varrho,L,\mathbf n}
                                  (p_i^m,\theta_{z,i}^*)\right)
       \notag\\
 &\hspace{7mm}
           -f(t_i,X_i^{\pi,m},
                     \widetilde{\mathcal U}_i^m,\widetilde{\mathcal V}_i^m)
                          \Bigg]
       -\mathcal{NN}_{y,i,1+d,n}^{\varrho,L,\mathbf n}
                                  (p_i^m,\theta_{y,i}^{r_j})\Bigg|
 +|\alpha_0|hL
               |\widetilde{\mathcal V}_i^m-\mathcal Z_i|.
 \label{implicit-residual-lower}
\end{align}
Similarly, the absolute value on the right-hand side is
bounded by
\begin{equation}\label{implicit-residual-upper}
 (1+|\alpha_0|hL)
 \left|\widetilde{\mathcal U}_i^m-
       \mathcal{NN}_{y,i,1+d,n}^{\varrho,L,\mathbf n}
                             (p_i^m,\theta_{y,i}^{r_j})\right|
       +|\alpha_0|hL
               |\widetilde{\mathcal V}_i^m-\mathcal Z_i|.
\end{equation}
Let $\mathcal F_{y,i}^*(\theta_{y,i}^*)$ denote the minimum
of \eqref{POP-LOSS-Y} with $\theta_{z,i}^*$ fixed.
By the minimizing property of $\theta_{y,i}^*$, \eqref{implicit-residual-upper}, and
\eqref{local-fit-z-sampled}, we obtain
\begin{align}
 \mathcal F_{y,i}^{*}(\theta_{y,i}^*)
 \leq
 2(1+|\alpha_0|hL)^2\widetilde{\varepsilon}_i^y
          +2\alpha_0^2h^2L^2\widetilde{\varepsilon}_i^z.
 \label{local-fit-optimality}
\end{align}
Applying \eqref{implicit-residual-lower} with
$\theta_{y,i}^{r_j}=\theta_{y,i}^*$, squaring both sides,
and using Young's inequality,
\eqref{local-fit-z-sampled}, and \eqref{local-fit-optimality},
we obtain
\begin{align}
 (1-|\alpha_0|hL)^2
       \mathbb E|\widetilde{\mathcal U}_i^m-\mathcal Y_i|^2
       \notag
 \leq
       4(1+|\alpha_0|hL)^2\widetilde{\varepsilon}_i^y
          +6\alpha_0^2h^2L^2\widetilde{\varepsilon}_i^z.
 \label{local-fit-before-division}
\end{align}
By \eqref{time-step-condition}, it follows that
\begin{equation}\label{sampled-y-fit}
 \mathbb E|\widetilde{\mathcal U}_i^m-\mathcal Y_i|^2
       \leq C\left(
          \widetilde{\varepsilon}_i^y
                      +h^2\widetilde{\varepsilon}_i^z\right).
\end{equation}

\smallskip
\noindent\emph{(c) Estimates in terms of
$\varepsilon_i^y$ and $\varepsilon_i^z$.}
By the definitions of $\widetilde{\varepsilon}_i^y$,
$\widetilde{\varepsilon}_i^z$, $\varepsilon_i^y$, and
$\varepsilon_i^z$, Young's inequality,
\eqref{MC-y-rate}, and \eqref{MC-z-rate}, we obtain
\begin{align}
 \widetilde{\varepsilon}_i^y
 &\leq2\mathbb E|\widetilde{\mathcal U}_i^m-\mathcal U_i^m|^2
                +2\varepsilon_i^y
 \leq C\left(\varepsilon_i^y+\frac1{K_i}\right),
 \label{sampled-eps-y}\\
 \widetilde{\varepsilon}_i^z
 &\leq2\mathbb E|\widetilde{\mathcal V}_i^m-\mathcal V_i^m|^2
                +2\varepsilon_i^z
 \leq C\left(\varepsilon_i^z+\frac1{K_i h}\right).
 \label{sampled-eps-z}
\end{align}
Combining \eqref{local-mc-split},
\eqref{local-fit-z-sampled}, \eqref{sampled-y-fit}, 
\eqref{sampled-eps-y}, and \eqref{sampled-eps-z}, we deduce
\begin{align}
 \mathbb E|\mathcal U_i^m-\mathcal Y_i|^2
 &\leq C\left(\varepsilon_i^y+h^2\varepsilon_i^z
                                      +\frac1{K_i}\right),
 \label{Deep-NN-E-10h}\\
 \mathbb E|\mathcal V_i^m-\mathcal Z_i|^2
 &\leq C\left(\varepsilon_i^z+\frac1{K_i h}\right).
 \label{local-fit-z}
\end{align}
Multiplying \eqref{Deep-NN-E-10h} by $\frac{1}{h}$
and \eqref{local-fit-z} by $h$, and adding the resulting
inequalities, we obtain
\begin{align}
 \frac{1}{h}\mathbb E|\mathcal U_i^m-\mathcal Y_i|^2
       +h\mathbb E|\mathcal V_i^m-\mathcal Z_i|^2    
 &\leq C\left(
       \frac{\varepsilon_i^y}{h}+h\varepsilon_i^z
                        +\frac1{hK_i}+\frac1{K_i}\right)\notag\\
 &\leq C\left(
       \frac{\varepsilon_i^y}{h}+h\varepsilon_i^z+\frac1{hK_i}\right).
 \label{local-fit-weighted}
\end{align}
Inserting \eqref{local-fit-weighted} into \eqref{MHDBR-err-esti-3-7} leads to
\begin{align}
 &\max_{0\leq i\leq N-J}\mathbb E|Y_i^{\pi,m}-\mathcal Y_i|^2
      +h\sum_{i=0}^{N-J}\mathbb E|Z_i^{\pi,m}-\mathcal Z_i|^2
       \notag\\
 &\leq C\sum_{i=N-J+1}^N\mathbb E\Big[
       |Y_i^{\pi,m}-\mathcal Y_i|^2
       +h|Z_i^{\pi,m}-\mathcal Z_i|^2\Big]
 +C\sum_{i=0}^{N-J}
       \left(\frac{\varepsilon_i^y}{h}
                    +h\varepsilon_i^z+\frac1{hK_i}\right).
 \label{MHDBR-err-esti-3-1b}
\end{align}

Step 3. Plugging  (\ref{noD0-time-error}) and (\ref{MHDBR-err-esti-3-1b}) into (\ref{MHDBR-err-esti-1}), we deduce
\begin{equation}
\begin{aligned}
&\max\limits_{0\le i\le N-J}\mathbb{E}\left[\left|Y_{t_i}-\mathcal{Y}_{i}\right|^2\right]
+\sum\limits_{i=0}^{N-J}h\mathbb{E}\left[|Z_{t_i}-\mathcal{Z}_{i}|^2 \right]
\le   Ch^{\min \{2J,~2\gamma_1\}} 
      + C\sum_{i=0}^{N-J}
       \left(\frac{\varepsilon_i^y}{h}
                    +h\varepsilon_i^z+\frac1{hK_i}\right).
\nonumber
\end{aligned}
\end{equation}
This completes the proof.
\end{proof}

Next, 
we derive a quantitative convergence result for
the proposed MHDBR method using GroupSort neural networks
\cite{ACLJGR19,GMPHWX22}.

\begin{theorem}\label{Thm-GS-convergence}
Suppose that the hypotheses of Theorem~\ref{MHDBR-err-esti} hold with $\gamma_1\geq J$, and  that the generator $f$
is independent of $z$, i.e.,
$f(t,x,y,z)=f(t,x,y)$.
Let $\delta>0$ be as in \eqref{MC-network-moment}, and let $g$
be Lipschitz continuous. 
Assume that the functions defining the starting values
$\{\mathcal Y_i\}_{i=N-J+1}^{N}$ have spatial Lipschitz
constants bounded independently of $N$.
There exist positive sequences
$\{\mathcal K_{y,i}\}_{i=0}^{N-J}$ and
$\{\mathcal K_{z,i}\}_{i=0}^{N-J}$, specified in Appendix~C,
such that
\[
\max_{0\leq i\leq N-J}\mathcal K_{y,i}
+
\max_{0\leq i\leq N-J}\sqrt h\,\mathcal K_{z,i}
\leq C,
\]
where $C$ is independent of $N$ and $h$.
For $i=0,\ldots,N-J$, perform the exact population
minimizations over GroupSort networks whose spatial
slices satisfy
$ x\longmapsto
 \mathcal{NN}_{y,i,1+d,n}^{\varrho,L,\mathbf n}
                    ((t_i,x),\theta_{y,i}^{r_j})
 \in\mathcal G_{\mathcal K_{y,i},d,n}^{\zeta_\kappa,L,\mathbf n},
 x\longmapsto
 \mathcal{NN}_{z,i,1+d,n\times d}^{\varrho,L,\mathbf n}
                    ((t_i,x),\theta_{z,i}^{r_j})
 \in\mathcal G_{\mathcal K_{z,i},d,nd}^{\zeta_\kappa,L,\mathbf n}.
$
Set
$K_{\mathrm{MC}}:=\min_{0\leq i\leq N-J}K_i$ and choose
\begin{equation}\label{GS-parameter-choice}
 \begin{aligned}
 R&\asymp N^{(2J+2)/\delta},
 &\eta&\asymp N^{-(J+1)-(2J+2)/\delta},
 &K_{\mathrm{MC}}&\asymp N^{2J+2}.
 \end{aligned}
\end{equation}
Then the approximations satisfy
\begin{equation}\label{GS-final-nodal}
\max_{0\leq i\leq N-J}\mathbb E|Y_{t_i}-\mathcal Y_i|^2
+h\sum_{i=0}^{N-J}\mathbb E|Z_{t_i}-\mathcal Z_i|^2
\leq Ch^{2J},
\end{equation}
where $C$ is independent of $N,h$ and $K_i$.
For $d>1$, the scalar component networks can be chosen with
\[
 \begin{aligned}
 \kappa&=O\!\left(\left\lceil2\sqrt d\,
                         N^{J+1+(2J+2)/\delta}\right\rceil\right),
 &L+1&=O(d^2),\\
 \sum_{\ell=0}^{L-1}n_\ell
   &=O\!\left(\left(2\sqrt d\,
                    N^{J+1+(2J+2)/\delta}\right)^{d^2-1}\right).
 \end{aligned}
\]
For $d=1$, one can take $L+1=3$ and
\[
 \kappa=O\big(N^{J+1+(2J+2)/\delta}\big),\qquad
 \sum_{\ell=0}^{L-1}n_\ell=O\big(N^{J+1+(2J+2)/\delta}\big).
\]
Parallel scalar networks give the vector and matrix outputs;
the neuron counts acquire factors $n$ and $nd$, respectively.
The displayed asymptotic estimates are for fixed dimensions.

\end{theorem}
\begin{proof}
A proof, including the spatial Lipschitz constraints and the truncation and approximation estimates, is given in Appendix C.
\end{proof}

\section{Numerical experiments}

This section presents three reported neural-network benchmarks and a
separate verification of the underlying time-discretization scheme.
The benchmarks describe finite-budget numerical performance; the final
test isolates temporal consistency by using analytic conditional
expectations and exact starting values.
The coefficients of the multistep high-order schemes are determined
by the order conditions \eqref{parameter-alpha} and
\eqref{parameter-beta}. For reference, Table~\ref{tex2} lists the
corresponding coefficients for $J=1,2,3,4$.

\begin{table}[!htbp]
  \centering
  \begin{threeparttable}
    \caption{The coefficients $\alpha_j$, $\beta_{J,j}$ for the multistep high-order scheme}
    \label{tex2}
    \begin{tabular}{l c c c c c c c c c c}  
      \toprule
      & $\alpha_0$ & $\alpha_1$ & $\alpha_2$ & $\alpha_3$ & $\alpha_4$ & $\beta_{J,0}$ & $\beta_{J,1}$ & $\beta_{J,2}$ & $\beta_{J,3}$ & $\beta_{J,4}$ \\
      \midrule
      $J=1$ & $\frac{1}{2}$ & $\frac{1}{2}$ & & & & $-\frac{1}{h}$ & $\frac{1}{h}$ & & &  \\
      $J=2$ & $\frac{5}{12}$ & $\frac{2}{3}$ & $-\frac{1}{12}$ & &  & $-\frac{3}{2h}$ & $\frac{2}{h}$ & $-\frac{1}{2h}$ & & \\
      $J=3$ & $\frac{3}{8}$ & $\frac{19}{24}$ & $-\frac{5}{24}$ & $\frac{1}{24}$ & & $-\frac{11}{6h}$ & $\frac{3}{h}$ & $-\frac{3}{2h}$ & $\frac{1}{3h}$ &\\
      $J=4$ & $\frac{251}{720}$ & $\frac{323}{360}$ & $-\frac{11}{30}$ & $\frac{53}{360}$ & $-\frac{19}{720}$ & $-\frac{25}{12h}$ & $\frac{4}{h}$ & $-\frac{3}{h}$ & $\frac{4}{3h}$ & $-\frac{1}{4h}$ \\
      \bottomrule
    \end{tabular}
  \end{threeparttable}
\end{table}

The three neural-network benchmarks each lie outside at least one
hypothesis of the convergence analysis. Their precise departures from
Assumptions 1--3 are identified below. In particular, the quantitative
GroupSort result in Theorem~\ref{Thm-GS-convergence} assumes a generator
independent of $z$ and exact minimization of population losses.
Results from finite-sample network training are therefore interpreted
as empirical benchmark outcomes, rather than as a direct verification
of that theorem. The additional manufactured problem at the end of
this section has bounded smooth solution data and a $z$-independent
generator; its verification concerns only the time-discrete scheme.

The neural-network benchmarks in Examples 1--3 are implemented in Python 3.10.18 and
performed on a GPU server running Ubuntu 22.04.5 LTS, equipped with
a 20-core Intel Xeon Gold 6133 processor (2.50 GHz), 48 GB RAM, and
an NVIDIA GeForce RTX 4090 GPU with 24 GB of memory. The corresponding code is available at: \url{https://github.com/22w2e/MHDBR-for-high-dimensional-nonlinear-parabolic-PDEs}.
Since the reported results are subject to stochastic variability from random sampling and neural-network training, each experiment is repeated independently 
$N_{\mathrm{rep}}=10$ times—using the same numerical configuration but independent random samples—a choice that offers a practical compromise between statistical reliability and computational cost, in line with the related literature \cite{HCPHWX20,GMPHWX22}.
A convergence-order study of the complete trained implementation is
not included. Such a study would require simultaneous control of the
inner sampling, network approximation, empirical fitting, optimization,
and starting-value errors. For example, the sufficient choice
$K_{\mathrm{MC}}\asymp N^{2J+2}$ in
Theorem~\ref{Thm-GS-convergence} is not the sampling schedule used in
Table~2, where $K=0.08N^2$. The corresponding term $N^2/K$ in the
proved upper bound does not decay under this schedule; this observation
limits what can be inferred from that bound, and does not assert
nonconvergence of the implementations.

The tabulated records provide the stated $N$, $K$, $M$, and repeated-run
statistics, but do not separately document all network architectures,
optimization schedules, and startup computations. These records do not
permit an independent verification of equal computational cost or of
the individual error contributions. Comparisons below are consequently
restricted to the reported configurations and statistics. In
particular, the finite-budget benchmarks do not certify the
starting-value accuracy required by \eqref{starting-errors}.

For clarity, let $u_{\mathrm{ref}}(0,x_0)$ denote the reference value at the
initial point: the analytic value in Examples 1 and 3, and the numerical
reference in Example 2. Let $\overline{\mathfrak{u}}_{\ell}(0,x_0)$ denote the
numerical approximation obtained in the $\ell$-th independent run.
The sample mean is defined by 
$\overline{\mathfrak{u}}(0,x_0):=\frac{1}{N_{rep}}\sum_{\ell=1}^{N_{rep}}\overline{\mathfrak{u}}_{\ell}(0,x_0), N_{rep}=10.$ 
The reported empirical standard deviation, mean absolute deviation,
and relative deviation are defined by
\[
\begin{aligned}
 \operatorname{Std.dev.}
 &=\left(\frac1{N_{\mathrm{rep}}}
    \sum_{\ell=1}^{N_{\mathrm{rep}}}
    \left|\overline{\mathfrak u}_{\ell}(0,x_0)
                 -\overline{\mathfrak u}(0,x_0)\right|^2\right)^{1/2},\\
 \operatorname{MAE.}
 &=\frac1{N_{\mathrm{rep}}}\sum_{\ell=1}^{N_{\mathrm{rep}}}
    \left|u_{\mathrm{ref}}(0,x_0)
                  -\overline{\mathfrak u}_{\ell}(0,x_0)\right|,\\
 \operatorname{Rel.err.}
 &=\frac{\left|\overline{\mathfrak u}(0,x_0)
                          -u_{\mathrm{ref}}(0,x_0)\right|}
              {|u_{\mathrm{ref}}(0,x_0)|}.
\end{aligned}
\]

Here $\operatorname{Rel.err.}$ measures the deviation of the sample
mean from the reference value, whereas $\operatorname{MAE.}$ measures
the average absolute deviation of individual runs. They answer
different questions: cancellation between positive and negative
errors can make the former much smaller. In Example 2 both statistics
are deviations from a numerical reference and include its unknown
reference error. No confidence interval for that reference is
available from the tabulated records.

\begin{table}[!htb]
  \centering
   \captionsetup{skip=4pt}
  \begin{threeparttable}
    \caption{Comparison of numerical results for different schemes in Example~1.}
    \setlength{\tabcolsep}{8pt}
  \renewcommand{\arraystretch}{0.90}
    \begin{tabular*}{\linewidth}
      {@{\extracolsep{\fill}}ccccccccc@{}}
      \toprule
      Scheme & $N$ & $K$ & $M$ & $u(0,x_0)$ & $\overline{\mathfrak{u}}(0,x_0)$ &
      $\operatorname{Std.dev.}$ &
      $\operatorname{MAE.}$ &
      $\operatorname{Rel.err.}$ \\
      \midrule
      \multirow{4}{*}{MHDBR (J=1) }  
    & 10 & 8 & 25 & 0.500000 & 0.484017 & 0.009518 & 0.015983 & 3.1965\%  \\
     & 20 & 32 & 100 & 0.500000 & 0.484432 & 0.008014 &  0.016292 & 3.1136\%  \\
    & 30 & 72 & 225 & 0.500000 & 0.494030 & 0.008737 & 0.007493 & 1.1939\%  \\ 
    & 40 & 128 & 400 & 0.500000 & 0.495502 & 0.014221 &  0.012822 & 0.8995\%  \\
    \midrule
      \multirow{4}{*}{MHDBR (J=2) }  
    & 10 & 8 & 25 & 0.500000 & 0.486736 & 0.016251 & 0.017812 & 2.6528\%  \\
    & 20 & 32 & 100 & 0.500000 & 0.490325 & 0.009863 &  0.011918 & 1.9351\%  \\
    & 30 & 72 & 225 & 0.500000 & 0.490605 & 0.009985 & 0.012076 & 1.8790\%  \\  
    & 40 & 128 & 400 & 0.500000 & 0.500081 & 0.010305 &  0.007644 & 0.0161\%  \\
     \midrule
      \multirow{4}{*}{MHDBR (J=3) }  
    & 10 & 8 & 25 & 0.500000 & 0.497102 & 0.009793 & 0.008586 & 0.5796\%  \\
    & 20 & 32 & 100 & 0.500000 & 0.490351 & 0.009499 &  0.011089 & 1.9297\%  \\
    & 30 & 72 & 225 & 0.500000 & 0.491011 & 0.010475 & 0.012334 & 1.7978\%  \\
    & 40 & 128 & 400 & 0.500000 & 0.491102 & 0.011110 &  0.011986 & 1.7796\%  \\
    \midrule
      \multirow{4}{*}{MHDBR (J=4) }  
    & 10 & 8 & 25 & 0.500000 & 0.493349 & 0.016596 & 0.013895 & 1.3302\%  \\
    & 20 & 32 & 100 & 0.500000 & 0.487880 & 0.008448 &  0.012547 & 2.4241\%  \\
    & 30 & 72 & 225 & 0.500000 & 0.491895 & 0.002976 & 0.008105 & 1.6209\%  \\
    & 40 & 128 & 400 & 0.500000 & 0.501081 & 0.007618 &  0.005383 & 0.2161\%  \\
    \midrule
      \multirow{4}{*}{MDBDP \cite{GMPHWX22}}  
    & 10 & 8 & 25 & 0.500000 & 0.533503 & 0.015276 & 0.033503 & 6.7006\%  \\
    & 20 & 32 & 100 & 0.500000 & 0.527751 & 0.007215 &  0.027751 & 5.5503\%  \\
    & 30 & 72 & 225 & 0.500000 & 0.530423 & 
    0.004106 & 0.030423 & 6.0847\%  \\ 
    & 40 & 128 & 400 & 0.500000 & 0.528022 & 0.002186  &  0.028022 & 5.6044\%  \\
      \bottomrule
    \end{tabular*}
  \end{threeparttable}
\end{table}

\subsection{Neural-network experiments on benchmark problems}
\label{sec:neural-network-experiments}

\textbf{Example 1.}
We first consider the following $d$-dimensional nonlinear parabolic PDE (taken from \cite{EWHJJA17}):
\begin{equation}
\label{PDEdetail-1}
\begin{cases}
\displaystyle
\frac{\partial u}{\partial t}(t,x)
+  \frac{\sigma^2}{2}\sum_{\ell=1}^{d}D^2_{x_\ell}u(t,x)
+ \Big(u(t,x) - \frac{d+2}{2d}\Big)
    \Big(\sigma \sum_{\ell=1}^{d}  D_{x_\ell}u(t,x)\Big)
= 0, & (t,x)\in [0,T)\times\mathbb{R}^d,\\[1.0em]
\displaystyle
u(T,x)=\frac{\exp\!\left(T+\tfrac{1}{d}\sum_{\ell=1}^{d}x_\ell\right)}
           {1+\exp\!\left(T+\tfrac{1}{d}\sum_{\ell=1}^{d}x_\ell\right)} .
\end{cases}
\end{equation}
The corresponding analytic solution is explicitly given by
$
u(t,x)
=\frac{\exp\!\left(t+\tfrac{1}{d}\sum_{\ell=1}^{d}x_\ell\right)}
        {1+\exp\!\left(t+\tfrac{1}{d}\sum_{\ell=1}^{d}x_\ell\right)}.
$
Let $T=1$, $\sigma=d$, and
$x_0=(0,\ldots,0)^\top\in\mathbb{R}^d$, which gives
$u(0,x_0)=0.5$.
This parameter choice is consistent with the stated analytic solution:
writing $v=u(t,x)$, substitution gives the residual
\[
 v(1-v)\left[1+\frac{\sigma^2}{2d}(1-2v)
       +\sigma\left(v-\frac{d+2}{2d}\right)\right]=0
 \quad\text{when }\sigma=d.
\]
Although $u$ and the along-solution function $F_u$ have the required
bounded Lie derivatives for fixed $d$, the generator
$f(y,z)=(y-(d+2)/(2d))\sum_\ell z_\ell$ is not globally Lipschitz in
$(y,z)$. It violates Assumption 1(iii) and depends on $z$, so neither
the general error theorem under its stated hypotheses nor the
$z$-independent GroupSort theorem is invoked for this raw benchmark.
The parameter convention for interpreting this example is $\sigma=d$.
The aggregate tables and plots are retained as recorded benchmark
outputs, rather than presented as fresh reruns or as an independent
check of the coefficients used in every training run.

We test the MHDBR schemes with $J=1,2,3,4$ on the $d=200$ problem.
For each $J$, we vary the number of time steps $N$ and adjust the Monte
Carlo sample sizes $K$ and $M$ accordingly. The MDBDP method in
\cite{GMPHWX22} is included as a benchmark. 

Table~2 reports the MHDBR and MDBDP results for $d=200$.
For the listed configurations, both the sample-mean deviations and
the MAEs are smaller for MHDBR. These are descriptive comparisons of
the reported runs, without a cost-normalized or statistical
significance claim. For $J=2$ and $N=40$, the relative sample-mean
deviation is $0.0161\%$, but the MAE is $0.007644$, corresponding to a
relative MAE of approximately $1.53\%$, and the empirical standard
deviation is $0.010305$. Thus the small deviation of the mean should
not be interpreted as the typical error of one trained solver.
The $J=4$, $N=40$ configuration has the smallest listed MAE,
$0.005383$. The nonmonotone errors in $N$ and $J$ do not establish
the temporal order of the fully trained method.

\begin{figure}[!htb]
  \centering
\begin{minipage}{0.6\textwidth}
\centering
  \begin{subfigure}[t]{0.485\linewidth}
    \centering
    \includegraphics[width=\linewidth]
    {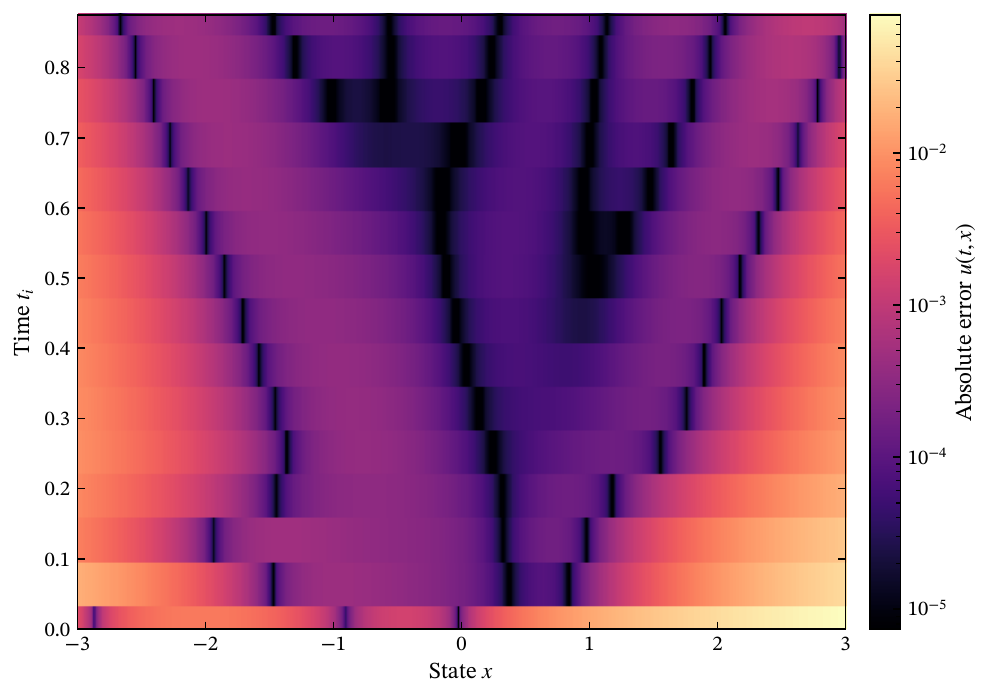}
    \caption{$N=16$.}
    \label{fig:ex1-j2-heatmap-n16}
  \end{subfigure}
  \hfill
  \begin{subfigure}[t]{0.485\linewidth}
    \centering
    \includegraphics[width=\linewidth]
    {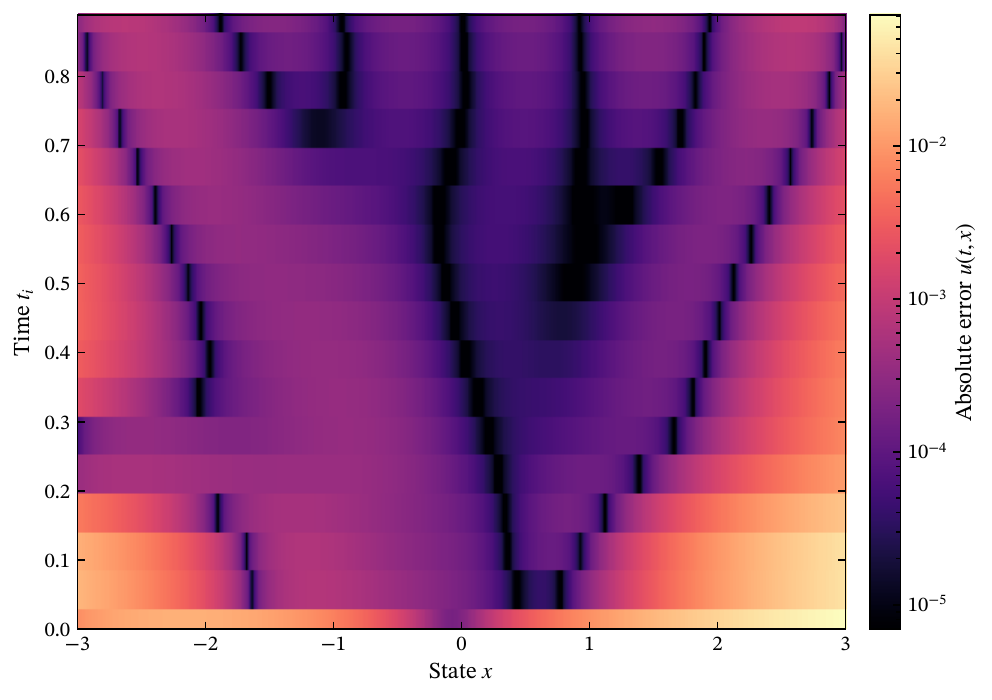}
    \caption{$N=18$.}
    \label{fig:ex1-j2-heatmap-n18}
  \end{subfigure}

  \vspace{0.6em}

  \begin{subfigure}[t]{0.485\linewidth}
    \centering
    \includegraphics[width=\linewidth]
    {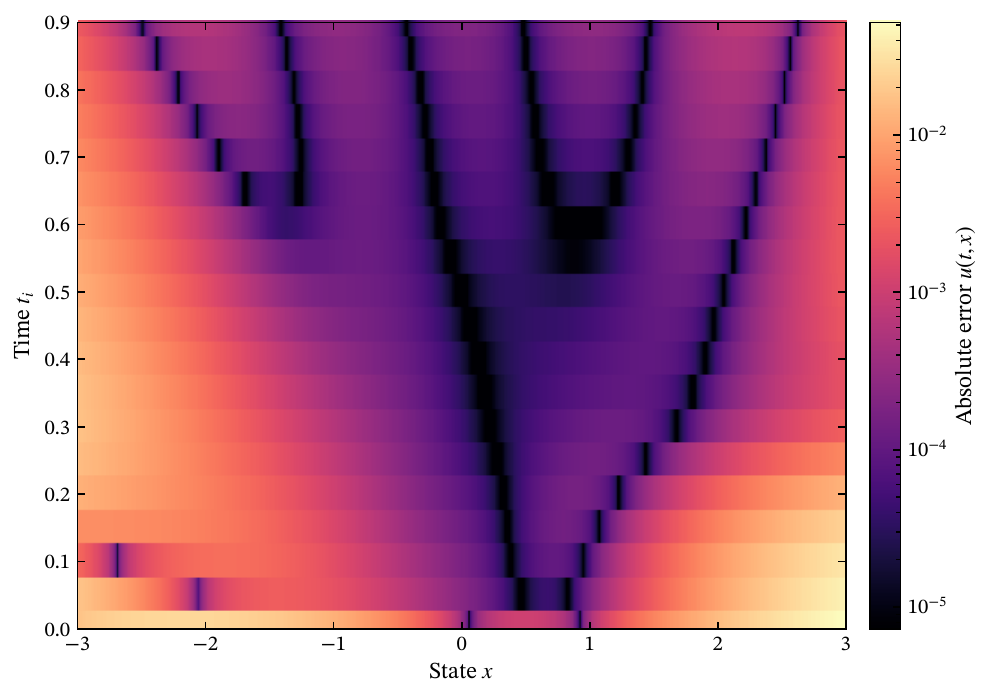}
    \caption{$N=20$.}
    \label{fig:ex1-j2-heatmap-n20}
  \end{subfigure}
  \hfill
  \begin{subfigure}[t]{0.485\linewidth}
    \centering
    \includegraphics[width=\linewidth]
    {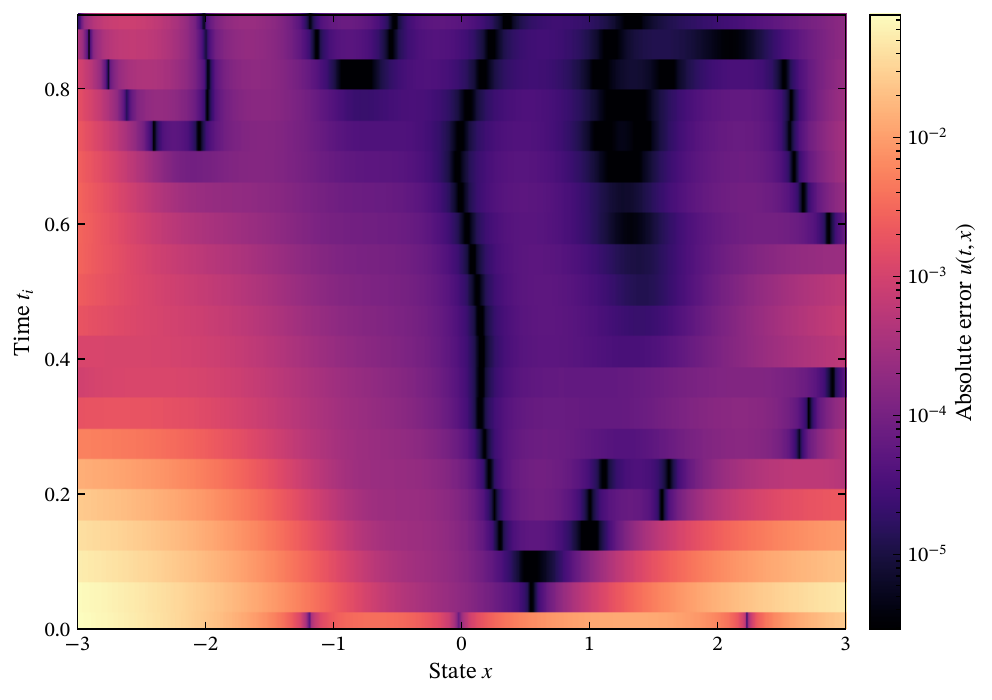}
    \caption{$N=22$.}
    \label{fig:ex1-j2-heatmap-n22}
  \end{subfigure}
  \end{minipage}

  \caption{
  Time--state distributions of the absolute error
for the one-dimensional case ($d=1$) of Example~1,
computed by the MHDBR $(J=2)$ scheme with different values of $N$.
  }
  \label{fig:ex1-j2-error-heatmaps}
\end{figure}

\begin{figure}[!htb]
  \centering
  \begin{minipage}{0.6\textwidth}
\centering

  \begin{subfigure}[t]{0.485\linewidth}
    \centering
    \includegraphics[width=\linewidth]
    {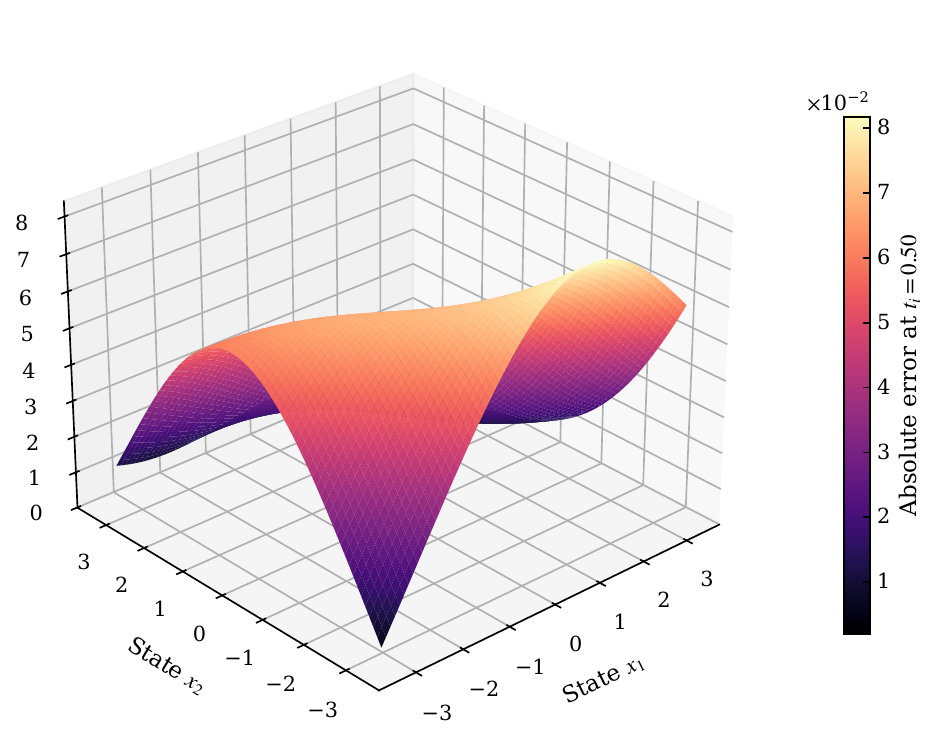}
    \caption{$J=1$.}
    \label{fig:ex1-j1-surface}
  \end{subfigure}
  \hfill
  \begin{subfigure}[t]{0.485\linewidth}
    \centering
    \includegraphics[width=\linewidth]
    {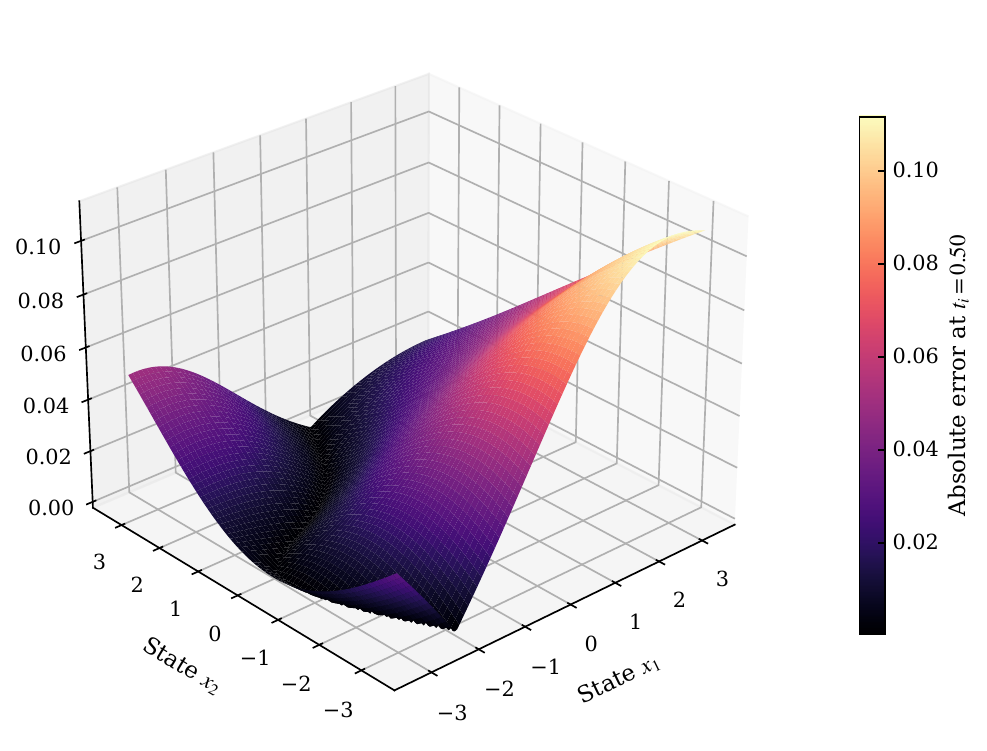}
    \caption{$J=2$.}
    \label{fig:ex1-j2-surface}
  \end{subfigure}

  \vspace{0.6em}

  \begin{subfigure}[t]{0.485\linewidth}
    \centering
    \includegraphics[width=\linewidth]
    {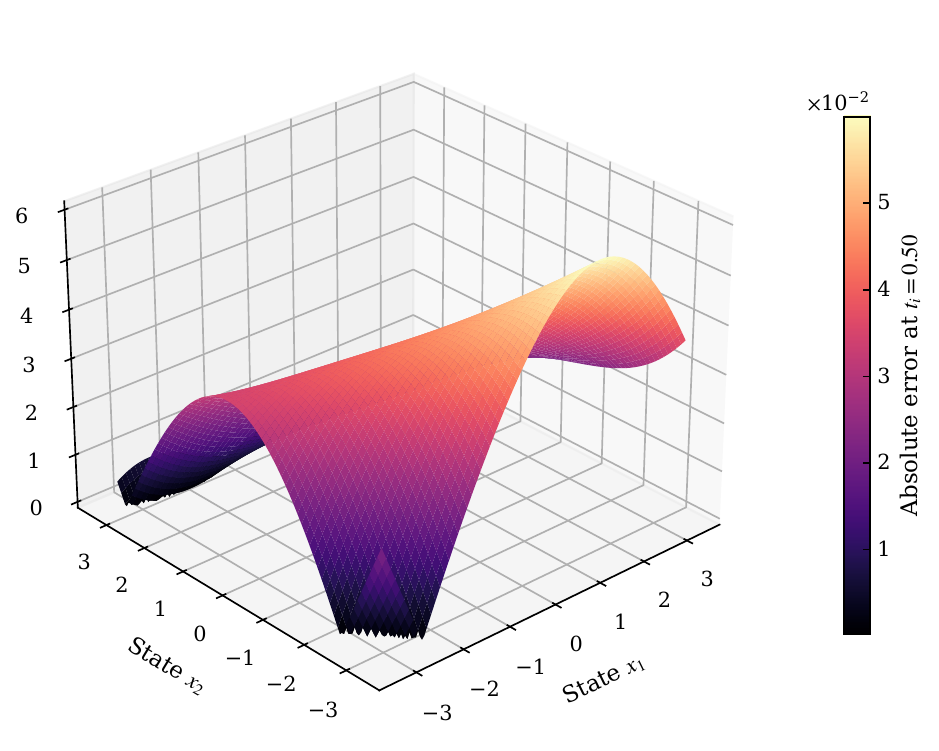}
    \caption{$J=3$.}
    \label{fig:ex1-j3-surface}
  \end{subfigure}
  \hfill
  \begin{subfigure}[t]{0.485\linewidth}
    \centering
    \includegraphics[width=\linewidth]
    {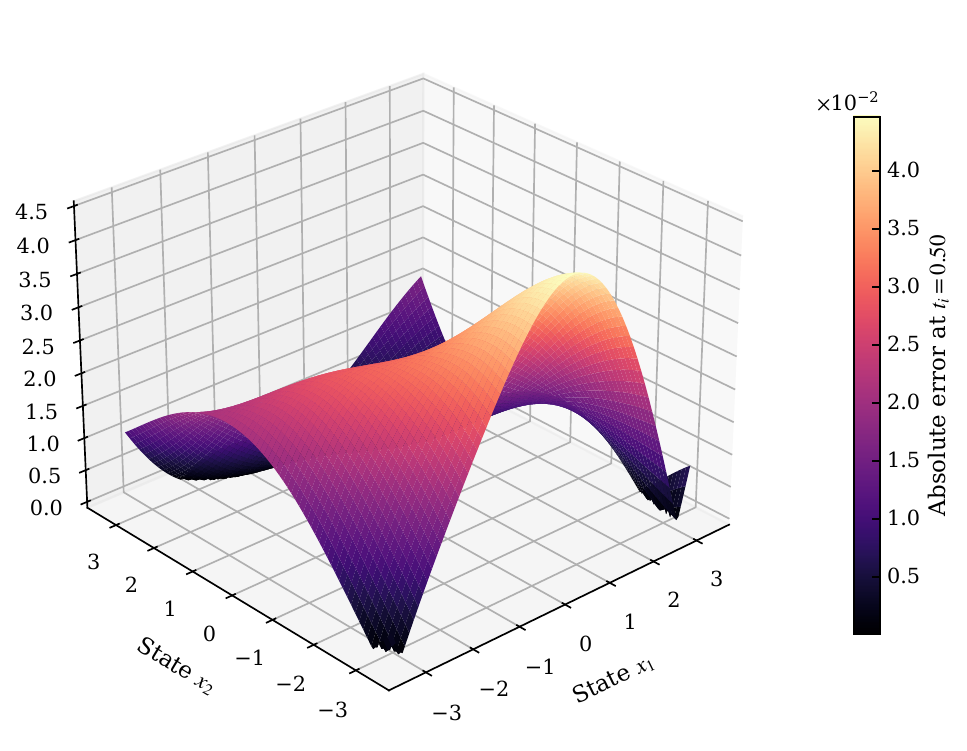}
    \caption{$J=4$.}
    \label{fig:ex1-j4-surface}
  \end{subfigure}
  \end{minipage}

  \caption{
    Surface plots of the absolute error
over the two-dimensional state space ($d=2$) for Example~1,
computed by the MHDBR schemes with $J=1,2,3,4$.
  }
  \label{fig:ex1-error-surface}
\end{figure}

Figures~\ref{fig:ex1-j2-error-heatmaps} and
\ref{fig:ex1-error-surface} display the recorded pointwise errors in
one and two dimensions. The heatmaps have comparable error magnitudes
for the plotted values of $N$, and the surfaces describe the error at
$t=0.5$. These views concern only the displayed states and times;
they do not establish a uniform error bound, a convergence rate,
or a numerical stability theorem.

 \begin{table}[!htb]
  \centering
  \captionsetup{skip=4pt}
  \begin{threeparttable}
    \caption{Numerical results of the MHDBR ($J=1$) scheme for
    Example~2 with $d=50$.}
    \label{tab:ex2-mhdbr-j1-d50}
    \setlength{\tabcolsep}{8pt}
  \renewcommand{\arraystretch}{0.90}
    \begin{tabular*}{\linewidth}
      {@{\extracolsep{\fill}}ccccccc@{}}
      \toprule
     $N$ & $K$ & $M$ 
      & $\overline{\mathfrak{u}}(0,x_0)$
      & $\operatorname{Std.dev.}$
      & $\operatorname{MAE.}$
      & $\operatorname{Rel.err.}$ \\
      \midrule
       4 & 64 & 256
      & 17.9231
      & 0.1663
      & 0.1423
      & 0.2849\% \\
       8 & 64 & 256
     & 17.9079
      & 0.2230
      & 0.1883
      & 0.3694\% \\
       16 & 64 & 256
    & 17.9349
      & 0.2352
      & 0.2266
      & 0.2189\% \\
      \midrule
      8 & 8 & 128
      & 18.0477
      & 0.1971
      & 0.1656
      & 0.4084\% \\
       8 & 16 & 128
     & 17.9807
      & 0.1749
      & 0.1360
      & 0.0353\% \\
       8 & 32 & 128
    & 17.9476
      & 0.1815
      & 0.1540
      & 0.1487\% \\
       8 & 40 & 128
    & 17.9505
      & 0.1625
      & 0.1291
      & 0.1322\% \\
       8 & 48 & 128
    & 17.9629
      & 0.1793
      & 0.1367
      & 0.0635\% \\
      8 & 56 & 128
    & 17.9387
      & 0.1774
      & 0.1336
      & 0.1981\% \\
      \midrule
       8 & 16 & 512
      & 17.9856
      & 0.2041
      & 0.1572
      & 0.0634\% \\
       8 & 16 & 768
     & 17.9940
      & 0.2029
      & 0.1636
      & 0.1096\% \\
       8 & 16 & 1024
    & 17.9716
      & 0.2640
      & 0.2260
      & 0.0153\% \\
      \bottomrule
    \end{tabular*}
  \end{threeparttable}
\end{table}

\begin{table}[!htb]
  \centering
  \captionsetup{skip=4pt}
  \begin{threeparttable}
    \caption{Numerical results of the MHDBR ($J=1$) scheme for
    Example~2 with $d=100$.}
    \label{tab:ex2-mhdbr-j1-d100}
    \setlength{\tabcolsep}{8pt}
  \renewcommand{\arraystretch}{0.90}
    \begin{tabular*}{\linewidth}
      {@{\extracolsep{\fill}}ccccccc@{}}
      \toprule
     $N$ & $K$ & $M$
      & $\overline{\mathfrak{u}}(0,x_0)$
      & $\operatorname{Std.dev.}$
      & $\operatorname{MAE.}$
      & $\operatorname{Rel.err.}$ \\
      \midrule
       4 & 64 & 256
      & 21.1309
      & 0.0741
      & 0.1679
      & 0.7883\% \\
       8 & 64 & 256
     & 21.0677
      & 0.0677
      & 0.2311
      & 1.0850\% \\
       16 & 64 & 256
    & 21.0329
      & 0.0880
      & 0.2659
      & 1.2483\% \\
       \midrule
      8 & 8 & 128
      & 21.1250
      & 0.1535
      & 0.1931
      & 0.8162\% \\
       8 & 16 & 128
     & 21.0950
      & 0.1230
      & 0.2038
      & 0.9568\% \\
       8 & 32 & 128
    & 21.1305
      & 0.1155
      & 0.1712
      & 0.7900\% \\
       8 & 40 & 128
    & 21.1442
      & 0.1115
      & 0.1546
      & 0.7257\% \\
       8 & 48 & 128
    & 21.1612
      & 0.1014
      & 0.1512
      & 0.6459\% \\
      8 & 56 & 128
    & 21.1579
      & 0.0940
      & 0.1438
      & 0.6613\% \\
      \midrule
       8 & 16 & 512
      & 21.0943
      & 0.2364
      & 0.2556
      & 0.9599\% \\
       8 & 16 & 768
     & 21.1153
      & 0.2465
      & 0.2577
      & 0.8614\% \\
       8 & 16 & 1024
    & 21.2162
      & 0.1616
      & 0.1461
      & 0.3880\% \\
      \bottomrule
    \end{tabular*}
  \end{threeparttable}
\end{table}
\textbf{Example 2.} We consider a financial model. This illustrates a classic nonlinear market framework featuring a dual-rate structure for borrowing and lending (\cite{EWHMJAKT19,KLTL25}). The model is
\begin{equation}
\label{PDEdetail-2}
\begin{cases}
\displaystyle
\frac{\partial u}{\partial t}(t,x)
+ \frac{\sigma^2}{2}\sum_{\ell=1}^{d}x^2_\ell \frac{\partial^2 u}{\partial x^2_\ell}(t,x)
- \min\left\{R_2\left(u - \sum_{\ell=1}^{d}x_\ell \frac{\partial u}{\partial x_\ell}(t,x)\right), 
R_1\left(u - \sum_{\ell=1}^{d}x_\ell \frac{\partial u}{\partial x_\ell}(t,x)\right)\right\}
= 0, \\
\displaystyle
u(T,x)=\max\left\{\max_{1\le \ell\le d}(x_\ell-K_1),\,0\right\}
-2\max\left\{\max_{1\le \ell\le d}(x_\ell-K_2),\,0\right\},
\end{cases}
\end{equation}
where \(R_1\) and \(R_2\) are the lending and borrowing rates, respectively.
Following \cite{KLTL25}, the benchmark is formulated in logarithmic
coordinates. To specify the correspondence, let
$q_\ell=\log x_\ell$ and $v(t,q)=u(t,e^{q_1},\ldots,e^{q_d})$.
Then $x_\ell u_{x_\ell}=v_{q_\ell}$ and
$x_\ell^2u_{x_\ell x_\ell}=v_{q_\ell q_\ell}-v_{q_\ell}$, so that
\begin{equation}\label{ex2-log-pde}
\begin{aligned}
 0={}&v_t+\frac{\sigma^2}{2}\Delta v
             -\frac{\sigma^2}{2}\sum_{\ell=1}^d v_{q_\ell}\\
 &-\min\left\{
 R_2\left(v-\sum_{\ell=1}^d v_{q_\ell}\right),
 R_1\left(v-\sum_{\ell=1}^d v_{q_\ell}\right)\right\}.
\end{aligned}
\end{equation}
With the auxiliary drift $\mu$, the equivalent forward process and
terminal function are
\begin{equation}\label{ex2-log-forward}
\begin{aligned}
 dQ_t&=(\mu-\sigma^2/2)\mathbf1_d\,dt+\sigma\,dW_t,
 \qquad Q_0=(\log100)\mathbf1_d,\\
 \widetilde g(q)
 &=\max\{\max_\ell(e^{q_\ell}-K_1),0\}\\
 &\quad-2\max\{\max_\ell(e^{q_\ell}-K_2),0\}.
\end{aligned}
\end{equation}
Writing $s(z)=\sigma^{-1}\sum_{\ell=1}^dz_\ell$, the compensating
BSDE generator is
\begin{equation}\label{ex2-log-generator}
 \widetilde f(y,z)
 =-\mu s(z)-\min\{R_2(y-s(z)),R_1(y-s(z))\}.
\end{equation}
The first term in \eqref{ex2-log-generator} cancels the auxiliary
forward drift and is required for equivalence with
\eqref{PDEdetail-2}. These formulas state the mathematical convention
needed to interpret a logarithmic implementation; the aggregate
benchmark records do not independently verify its implementation.

This example is outside the regularity regime of Assumption 2:
the terminal payoff is unbounded and nonsmooth, whereas
$\mathcal A_b^k$ includes boundedness of the function itself and
regularity up to the terminal time. In logarithmic variables,
$\widetilde g$ also has exponential rather than linear growth.
Moreover, $\widetilde f$ depends on $z$, excluding the quantitative
GroupSort theorem. The benchmark is therefore an empirical test with
a nonsmooth payoff.

We consider $d=50$ and $d=100$, while keeping $T=0.5$, 
$x_0=100\mathbf{1}_d$, 
$\mu=0.06$, 
$\sigma=0.2$,
and
$
K_1=120$, $K_2=150$,
$R_1=0.04$, $R_2=0.06$.
Since no closed-form analytic solution is available, we use the
benchmark values reported in \cite{EWHMJAKT19}, which were obtained
using a multilevel Monte Carlo method with seven Picard iterations.
More precisely, the numerical reference values are
$u_{\mathrm{ref}}(0,x_0)=17.9743$ for $d=50$ and
$u_{\mathrm{ref}}(0,x_0)=21.2988$ for $d=100$.
They are not analytic exact values; all reported deviations in this
example are measured against these references.

Tables~\ref{tab:ex2-mhdbr-j1-d50} and
\ref{tab:ex2-mhdbr-j1-d100} report the $J=1$ scheme in dimensions
$50$ and $100$. Most relative deviations of the sample mean from
the reference are below $1\%$, although the errors are not monotone
in $N$, $K$, or $M$. These observations describe the listed
configurations; they do not establish a high-order convergence rate,
dimension-uniform accuracy, or a stability theorem.

\begin{table}[!htb]
  \centering
  \captionsetup{skip=4pt}
  \caption{Numerical results of different schemes for
  Example~3 with different dimensions.}
  \label{tab:ex3-mhdbr-j1-dimensions}

  \setlength{\tabcolsep}{8pt}
  \renewcommand{\arraystretch}{0.90}

  \begin{tabular*}{\linewidth}{@{\extracolsep{\fill}}ccccccc@{}}
    \toprule
      Scheme
      & $d$
      & $u(0,x_0)$
      & $\overline{\mathfrak{u}}(0,x_0)$
      & $\operatorname{Std.dev.}$
      & $\operatorname{MAE.}$
      & $\operatorname{Rel.err.}$ \\
      \midrule

      \multirow{5}{*}{MHDBR $(J=1)$}
      & 2   & 1.542098   & 1.549792   & 0.012676 & 0.011903 & 0.4989\% \\
      & 10  & 7.710488   & 7.684636   & 0.090642 & 0.080368 & 0.3353\% \\
      & 50  & 38.552441  & 38.708513  & 0.601159 & 0.487889 & 0.4048\% \\
      & 100 & 77.104881  & 76.583385  & 0.775599 & 0.772751 & 0.6763\% \\
      & 150 & 115.657326 & 111.881879 & 0.801596 & 3.775446 & 3.2643\% \\
      \midrule

      \multirow{5}{*}{MHDBR $(J=2)$}
      & 2   & 1.542098   & 1.542060   & 0.023527 & 0.019246 & 0.0024\% \\
      & 10  & 7.710488   & 7.655181   & 0.083856 & 0.075393 & 0.7173\% \\
      & 50  & 38.552441  & 38.603999  & 0.451697 & 0.358061 & 0.1337\% \\
      & 100 & 77.104881  & 76.777737  & 0.924413 & 0.704337 & 0.4243\% \\
      & 150 & 115.657326 & 110.476669 & 1.019467 & 5.180656 & 4.4793\% \\
      \midrule

      \multirow{5}{*}{MHDBR $(J=3)$}
      & 2   & 1.542098   & 1.542756   & 0.023197 & 0.018029 & 0.0427\% \\
      & 10  & 7.710488   & 7.710221   & 0.089625 & 0.069966 & 0.0035\% \\
      & 50  & 38.552441  & 38.579995  & 0.496429 & 0.308706 & 0.0715\% \\
      & 100 & 77.104881  & 76.965522  & 0.676417 & 0.517965 & 0.1807\% \\
      & 150 & 115.657326 & 109.892693 & 0.909541 & 5.764632 & 4.9842\% \\
      \midrule

      \multirow{5}{*}{MHDBR $(J=4)$}
      & 2   & 1.542098   & 1.540237   & 0.024774 & 0.020762 & 0.1206\% \\
      & 10  & 7.710488   & 7.723910   & 0.075118 & 0.062233 & 0.1741\% \\
      & 50  & 38.552441  & 38.430746  & 0.372579 & 0.285289 & 0.3157\% \\
      & 100 & 77.104881  & 76.826250  & 0.374783 & 0.380183 & 0.3614\% \\
      & 150 & 115.657326 & 110.061488 & 0.935572 & 5.595838 & 4.8382\% \\
      \midrule

      \multirow{5}{*}{MDBDP \cite{GMPHWX22}}
      & 2   & 1.542098   & 1.539776   & 0.004799 & 0.003827 & 0.1506\% \\
      & 10  & 7.710488   & 7.713663   & 0.007350 & 0.005833 & 0.0412\% \\
      & 50  & 38.552441  & 38.473212  & 0.016998 & 0.079228 & 0.2055\% \\
      & 100 & 77.104881  & 76.555768  & 0.125841 & 0.549113 & 0.7122\% \\
      & 150 & 115.657326 & 112.471697 & 0.319760 & 3.185629 & 2.7544\% \\
      \bottomrule
    \end{tabular*}
\end{table}

\textbf{Example 3.} 
We consider a linear benchmark associated with the
Black--Scholes--Barenblatt model, which arises in option
pricing and hedging under volatility uncertainty
\cite{RM24}. With the fixed volatility used here,
the PDE is linear and admits an explicit solution,
providing a benchmark for evaluating the accuracy
of the MHDBR schemes. The model reads
\begin{equation}
\label{PDEdetail-3}
\begin{cases}
\frac{\partial u}{\partial t}(t,x)
+\frac{\sigma^2}{2} \operatorname{Tr}\left(\operatorname{diag}(x^2) D_x^2 u(t,x)\right)
-  r\left(u - x D_x u(t,x)\right)
= 0, \\
u(T,x)=|x|^2,
\end{cases}
\end{equation}
the  analytic solution of the PDE (\ref{PDEdetail-3}) is explicitly given by
$
u(t,x)
=\exp\!\left((r + \sigma^2)(T-t)\right)|x|^2.
$

The stated solution satisfies the PDE: its time derivative is
$-(r+\sigma^2)u$, its diffusion term is $\sigma^2u$, and its remaining
drift and reaction terms sum to $ru$.
However, $g(x)=|x|^2$ violates the linear-growth condition in
Assumption 1(iv) and the global Lipschitz condition on $g$ in
Theorem~\ref{Thm-GS-convergence}. The solution is unbounded and hence
does not belong to $\mathcal A_b^{J+2}$. This is therefore a smooth
but unbounded benchmark beyond the present theory.

We use $T=1$, $r=0.05$, $\sigma=0.4$ and
$X_0=(1,0.5,\ldots,1,0.5)\in\mathbb R^d$.
Table~\ref{tab:ex3-mhdbr-j1-dimensions} uses the fixed values
$N=20$, $K=64$, and $M=256$ in dimensions $2$ through $150$.
The comparison depends on the error statistic: MDBDP has a smaller
MAE than every listed MHDBR variant at $d=2,10,50,150$, while some
MHDBR variants have a smaller MAE at $d=100$. Several MHDBR
configurations have a smaller deviation of the sample mean despite a
larger MAE. For instance, at $d=2$, $J=2$ gives a relative
sample-mean deviation of $0.0024\%$, but its relative MAE is about
$1.25\%$. Thus these results do not show a uniform accuracy advantage
over MDBDP. The larger errors at $d=150$ also limit conclusions about
dimension-independent performance under a fixed budget.

\begin{figure}[!htb]
  \centering
  \begin{minipage}{0.6\textwidth}
\centering

  \begin{subfigure}[t]{0.485\linewidth}
    \centering
    \includegraphics[width=\linewidth]
    {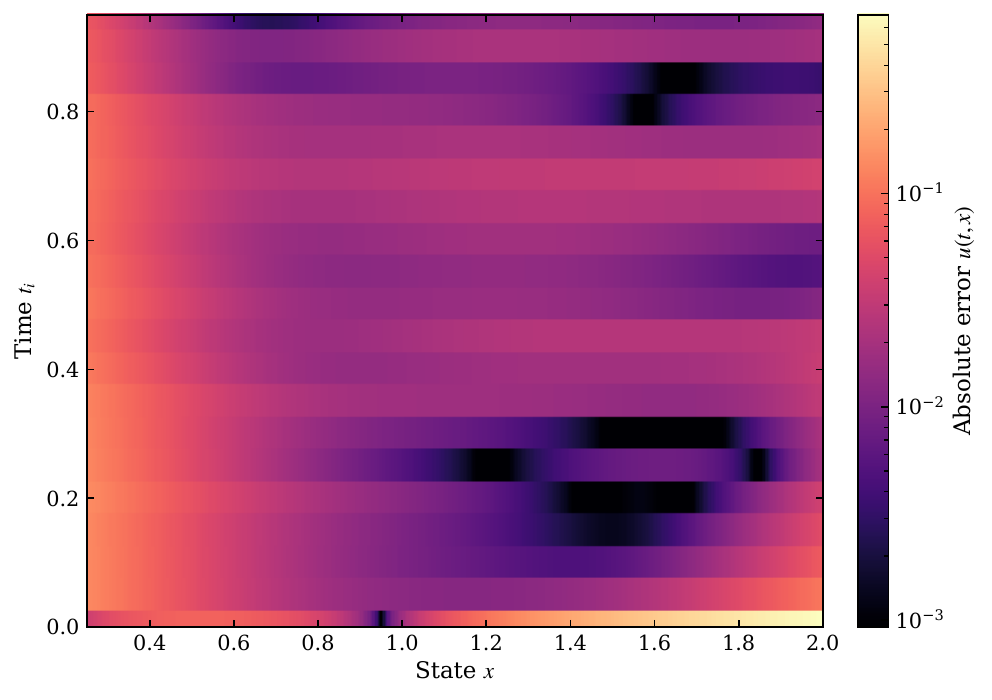}
    \caption{$J=1$.}
    \label{fig:ex3-j1-heatmap}
  \end{subfigure}
  \hfill
  \begin{subfigure}[t]{0.485\linewidth}
    \centering
    \includegraphics[width=\linewidth]
    {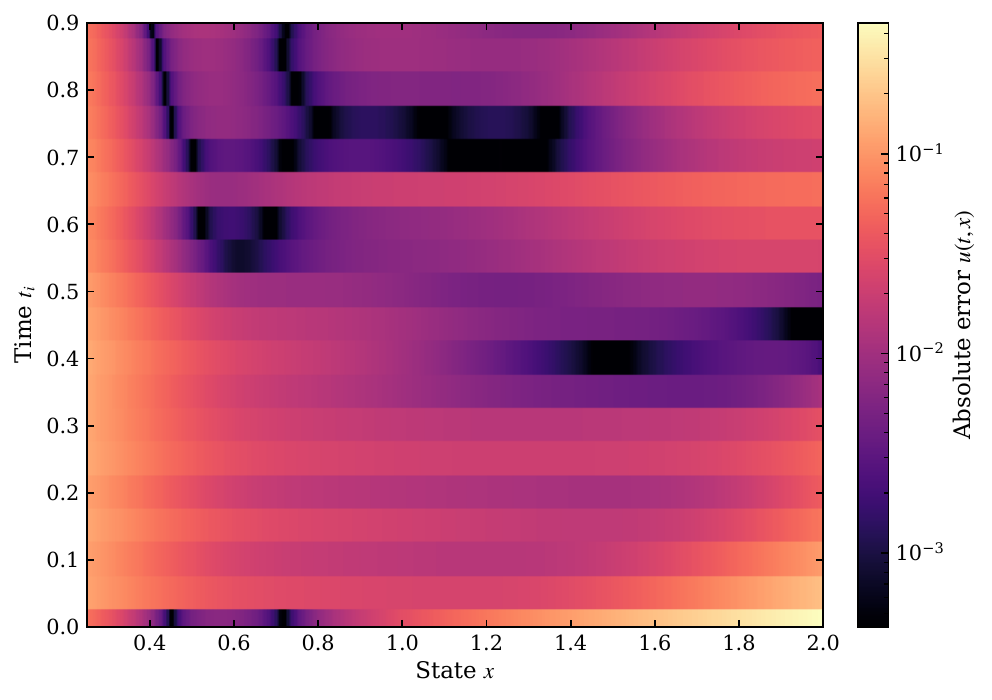}
    \caption{$J=2$.}
    \label{fig:ex3-j2-heatmap}
  \end{subfigure}

  \vspace{0.3em}

  \begin{subfigure}[t]{0.485\linewidth}
    \centering
    \includegraphics[width=\linewidth]
    {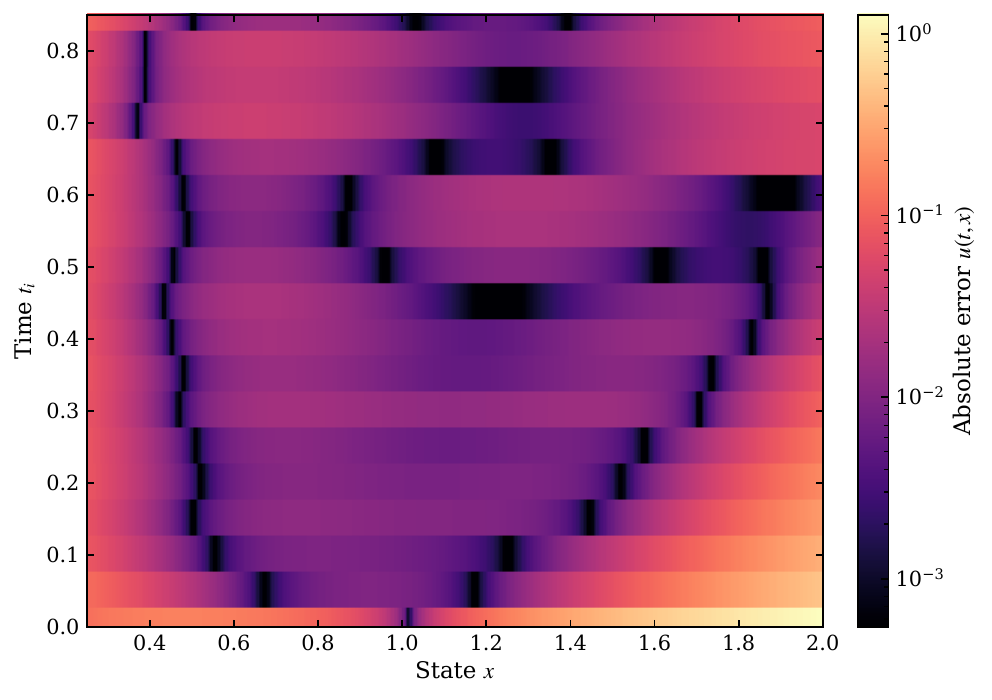}
    \caption{$J=3$.}
    \label{fig:ex3-j3-heatmap}
  \end{subfigure}
  \hfill
  \begin{subfigure}[t]{0.485\linewidth}
    \centering
    \includegraphics[width=\linewidth]
    {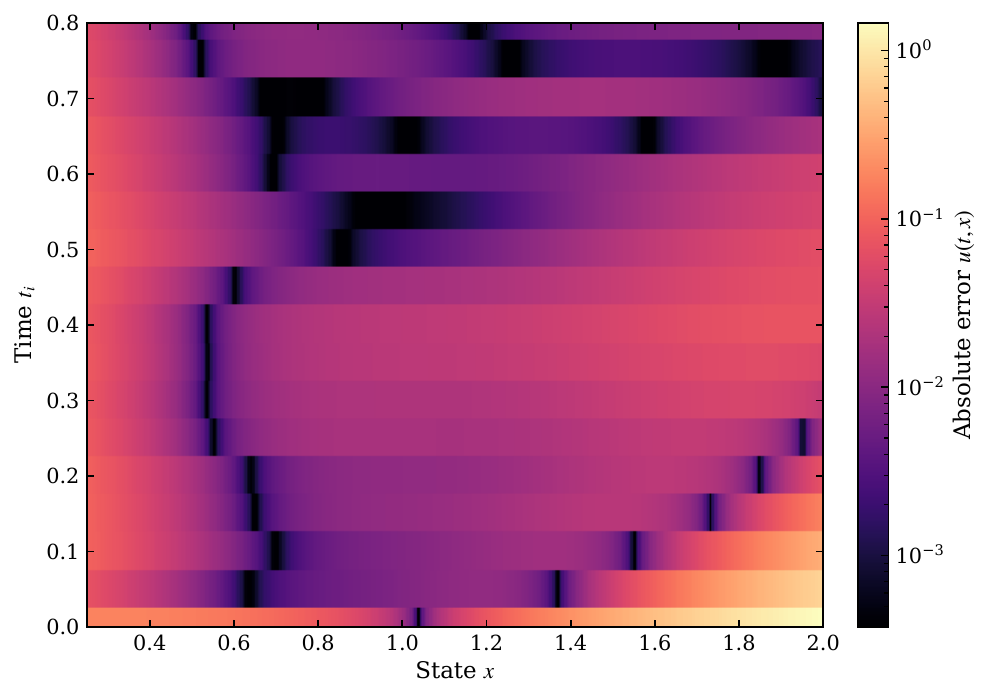}
    \caption{$J=4$.}
    \label{fig:ex3-j4-heatmap}
  \end{subfigure}
  \end{minipage}

  \caption{
    Time--state distributions of the absolute error
for the one-dimensional case ($d=1$) of Example~3,
computed by the MHDBR schemes with $J=1,2,3,4$.
  }
  \label{fig:ex3-error-heatmaps}
\end{figure}
\begin{figure}[!htb]
  \centering
  \begin{minipage}{0.6\textwidth}
\centering

  \begin{subfigure}[t]{0.485\linewidth}
    \centering
    \includegraphics[width=\linewidth]
    {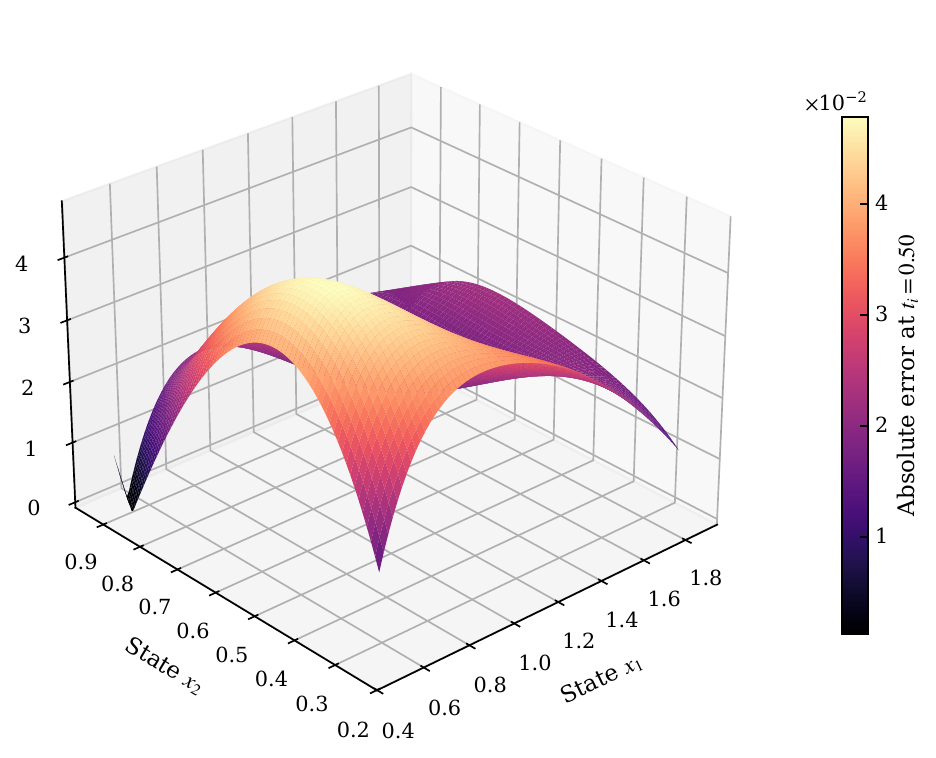}
    \caption{$J=1$.}
    \label{fig:ex3-j1-surface}
  \end{subfigure}
  \hfill
  \begin{subfigure}[t]{0.485\linewidth}
    \centering
    \includegraphics[width=\linewidth]
    {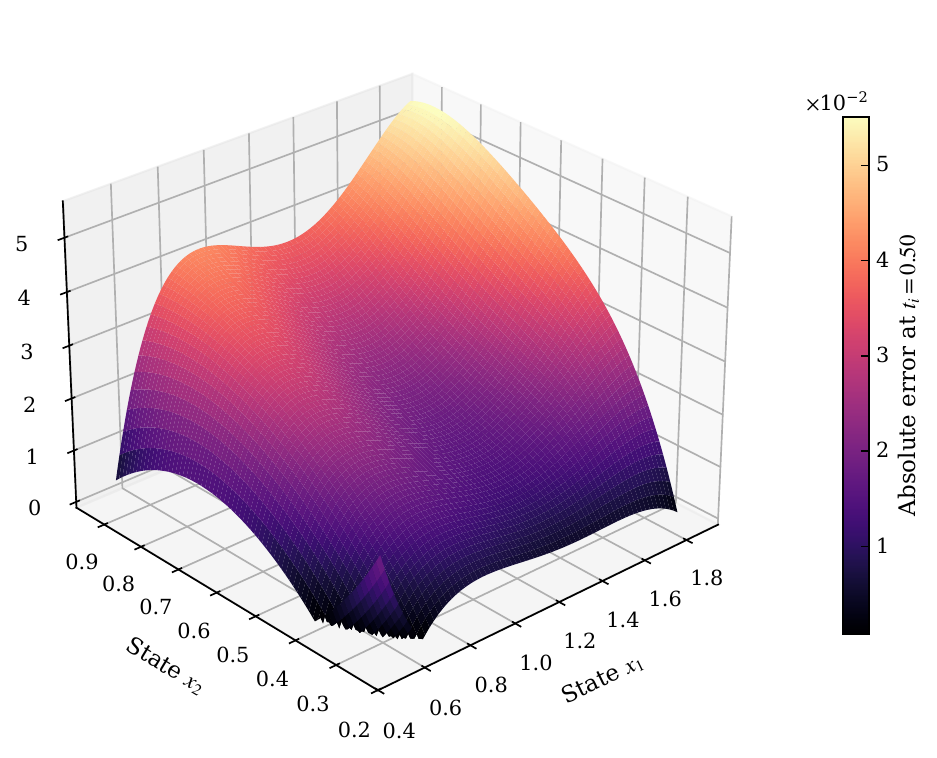}
    \caption{$J=2$.}
    \label{fig:ex3-j2-surface}
  \end{subfigure}

  \vspace{0.3em}

  \begin{subfigure}[t]{0.485\linewidth}
    \centering
    \includegraphics[width=\linewidth]
    {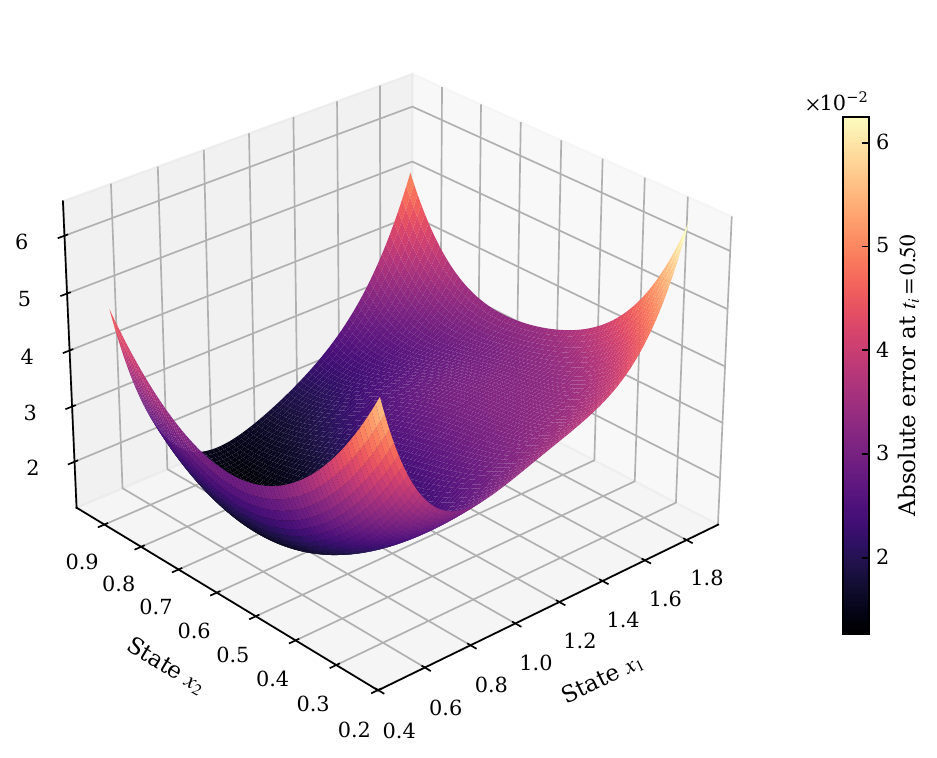}
    \caption{$J=3$.}
    \label{fig:ex3-j3-surface}
  \end{subfigure}
  \hfill
  \begin{subfigure}[t]{0.485\linewidth}
    \centering
    \includegraphics[width=\linewidth]
    {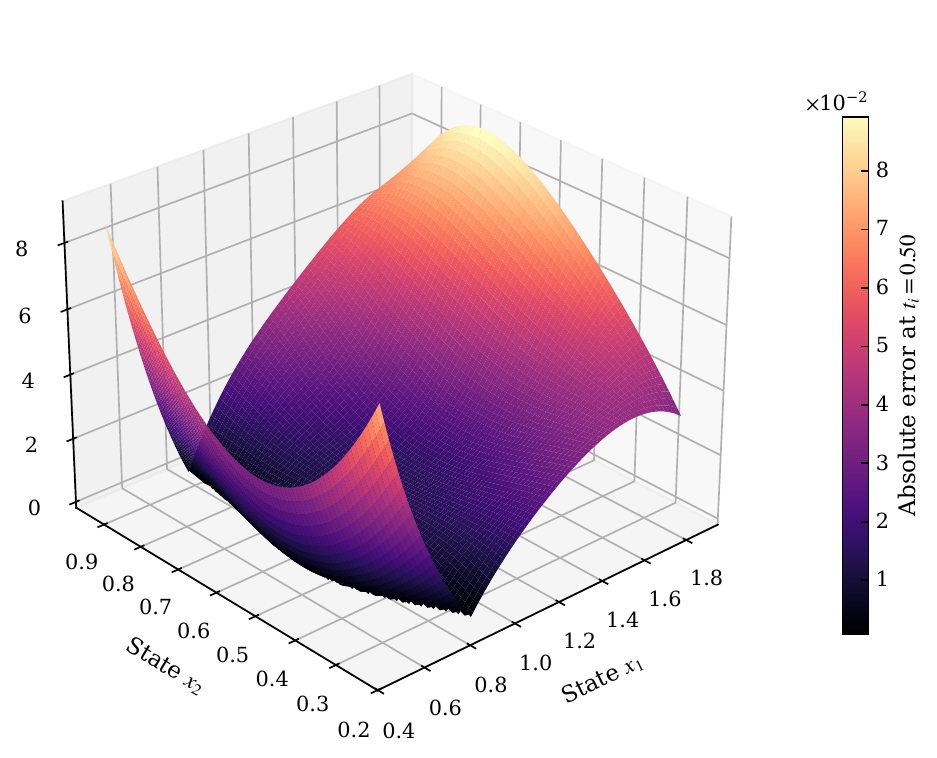}
    \caption{$J=4$.}
    \label{fig:ex3-j4-surface}
  \end{subfigure}
  \end{minipage}

  \caption{
    Surface plots of the absolute error
over the two-dimensional state space ($d=2$) for Example~3,
computed by the MHDBR schemes with $J=1,2,3,4$.
  }
  \label{fig:ex3-error-surface}
\end{figure}

Figures~\ref{fig:ex3-error-heatmaps} and
\ref{fig:ex3-error-surface} display the pointwise errors on the
selected one- and two-dimensional regions. Their appearance is
consistent with the reported finite-resolution accuracy, but does
not replace estimates in the space--time norms of the theorems.

\subsection{Independent verification of the time-discrete scheme}
\label{sec:manufactured-time-check}

To separate temporal discretization from learning and sampling errors,
consider the one-dimensional manufactured problem
\begin{equation}\label{manufactured-pde}
 u_t+\tfrac12u_{xx}+u=0,\qquad u(1,x)=\sin x,
 \qquad u(t,x)=e^{(1-t)/2}\sin x.
\end{equation}
Here $X_t=\pi/4+W_t$, $f(t,x,y)=y$, and
$Z_t=e^{(1-t)/2}\cos X_t$.
The data satisfy the coefficient and solution regularity conditions
of Assumptions 1--3: the forward process is simulated exactly,
$g$ is globally Lipschitz, and $u$ and $F_u=u$ have bounded Lie
derivatives of every required order. This test uses no neural
network, Monte Carlo sampling, or GPU training, and does not test
the population minimization or learning-error parts of the theory.

For $h=1/N$, analytic conditional expectations give
\begin{equation}\label{manufactured-expectations}
\begin{aligned}
 \mathbb E_i[\sin X_{t_{i+j}}]
   &=e^{-jh/2}\sin X_{t_i},\\
 \mathbb E_i[\sin X_{t_{i+j}}\Delta W_{i,i+j}]
   &=jh\,e^{-jh/2}\cos X_{t_i}.
\end{aligned}
\end{equation}
Prescribe exact starting values at $i=N-J+1,\ldots,N$ and write
$Y_i^\pi=A_i\sin X_{t_i}$ and $Z_i^\pi=B_i\cos X_{t_i}$.
The multistep scheme then reduces to the scalar recurrence
\begin{equation}\label{manufactured-recurrence}
\begin{aligned}
 A_i&=\frac{e^{-h/2}A_{i+1}
       +h\sum_{j=1}^J\alpha_j e^{-jh/2}A_{i+j}}
                {1-h\alpha_0},\\
 B_i&=\sum_{j=1}^J(h\beta_{J,j})\,j e^{-jh/2}A_{i+j},
 \qquad i=N-J,\ldots,0.
\end{aligned}
\end{equation}
The exact coefficients are $A_i^{\mathrm{ex}}=e^{(1-t_i)/2}$.
Since $X_0=\pi/4$,
$\mathbb E\sin^2X_{t_i}=\mathbb E\cos^2X_{t_i}=1/2$.
Thus the errors in the nodal norms are computed without sampling as
\begin{equation}\label{manufactured-errors}
\begin{aligned}
 E_Y(N)&=\max_{0\leq i\leq N-J}
        \|Y_{t_i}-Y_i^\pi\|_{L^2}\\
       &=\frac1{\sqrt2}\max_{0\leq i\leq N-J}
                         |A_i^{\mathrm{ex}}-A_i|,\\
 E_Z(N)&=\left(h\sum_{i=0}^{N-J}
                 \|Z_{t_i}-Z_i^\pi\|_{L^2}^2\right)^{1/2}\\
       &=\left(\frac h2\sum_{i=0}^{N-J}
                    |A_i^{\mathrm{ex}}-B_i|^2\right)^{1/2}.
\end{aligned}
\end{equation}
The recurrence was evaluated with 60-digit decimal arithmetic.
Table~\ref{tab:manufactured-time-check} reports the resulting errors
and the observed order $p_{64}=\log_2(E_Z(32)/E_Z(64))$.

\begin{table}[!htbp]
\centering
\small
\setlength{\tabcolsep}{4pt}
\caption{Independent time-discretization check with exact starting
values and analytic conditional expectations; no network training or
Monte Carlo sampling.}
\label{tab:manufactured-time-check}
\begin{tabular}{ccccccc}
\toprule
$J$ & $E_Y(64)$ & $E_Z(8)$ & $E_Z(16)$ & $E_Z(32)$ & $E_Z(64)$ & $p_{64}$\\
\midrule
1 & $2.3720\times10^{-5}$ & $1.1176\times10^{-1}$ & $5.6879\times10^{-2}$ & $2.8699\times10^{-2}$ & $1.4416\times10^{-2}$ & 0.993 \\
2 & $1.8136\times10^{-7}$ & $1.2641\times10^{-2}$ & $3.3851\times10^{-3}$ & $8.7539\times10^{-4}$ & $2.2255\times10^{-4}$ & 1.976 \\
3 & $1.7546\times10^{-9}$ & $1.4146\times10^{-3}$ & $2.0091\times10^{-4}$ & $2.6678\times10^{-5}$ & $3.4345\times10^{-6}$ & 2.957 \\
4 & $1.9032\times10^{-11}$ & $1.5610\times10^{-4}$ & $1.1892\times10^{-5}$ & $8.1248\times10^{-7}$ & $5.2994\times10^{-8}$ & 3.938 \\
\bottomrule
\end{tabular}
\end{table}

The observed orders of the nodal $Z$ error approach $J$ for
$J=1,2,3,4$. This independent calculation supports the temporal
consistency of the underlying multistep formula on this smooth
problem. It neither establishes a convergence rate for the trained
MHDBR implementation nor changes the scope of the preceding
finite-budget benchmark comparisons.

\section{Conclusion}

We have developed a multistep neural backward regression method for
semilinear parabolic PDEs. Under the stated assumptions, its population
approximation error separates into time discretization, neural
approximation and conditional Monte Carlo terms. For $z$-independent
generators, the constrained GroupSort construction yields nodal
$L^2$ order $J$ with a strong forward order at least $J$, sufficiently
accurate starting values and the prescribed network and sample sizes.
This is a conditional approximation theorem: it neither establishes
finite-sample training convergence nor removes dimensional dependence
from the network complexity.

The manufactured test verifies the temporal order of the multistep
scheme using exact conditional expectations. The three neural-network
benchmarks illustrate behavior in broader settings, including problems
outside the global assumptions; their reported errors do not establish
the full convergence theorem. Quantifying optimization and
finite-training-sample errors, and extending the theory to less regular
or nonglobally Lipschitz data, remain separate tasks.

\begin{appendices}

\section{Multidimensional It\^{o}--Taylor discretization schemes}
\label{app:Ito-Taylor-schemes}

This appendix presents several particular realizations of the general
forward discretization \eqref{SDE-general-scheme}. Let
\[
0=t_0<t_1<\cdots<t_N=T,
\qquad
h=t_{i+1}-t_i,
\]
be a uniform temporal partition. We write
\[
\Delta W_i^\ell
=
W_{t_{i+1}}^\ell-W_{t_i}^\ell,
\qquad \ell=1,\ldots,d,
\]
and denote by \(\sigma^{(\ell)}\) the \(\ell\)-th column of the diffusion
matrix \(\sigma\).

For a sufficiently smooth scalar- or vector-valued function
\(\varphi:[0,T]\times\mathbb{R}^d\to\mathbb{R}^q\), define
\[
L^0\varphi
=
\frac{\partial\varphi}{\partial t}
+
\sum_{a=1}^d
\mu^a\frac{\partial\varphi}{\partial x_a}
+
\frac{1}{2}
\sum_{\ell=1}^d\sum_{a,b=1}^d
\sigma^{a\ell}\sigma^{b\ell}
\frac{\partial^2\varphi}
{\partial x_a\partial x_b},
\]
and, for \(\ell=1,\ldots,d\),
\[
L^\ell\varphi
=
\sum_{a=1}^d
\sigma^{a\ell}
\frac{\partial\varphi}{\partial x_a}.
\]
For vector-valued functions, these operators are applied componentwise,
and compositions are denoted by
\[
L^rL^s\varphi=L^r\bigl(L^s\varphi\bigr).
\]

Let
\[
\mathrm{d}\mathcal{W}_s^0=\mathrm{d}s,
\qquad
\mathrm{d}\mathcal{W}_s^\ell=\mathrm{d}W_s^\ell,
\quad \ell=1,\ldots,d.
\]
For a multi-index
\[
\alpha=(j_1,\ldots,j_k)
\in\{0,1,\ldots,d\}^k,
\]
the corresponding iterated It\^{o} integral is defined by
\[
I_\alpha^i
=
I_{(j_1,\ldots,j_k)}^i
:=
\int_{t_i}^{t_{i+1}}
\int_{t_i}^{s_k}\cdots
\int_{t_i}^{s_2}
\mathrm{d}\mathcal{W}_{s_1}^{j_1}
\cdots
\mathrm{d}\mathcal{W}_{s_k}^{j_k}.
\]
In particular,
\[
I_{(0)}^i=h,
\qquad
I_{(\ell)}^i=\Delta W_i^\ell,
\qquad
I_{(0,0)}^i=\frac{1}{2}h^2.
\]
We also introduce
\[
\Delta\zeta_i^\ell
:=
I_{(\ell,0)}^i
=
\int_{t_i}^{t_{i+1}}
\int_{t_i}^{s}
\mathrm{d}W_r^\ell\,\mathrm{d}s,
\]
so that
\[
I_{(0,\ell)}^i
=
h\Delta W_i^\ell-\Delta\zeta_i^\ell.
\]
Moreover,
\[
\mathbb{E}
\left[
\left(\Delta\zeta_i^\ell\right)^2
\right]
=
\frac{1}{3}h^3,
\qquad
\mathbb{E}
\left[
\Delta W_i^r\Delta\zeta_i^\ell
\right]
=
\frac{1}{2}\delta_{r\ell}h^2,
\]
where \(\delta_{r\ell}\) denotes the Kronecker delta. Unless otherwise
specified, all coefficients and their derivatives below are evaluated
at \((t_i,X_i^\pi)\).

The general formulation \eqref{SDE-general-scheme} includes the
following classical It\^{o}--Taylor-type schemes.

\begin{enumerate}

\item[$\bullet$]
\textbf{The multidimensional Euler scheme.}
It is given by
\begin{equation}
X_{i+1}^{\pi}
=
X_i^\pi
+
\mu(t_i,X_i^\pi)h
+
\sum_{\ell=1}^d
\sigma^{(\ell)}(t_i,X_i^\pi)\Delta W_i^\ell.
\label{app-SDE-euler}
\end{equation}
Equivalently,
\[
X_{i+1}^{\pi}
=
X_i^\pi
+
\mu(t_i,X_i^\pi)h
+
\sigma(t_i,X_i^\pi)\Delta W_i.
\]

\item[$\bullet$]
\textbf{The multidimensional Milstein scheme.}
It takes the form
\begin{equation}
\begin{aligned}
X_{i+1}^{\pi}
={}&
X_i^\pi
+
\mu(t_i,X_i^\pi)h
+
\sum_{\ell=1}^d
\sigma^{(\ell)}(t_i,X_i^\pi)\Delta W_i^\ell
\\
&+
\sum_{r=1}^d\sum_{\ell=1}^d
L^r\sigma^{(\ell)}(t_i,X_i^\pi)
I_{(r,\ell)}^i,
\end{aligned}
\label{app-SDE-milstein}
\end{equation}
where
\[
I_{(r,\ell)}^i
=
\int_{t_i}^{t_{i+1}}
\int_{t_i}^{s}
\mathrm{d}W_v^r\,\mathrm{d}W_s^\ell.
\]
For repeated indices,
\[
I_{(\ell,\ell)}^i
=
\frac{1}{2}
\left[
\left(\Delta W_i^\ell\right)^2-h
\right].
\]

\item[$\bullet$]
\textbf{The multidimensional strong-order \(1.5\)
It\^{o}--Taylor scheme.}
Define
\begin{equation}
\begin{aligned}
\Psi_i^{[1.5]}
={}&
\mu(t_i,X_i^\pi)h
+
\sum_{\ell=1}^d
\sigma^{(\ell)}(t_i,X_i^\pi)\Delta W_i^\ell
+
\frac{1}{2}
L^0\mu(t_i,X_i^\pi)h^2
\\
&+
\sum_{r=1}^d
L^r\mu(t_i,X_i^\pi)\Delta\zeta_i^r
+
\sum_{\ell=1}^d
L^0\sigma^{(\ell)}(t_i,X_i^\pi)
\left(
h\Delta W_i^\ell-\Delta\zeta_i^\ell
\right)
\\
&+
\sum_{r=1}^d\sum_{\ell=1}^d
L^r\sigma^{(\ell)}(t_i,X_i^\pi)
I_{(r,\ell)}^i
+
\sum_{r=1}^d\sum_{s=1}^d\sum_{\ell=1}^d
L^rL^s\sigma^{(\ell)}(t_i,X_i^\pi)
I_{(r,s,\ell)}^i.
\end{aligned}
\label{app-SDE-strong-1.5-increment}
\end{equation}
The corresponding scheme is
\begin{equation}
X_{i+1}^{\pi}
=
X_i^\pi+\Psi_i^{[1.5]}.
\label{app-SDE-strong-1.5}
\end{equation}

\item[$\bullet$]
\textbf{The multidimensional strong-order \(2.0\)
It\^{o}--Taylor scheme.}
Using \(\Psi_i^{[1.5]}\) defined in
\eqref{app-SDE-strong-1.5-increment}, the scheme is
\begin{equation}
\begin{aligned}
X_{i+1}^{\pi}
={}&
X_i^\pi+\Psi_i^{[1.5]}
+
\sum_{r=1}^d\sum_{s=1}^d
L^rL^s\mu(t_i,X_i^\pi)
I_{(r,s,0)}^i
+
\sum_{r=1}^d\sum_{\ell=1}^d
L^rL^0\sigma^{(\ell)}(t_i,X_i^\pi)
I_{(r,0,\ell)}^i
\\
&+
\sum_{r=1}^d\sum_{\ell=1}^d
L^0L^r\sigma^{(\ell)}(t_i,X_i^\pi)
I_{(0,r,\ell)}^i
+
\sum_{r=1}^d\sum_{s=1}^d
\sum_{q=1}^d\sum_{\ell=1}^d
L^rL^sL^q\sigma^{(\ell)}(t_i,X_i^\pi)
I_{(r,s,q,\ell)}^i.
\end{aligned}
\label{app-SDE-strong-2.0}
\end{equation}

\end{enumerate}

For repeated indices, the iterated stochastic integrals satisfy
\[
I_{(\ell,\ell,\ell)}^i
=
\frac{1}{6}
\left[
\left(\Delta W_i^\ell\right)^3
-
3h\Delta W_i^\ell
\right],
\]
and
\[
I_{(\ell,\ell,\ell,\ell)}^i
=
\frac{1}{24}
\left[
\left(\Delta W_i^\ell\right)^4
-
6h\left(\Delta W_i^\ell\right)^2
+
3h^2
\right].
\]
Consequently, when \(d=1\), the multidimensional Milstein and
higher-order It\^{o}--Taylor schemes reduce to their corresponding
one-dimensional forms.

\section{GroupSort neural networks}
\label{GroupSort-neural-networks}

We use componentwise Lipschitz constraints for vector-valued networks.
For a vector or a matrix $a$, let $\|a\|_\infty$ denote the largest
absolute value of its entries. The notation $|a|$ continues to denote
the Euclidean norm for vectors and the Frobenius norm for matrices.
In particular,
\[
 \|a\|_\infty\leq |a|\leq\sqrt{d_1}\,\|a\|_\infty
\]
when $a$ has $d_1$ scalar entries. Throughout the GroupSort construction,
input distances are Euclidean. A componentwise $\mathcal K$-Lipschitz
map $\Psi$ therefore satisfies
\[
 \|\Psi(x)-\Psi(x')\|_\infty\leq\mathcal K|x-x'|.
\]

Let $\kappa\geq2$ be an integer dividing every hidden-layer width
$n_\ell$, $\ell=0,\ldots,L-1$. The GroupSort activation
$\zeta_\kappa^\ell$ partitions its input into groups of size $\kappa$
and sorts each group in nonincreasing order. For a fixed bias bound
$M_\beta>0$, define the scalar network class
\[
\begin{aligned}
 \mathcal S_{d_0}^{\zeta_\kappa,L,\mathbf n}
 :=\bigg\{&\varphi(\mathcal W_0,\beta_0,\ldots,
                         \mathcal W_L,\beta_L)
          \in\mathcal{NN}_{d_0,1}^{\zeta_\kappa,L,\mathbf n}:\\
 &\sup_{|x|=1}\|\mathcal W_0x\|_\infty\leq1,\qquad
   \sup_{\|x\|_\infty=1}\|\mathcal W_\ell x\|_\infty\leq1,
                   \quad \ell=1,\ldots,L,\\
 &\|\beta_\ell\|_\infty\leq M_\beta,
                   \quad \ell=0,\ldots,L\bigg\}.
\end{aligned}
\]
GroupSort is nonexpansive in the maximum norm, so each member of
this scalar class is $1$-Lipschitz with respect to its Euclidean input.
For $\mathcal K>0$, set
\[
\begin{aligned}
 \mathcal G_{\mathcal K,d_0,d_1}^{\zeta_\kappa,L,\mathbf n}
 :=\bigg\{\Psi=(\Psi_a)_{a=1}^{d_1}:\;&
   \Psi_a(x)=\mathcal K\bar\beta_a
        \psi_a\!\left(\frac{x+\alpha_a}{\bar\beta_a}\right),\\
 &\psi_a\in\mathcal S_{d_0}^{\zeta_\kappa,L,\mathbf n},\quad
   \alpha_a\in\mathbb R^{d_0},\quad\bar\beta_a>0\bigg\}.
\end{aligned}
\]
Every member of this class is componentwise $\mathcal K$-Lipschitz.
Its Euclidean Lipschitz constant is at most $\sqrt{d_1}\mathcal K$.
Matrix outputs are identified with their scalar entries.
The depth is $L+1$, and $\sum_{\ell=0}^{L-1}n_\ell$ is the total
number of hidden neurons in a scalar component network.

\begin{lemma}[Componentwise GroupSort approximation]
\label{gropusort-est}
Let $\widetilde f:[-R,R]^d\to\mathbb R^{d_1}$ satisfy
\[
 \|\widetilde f(x)-\widetilde f(x')\|_\infty
       \leq\mathcal K|x-x'|,
       \qquad x,x'\in[-R,R]^d.
\]
For every $\varepsilon>0$, there exists
$g\in\mathcal G_{\mathcal K,d,d_1}^{\zeta_\kappa,L,\mathbf n}$
such that
\[
 \sup_{x\in[-R,R]^d}\|\widetilde f(x)-g(x)\|_\infty
       \leq2R\mathcal K\varepsilon,
 \qquad
 \sup_{x\in[-R,R]^d}|\widetilde f(x)-g(x)|
       \leq2\sqrt{d_1}R\mathcal K\varepsilon.
\]
For $d>1$, one can take
\[
 \kappa=\max\!\left\{2,
            \left\lceil\frac{2\sqrt d}{\varepsilon}\right\rceil\right\},
 \qquad L+1=O(d^2),\qquad
 \sum_{\ell=0}^{L-1}n_\ell
       =O\!\left(\left(\frac{2\sqrt d}{\varepsilon}\right)^{d^2-1}\right).
\]
For $d=1$, one can take
\[
 \kappa=\max\{2,\lceil1/\varepsilon\rceil\},\qquad
 L+1=3,\qquad
 \sum_{\ell=0}^{L-1}n_\ell=O(1/\varepsilon).
\]
The size estimates are as $\varepsilon\downarrow0$.
For a vector output, the $d_1$ scalar networks can be run in parallel;
the resulting total hidden-neuron count is multiplied by $d_1$.
This factor is absorbed in the size estimates when the output dimension
is fixed.
\end{lemma}
\begin{proof}
Each scalar component of $\widetilde f$ is $\mathcal K$-Lipschitz.
Apply the scalar case of Proposition~2.1 in \cite{GMPHWX22} to each
component, with the same $R,\mathcal K,\varepsilon$ and architecture.
The resulting $g_a$ are $\mathcal K$-Lipschitz and satisfy
$\sup_{[-R,R]^d}|\widetilde f_a-g_a|\leq2R\mathcal K\varepsilon$.
Combining the components gives both displayed bounds and the stated
network sizes. The lower bound $\kappa\geq2$ does not change the
asymptotic size estimates.
\end{proof}

\section{Proof of Theorem \ref{Thm-GS-convergence}}
\label{Proof-of-Theorem}

\begin{proof}
Write $n_*=N-J$. We use the output maximum norm only for the
Lipschitz estimates; all approximation errors and moment estimates
below use the Euclidean or Frobenius norm $|\cdot|$.
Because $n,d$ are fixed, Assumption~1 implies, after increasing a
constant $L_\infty$ by a factor depending only on $n$,
\[
 \|f(t,x,y)-f(t,x',y')\|_\infty
       \leq L_\infty\bigl(|x-x'|+\|y-y'\|_\infty\bigr).
\]
Here we use the hypothesis that $f$ is independent of $z$.
Take $h$ sufficiently small that $|\alpha_0|hL_\infty<1$.
This is compatible with the small-step requirement in the theorem.

For $i\leq n_*$, denote the optimized spatial network by
\[
 \Psi_{y,i}(x):=
  \mathcal{NN}_{y,i,1+d,n}^{\varrho,L,\mathbf n}
                 ((t_i,x),\theta_{y,i}^*).
\]
For $i>n_*$, let $\Psi_{y,i}$ be the prescribed starting function.
At each fixed time $t_i$, the GroupSort approximation has spatial
input dimension $d$. It is realized in the notation with input
dimension $1+d$ by setting the weight of the time input to zero.

\medskip
\noindent Step 1. Componentwise Lipschitz estimates.
Choose $C_X$ so that \eqref{GS-X-stability} holds with $C_X$ in
place of $C$, and set
\[
 s_h=(1+C_Xh)^{1/2},\qquad
 A_\alpha=\sum_{j=1}^J|\alpha_j|,\qquad
 B_\beta=\sum_{j=1}^J|h\beta_{J,j}|\sqrt j.
\]
For fixed $J$, $A_\alpha$ and $B_\beta$ are independent of $h$.
Choose $K_*>0$ bounding the componentwise Lipschitz constants
of all starting functions, uniformly in $N$. Set
$\mathcal K_{y,j}=K_*$ for $n_*<j\leq N$ and, recursively, define
\begin{equation}\label{GS-K-recursion}
 \mathcal K_{y,i}
 :=\frac{s_h(1+hL_\infty A_\alpha)\mathcal K_{y,i+1}
        +hL_\infty(|\alpha_0|+s_hA_\alpha)}
        {1-|\alpha_0|hL_\infty},
 \qquad i=n_*,\ldots,0.
\end{equation}
Define independently
\begin{equation}\label{GS-Kz-definition}
 \mathcal K_{z,i}:=
       \sqrt{\frac d h}\,s_hB_\beta\mathcal K_{y,i+1},
       \qquad 0\leq i\leq n_*.
\end{equation}
These are the bounds specifying the admissible network classes in
the theorem. In particular, every admissible $\Psi_{y,i}$ satisfies
$\|\Psi_{y,i}(x)-\Psi_{y,i}(x')\|_\infty
 \leq\mathcal K_{y,i}|x-x'|$.

The deterministic regression targets are
\begin{align}
 y_i(x)
 &=\mathbb E[\Psi_{y,i+1}(X_{i+1}^{i,x})]
      +\alpha_0h f(t_i,x,y_i(x))\notag\\
 &\quad+h\sum_{j=1}^J\alpha_j\mathbb E\!
       \left[f(t_{i+j},X_{i+j}^{i,x},
                         \Psi_{y,i+j}(X_{i+j}^{i,x}))\right],
       \label{GS-y-def}\\
 z_i(x)
 &=\sum_{j=1}^J\beta_{J,j}\mathbb E\!
       \left[\Psi_{y,i+j}(X_{i+j}^{i,x})
                                  \Delta W_{i,i+j}^{\top}\right].
       \label{GS-z-def}
\end{align}
The equation for $y_i(x)$ is a contraction in the output maximum
norm. The Markov property and the common transition rule give
\[
 \mathcal U_i^m=y_i(X_i^{\pi,m}),\qquad
 \mathcal V_i^m=z_i(X_i^{\pi,m}).
\]

The coefficient of $\mathcal K_{y,i+1}$ in
\eqref{GS-K-recursion} is at least one, and its remaining term
is nonnegative. Thus
\begin{equation}\label{GS-cap-monotonicity}
 \mathcal K_{y,i}\geq\mathcal K_{y,i+1}
          \geq\mathcal K_{y,i+j},\qquad 1\leq j\leq J.
\end{equation}
For branches driven by the same random inputs, put
\[
 D_{i,j}(x,x'):=\Psi_{y,i+j}(X_{i+j}^{i,x})
                     -\Psi_{y,i+j}(X_{i+j}^{i,x'}).
\]
The componentwise network constraints and \eqref{GS-X-stability}
imply
\begin{align}
 \left(\mathbb E\|D_{i,j}(x,x')\|_\infty^2\right)^{1/2}
 &\leq\mathcal K_{y,i+j}
       \left(\mathbb E|X_{i+j}^{i,x}-X_{i+j}^{i,x'}|^2\right)^{1/2}
       \notag\\
 &\leq s_h\mathcal K_{y,i+1}|x-x'|.
       \label{GS-future-difference}
\end{align}
The same argument applies to the starting functions.
Subtracting \eqref{GS-y-def} at $x$ and $x'$, using the convexity
of the maximum norm and the Lipschitz bound for $f$, yields
\begin{align}
 &(1-|\alpha_0|hL_\infty)\|y_i(x)-y_i(x')\|_\infty\notag\\
 &\quad\leq
 \left[s_h\mathcal K_{y,i+1}+|\alpha_0|hL_\infty
       +hL_\infty A_\alpha s_h(1+\mathcal K_{y,i+1})\right]
                         |x-x'|.
       \label{GS-y-propagation}
\end{align}
Hence \eqref{GS-K-recursion} gives
\begin{equation}\label{GS-y-final-Lip}
 \|y_i(x)-y_i(x')\|_\infty\leq\mathcal K_{y,i}|x-x'|.
\end{equation}
For the matrix target, the elementary bound
$\|ab^\top\|_\infty\leq\|a\|_\infty|b|$ and
$\mathbb E|\Delta W_{i,i+j}|^2=djh$ give
\begin{align}
 \|z_i(x)-z_i(x')\|_\infty
 &\leq\sum_{j=1}^J|\beta_{J,j}|
       \left(\mathbb E\|D_{i,j}(x,x')\|_\infty^2\right)^{1/2}
                       (djh)^{1/2}\notag\\
 &\leq\sqrt{\frac d h}\,s_hB_\beta
                           \mathcal K_{y,i+1}|x-x'|
  =\mathcal K_{z,i}|x-x'|.
       \label{GS-z-final-Lip}
\end{align}
Thus both targets have exactly the componentwise Lipschitz bounds
required by their admissible GroupSort classes.

For sufficiently small $h$, \eqref{GS-K-recursion} implies
\[
 \mathcal K_{y,i}\leq(1+C_0h)\mathcal K_{y,i+1}+C_0h,
\]
where $C_0$ depends only on the fixed coefficients, $J,n,d$ and
the starting bounds. The discrete Gronwall inequality gives
\[
 \mathcal K_{y,i}
 \leq(1+C_0h)^{n_*+1-i}K_*
          +C_0h\sum_{q=0}^{n_*-i}(1+C_0h)^q
 \leq C.
\]
Together with \eqref{GS-Kz-definition}, this proves
\begin{equation}\label{GS-uniform-caps}
 \max_{0\leq i\leq n_*}\mathcal K_{y,i}
       +\max_{0\leq i\leq n_*}\sqrt h\,\mathcal K_{z,i}\leq C,
\end{equation}
with $C$ independent of $N$. The bounds are fixed before each
population minimization. The selected minimizer remains within
its constrained class, so the backward induction applies to
the optimized future networks as well as to the starting functions.

\medskip
\noindent Step 2. GroupSort approximation in Euclidean norms.
Let $R\geq1$, $\eta>0$, and $R\eta\leq1$.
Apply Lemma~\ref{gropusort-est} componentwise to $y_i$ and to the
$nd$ entries of $z_i$. There exist comparison networks
\[
 \Phi_{y,i}\in\mathcal G_{\mathcal K_{y,i},d,n}^{\zeta_\kappa,L,\mathbf n},
 \qquad
 \Phi_{z,i}\in\mathcal G_{\mathcal K_{z,i},d,nd}^{\zeta_\kappa,L,\mathbf n},
\]
where the latter output is reshaped as an $n\times d$ matrix,
such that
\begin{align}
 \sup_{x\in[-R,R]^d}|y_i(x)-\Phi_{y,i}(x)|
 &\leq2\sqrt n R\mathcal K_{y,i}\eta
  \leq CR\eta,\label{GS-y-uniform-error}\\
 \sup_{x\in[-R,R]^d}\sqrt h\,|z_i(x)-\Phi_{z,i}(x)|
 &\leq2\sqrt{ndh}R\mathcal K_{z,i}\eta
  \leq CR\eta.\label{GS-z-uniform-error}
\end{align}
These are comparison networks for the infima defining
$\varepsilon_i^y,\varepsilon_i^z$; they need not equal the
population minimizers. The factors $\sqrt n$ and $\sqrt{nd}$
enter only in these norm conversions and the subsequent error
estimates, not as multipliers in the time-step Lipschitz recursion.

\medskip
\noindent Step 3. Whole-space approximation errors.
Write $X_i=X_i^{\pi,m}$ and set
$D_{y,i}(x)=y_i(x)-\Phi_{y,i}(x)$,
$D_{z,i}(x)=z_i(x)-\Phi_{z,i}(x)$.
Then
\begin{equation}\label{GS-y-split}
 \varepsilon_i^y\leq
   \mathbb E[|D_{y,i}(X_i)|^2\mathbf1_{\{|X_i|\leq R\}}]
  +\mathbb E[|D_{y,i}(X_i)|^2\mathbf1_{\{|X_i|>R\}}].
\end{equation}
The local term satisfies
\begin{equation}\label{GS_y}
 \mathbb E[|D_{y,i}(X_i)|^2\mathbf1_{\{|X_i|\leq R\}}]
       \leq CR^2\eta^2.
\end{equation}
The componentwise Lipschitz bounds on both $y_i$ and
$\Phi_{y,i}$, followed by a single Euclidean norm conversion, give
\begin{align}
 |D_{y,i}(x)|
 &\leq |D_{y,i}(0)|+2\sqrt n\,\mathcal K_{y,i}|x|\notag\\
 &\leq C(R\eta+|x|),\qquad x\in\mathbb R^d.
       \label{GS-y-difference-growth}
\end{align}
Assumption~3 and the deterministic initial state $x_0$ provide
$\sup_i\mathbb E|X_i|^{2+\delta}\leq C$. Hence
\begin{equation}\label{GS-y-moment}
 \mathbb E|D_{y,i}(X_i)|^{2+\delta}\leq C,
\end{equation}
and Markov's inequality yields
\begin{equation}\label{GS-X-tail}
 \mathbb P(|X_i|>R)
       \leq R^{-(2+\delta)}\mathbb E|X_i|^{2+\delta}
       \leq CR^{-(2+\delta)}.
\end{equation}
By H\"older's inequality,
\begin{align}
 \mathbb E[|D_{y,i}(X_i)|^2\mathbf1_{\{|X_i|>R\}}]
 &\leq\bigl(\mathbb E|D_{y,i}(X_i)|^{2+\delta}\bigr)^{2/(2+\delta)}
                \mathbb P(|X_i|>R)^{\delta/(2+\delta)}\notag\\
 &\leq CR^{-\delta}.
       \label{GS-y-tail}
\end{align}
Combining \eqref{GS-y-split}, \eqref{GS_y}, and
\eqref{GS-y-tail} gives
\begin{equation}\label{Deep-NN-E-r-7}
 \varepsilon_i^y\leq C(R^2\eta^2+R^{-\delta}).
\end{equation}

Similarly,
\begin{equation}\label{GS-z-split}
 h\varepsilon_i^z\leq
   \mathbb E[h|D_{z,i}(X_i)|^2\mathbf1_{\{|X_i|\leq R\}}]
  +\mathbb E[h|D_{z,i}(X_i)|^2\mathbf1_{\{|X_i|>R\}}].
\end{equation}
The local error satisfies
\begin{equation}\label{GS_z}
 \mathbb E[h|D_{z,i}(X_i)|^2\mathbf1_{\{|X_i|\leq R\}}]
       \leq CR^2\eta^2.
\end{equation}
Using the componentwise bounds and then the Frobenius norm gives
\begin{align}
 \sqrt h\,|D_{z,i}(x)|
 &\leq\sqrt h\,|D_{z,i}(0)|
       +2\sqrt{ndh}\,\mathcal K_{z,i}|x|
  \leq C(R\eta+|x|),\label{GS-z-difference-growth}\\
 \mathbb E\bigl(\sqrt h\,|D_{z,i}(X_i)|\bigr)^{2+\delta}
 &\leq C.\label{GS-z-moment}
\end{align}
Consequently,
\begin{align}
 \mathbb E[h|D_{z,i}(X_i)|^2\mathbf1_{\{|X_i|>R\}}]
 &\leq\left[\mathbb E
       \bigl(\sqrt h\,|D_{z,i}(X_i)|\bigr)^{2+\delta}\right]^{2/(2+\delta)}
                  \mathbb P(|X_i|>R)^{\delta/(2+\delta)}\notag\\
 &\leq CR^{-\delta}.
       \label{GS-z-tail}
\end{align}
It follows that
\begin{equation}\label{Deep-NN-E-r-8}
 h\varepsilon_i^z\leq C(R^2\eta^2+R^{-\delta}).
\end{equation}
All constants in these estimates may depend on $n,d,J,T$,
the coefficient bounds, the starting Lipschitz bounds and the
initial moment, but are independent of $N,R,\eta,K_i$.

\medskip
\noindent Step 4. Convergence estimate.
Since $h=T/N$, \eqref{Deep-NN-E-r-7} and
\eqref{Deep-NN-E-r-8} imply
\begin{align}
 \sum_{i=0}^{n_*}\left(\frac{\varepsilon_i^y}{h}
                                   +h\varepsilon_i^z\right)
 &\leq C\left(\frac Nh+N\right)(R^2\eta^2+R^{-\delta})\notag\\
 &\leq CN^2(R^2\eta^2+R^{-\delta}).
       \label{GS-sum-eps}
\end{align}
The choices in \eqref{GS-parameter-choice} give
\begin{equation}\label{GS-R}
 R^{-\delta}=O(N^{-2J-2}),\qquad
 R^2\eta^2=O(N^{-2J-2}),\qquad R\eta=O(N^{-J-1}),
\end{equation}
and the Monte Carlo term satisfies
\begin{equation}\label{GS-sum-MC}
 \sum_{i=0}^{n_*}\frac1{hK_i}
 \leq\frac{N-J+1}{hK_{\mathrm{MC}}}
 \leq\frac{CN^2}{K_{\mathrm{MC}}}=O(N^{-2J}).
\end{equation}
Lemma~\ref{gropusort-est} gives, for $d>1$,
\begin{equation}\label{GS-grouping-final}
 \kappa=O\!\left(
    \left\lceil2\sqrt d\,
          N^{J+1+(2J+2)/\delta}\right\rceil\right),
\end{equation}
depth $L+1=O(d^2)$, and
\begin{equation}\label{GS-width-final}
 \sum_{\ell=0}^{L-1}n_\ell
   =O\!\left(\left(2\sqrt d\,
          N^{J+1+(2J+2)/\delta}\right)^{d^2-1}\right).
\end{equation}
For $d=1$, one can take $L+1=3$ and
\begin{equation}\label{GS-one-dimensional-final}
 \kappa=O\bigl(N^{J+1+(2J+2)/\delta}\bigr),\qquad
 \sum_{\ell=0}^{L-1}n_\ell
       =O\bigl(N^{J+1+(2J+2)/\delta}\bigr).
\end{equation}
Combining \eqref{GS-sum-eps}, \eqref{GS-R} and
\eqref{GS-sum-MC} with Theorem~\ref{MHDBR-err-esti}, and using
$\gamma_1\geq J$, yields
\begin{equation}\label{GS-trained-nodes}
 \max_{0\leq i\leq n_*}\mathbb E|Y_{t_i}-\mathcal Y_i|^2
       +h\sum_{i=0}^{n_*}\mathbb E|Z_{t_i}-\mathcal Z_i|^2
       \leq Ch^{2J}.
\end{equation}
This completes the proof.
\end{proof}
\end{appendices}
\\
\\
\\
\textbf{ \Large Declarations}
\\
\\
\textbf{Conflict of interest}\\
The authors declare that they have no conflict of interest.
The  authors contributed equally to this article. The authors read and approved the final manuscript.
\\
\\
\textbf{Funding}\\
This work was supported by the National Key R\&D Program of China (No.2023YFA1008701), the Key Project of the National Natural Science Foundation of China (No.12431017),
the National Natural Science of China (No.12501664) and
the National Natural Science of Yangzhou, China (No.YZ2025145).

\end{document}